\documentclass[nohyperref]{article}

\usepackage{microtype}
\usepackage{graphicx}
\usepackage{booktabs} %

\usepackage{hyperref}

\usepackage{algorithm}
\usepackage{algpseudocode}

\usepackage[accepted]{icml2023}

\usepackage{amsmath}
\usepackage{amssymb}
\usepackage{mathtools}
\usepackage{amsthm}
\usepackage{amsfonts}
\usepackage{caption}
\usepackage{subcaption}
\usepackage{booktabs}
\usepackage{bbm}
\usepackage{tikz}
\usetikzlibrary{bayesnet}
\usetikzlibrary{arrows}
\usetikzlibrary{backgrounds}
\usepackage{color}
\usetikzlibrary{decorations.pathmorphing}
\usetikzlibrary{hobby}
\usepackage{pst-all}

\usepackage[capitalize]{cleveref}

\usepackage{siunitx}

\usepackage{multirow}
\usepackage{lipsum}
\usepackage{pifont}

\theoremstyle{plain}

\theoremstyle{definition}

\theoremstyle{remark}

\usepackage[textsize=tiny]{todonotes}

\icmltitlerunning{Causal Modeling of Policy Interventions}

\DeclareMathOperator{\EX}{\mathbb{E}}
\newcommand*{\doOperator}{\operatorname{do}}
\newcommand*{\D}{\mathrm{d}}
\newcommand*{\hack}[1]{#1}

\newcommand{\cmark}{\ding{51}}%
\newcommand{\xmark}{\ding{55}}%

\newcommand\independent{\protect\mathpalette{\protect\independenT}{\perp}}
\def\independenT#1#2{\mathrel{\rlap{$#1#2$}\mkern2mu{#1#2}}}

\begin{document}

\twocolumn[
\icmltitle{Causal Modeling of Policy Interventions From Treatment--Outcome Sequences%
}

\icmlsetsymbol{equal}{*}

\begin{icmlauthorlist}
\icmlauthor{\c{C}a\u{g}lar H{\i}zl{\i}}{aaltocs}
\icmlauthor{ST John}{aaltocs}
\icmlauthor{Anne Juuti}{hus}
\icmlauthor{Tuure Saarinen}{hus}
\icmlauthor{Kirsi Pietiläinen}{hus}
\icmlauthor{Pekka Marttinen}{aaltocs}
\end{icmlauthorlist}

\icmlaffiliation{aaltocs}{Department of Computer Science, Aalto University, Helsinki, Finland}
\icmlaffiliation{hus}{Helsinki University Hospital and University of Helsinki, Helsinki, Finland}

\icmlcorrespondingauthor{\c{C}a\u{g}lar H{\i}zl{\i}}{caglar.hizli@aalto.fi}
\icmlkeywords{Machine Learning, ICML}

\vskip 0.3in
]

\printAffiliationsAndNotice{}  %
\begin{abstract}
A \emph{treatment policy} defines when and what treatments are applied to affect some outcome of interest. Data-driven decision-making requires the ability to predict \emph{what happens if a policy is changed}. Existing methods that predict how the outcome evolves under different scenarios assume that the tentative sequences of future treatments are fixed in advance, while in practice the treatments are determined stochastically by a policy and may depend, for example, on the efficiency of previous treatments. Therefore, the current methods are not applicable if the treatment policy is unknown or a counterfactual analysis is needed. To handle these limitations, we model the treatments and outcomes jointly in continuous time, by combining Gaussian processes and point processes. Our model enables the estimation of a treatment policy from observational sequences of treatments and outcomes, and it can predict the interventional and counterfactual progression of the outcome \emph{after an intervention on the treatment policy} (in contrast with the causal effect of a single treatment). We show with real-world and semi-synthetic data on blood glucose progression that our method can answer causal queries more accurately than existing alternatives.
\end{abstract}

\begin{figure*}[t]
    \centering
    \includegraphics[width=1.0\textwidth]{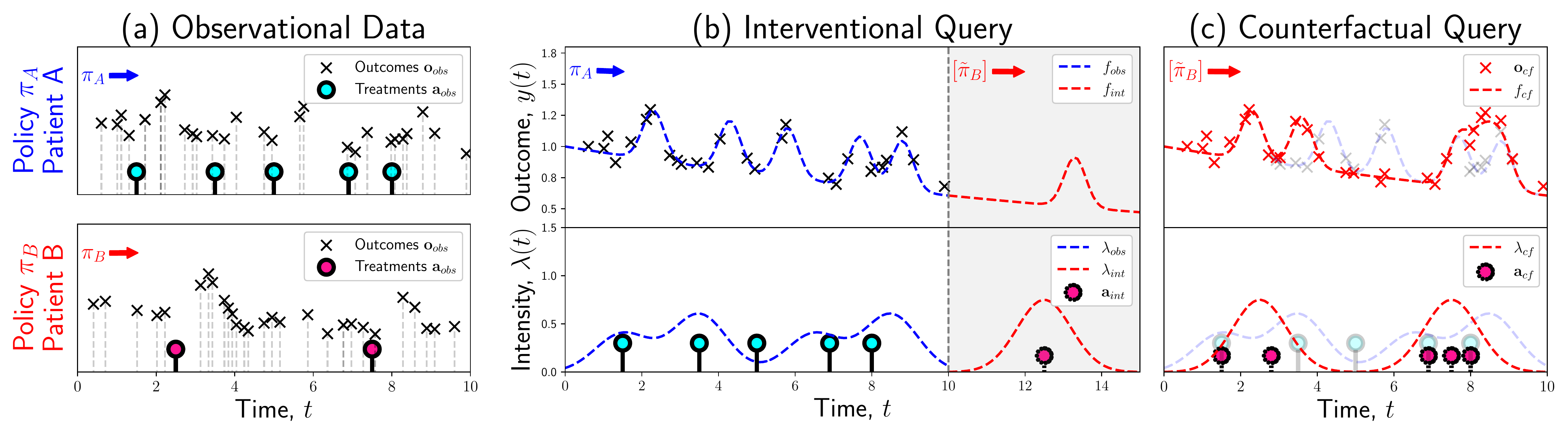}
    \caption{\textbf{(a)} Sequential treatment--outcome data for patients {\color{blue} A} (top) and {\color{red} B} (bottom). Patients follow distinct policies {\color{blue} $\pi_A$} and {\color{red}$\pi_B$} in the observation period $[0,10]$, where patient {\color{blue}A} is treated more frequently. A treatment causes a (slightly delayed) increase or `bump' in the outcome.
    \textbf{(b, c)} We now focus on patient {\color{blue}A} and show outcome trajectory $f$ (top) and treatment intensity $\lambda$ (bottom) in dashed lines. The estimates based on the observational data are in {\color{blue}blue}.
    \textbf{(b)} The interventional query corresponds to how the outcome  trajectory of {\color{blue}A} will progress \emph{after} the observation period (shaded area) under a treatment intensity induced by a \emph{different} policy {\color{red}$\pi_B$}, shown in {\color{red}red}.
    \textbf{(c)} The counterfactual query corresponds to how the outcome trajectory of {\color{blue}A} \emph{would have} progressed if the policy in the observation period $[0, 10]$ had been set to {\color{red}$\pi_B$} instead. 
    The observational data and estimations are shown in the background, while the counterfactual data and predictions are shown in {\color{red}red}.
    Notice some of the observed treatments are kept as counterfactual treatments $\mathbf{a}_\textit{cf}$, where the counterfactual intensity $\lambda_\textit{cf}$ is higher than the observational intensity $\lambda_\textit{obs}$. Counterfactual treatments in turn affect the counterfactual trajectory $f_\textit{cf}$.}
    \label{fig:compare}
\end{figure*}

\section{Introduction}

What policy should we adopt? In healthcare, for example, we observe patients' physiological markers (\emph{outcomes}) that change over time. %
We want to (positively) affect these outcomes %
by actions (\emph{treatments}) such as doses of a medicine. Sequences of outcomes and treatments are recorded as a time series. %
A \emph{policy} dictates what actions to take and when. To improve our policies, we must be able to assess their consequences: What is the effect of a given policy? What will be the effect of a change to a different policy? What would have happened if a patient had followed a different treatment policy? These questions correspond to observational, interventional, and counterfactual queries, %
and answering them is crucial, especially in domains such as public health and healthcare, where quantifying risks and expectations accompanying a policy decision, as well as evaluating past policies, is essential \citep{schulam2017reliable,bica2020estimating,oberst2019counterfactual,tsirtsis2020decisions,tsirtsis2021counterfactual}. Answering the interventional and counterfactual queries requires modeling policy interventions using a \emph{causal} model.

We want to infer a causal model of recurrent treatments from sequential treatment--outcome data.
Such data are always created by some policy, which induces dependencies between past events and future treatments.
However, the policy itself is generally not recorded and may be known only implicitly from the observed distribution of treatments and outcomes. 
In epidemiology, this problem has been studied using linear models for discrete-time sequences \citep{robins1986new,robins2009estimation,taubman2009intervening,zhang2018comparing},
though these works are unable to capture non-linear, long-term dependencies in real-world, continuous-time treatment--outcome sequences \citep{bica2021real}. \looseness-1

To handle such complex, long-term dependencies, sequential treatment--outcome models based on Bayesian nonparametrics and neural networks have been proposed \citep{xu2016bayesian,schulam2017reliable,soleimani2017treatment,bica2020estimating,seedat2022continuous}. However, they neglect the policy-induced link between past events and future treatments, and their causal analyses are limited to estimating the impact of a fixed sequence of treatments set by hand or generated by a simplistic parametric model. Such models do not generalize beyond simulations to the analysis of realistic treatment policies in real-world applications. Recently, \citet{hua2021personalized} modeled treatments explicitly; however, they focused on optimizing treatment strategies instead of causal estimation and only considered a parametric model tailored for a particular application.

With an appropriate causal model, we can also evaluate treatment policies using counterfactual reasoning, which allows for learning from mistakes by considering alternative scenarios to past events \citep{epstude2008functional,oberst2019counterfactual}. This is not considered by most of the literature, which focuses on future outcome progression. One recent work \citep{noorbakhsh2021counterfactual} applies counterfactual reasoning to event data using a counterfactual temporal point process, but does not consider a treatment--outcome setup (for further related work, see \cref{sec:related_work}).

To address these limitations, we propose a joint treatment--outcome model to estimate treatment policies and treatment responses in continuous time. Our contributions are:

\textbf{Joint treatment--outcome model.} We combine a marked point process and a conditional Gaussian process (\textsc{Gp}). Our \emph{non-parametric} model can be learned from observational sequential treatment--outcome data (Fig. 1(a)) and can estimate future and counterfactual progression.

\textbf{Interventional and counterfactual queries of interventions on policies.} We show that an intervention on a treatment policy is equivalent to a sequence of stochastic interventions on treatments, which we can model with our joint model, and use this to answer interventional (Fig. 1(b)) and counterfactual queries (Fig. 1(c)).

\textbf{Counterfactual sampling of arbitrary point processes.} We extend the algorithm of \citet{noorbakhsh2021counterfactual} from Poisson processes (independent events) to arbitrary point processes, allowing events to depend on past events.

We fit our model to a real-world data set on blood glucose progression, and from this generate realistic semi-synthetic use cases, where we show that our model accurately answers interventional and counterfactual policy queries. %

\section{Preliminaries}

Our work builds on Gaussian processes (for more details see \citet{williams2006gaussian}), marked point processes (\cref{sec:markedpointprocesses}) and causal inference (\cref{sec:causalinference}).

\subsection{Marked Point Processes (\textsc{Mpp})}
\label{sec:markedpointprocesses}

A temporal point process is a stochastic process that models a set of ordered random points $\{t_i\}_{i=1}^N$ on an interval $[0,T]$
\citep{daley2003introduction,rasmussen2011temporal}. This process is uniquely determined by its conditional intensity function $\lambda^*(\tau) = \lambda(\tau \mid \mathcal{H}_{<\tau})$, where the star superscript ${}^*$ denotes dependence on the past history $\mathcal{H}_{<\tau}$. The intensity function describes the instantaneous rate, i.e., the expected number of events in an infinitesimal interval $[\tau, \tau+\D\tau)$.

When each event time $t_i$ is associated with additional information (`mark')  $m_i$, we model $\mathcal{D} = \{(t_i, m_i)\}_{i=1}^N$ with a marked point process (\textsc{Mpp}). For an \textsc{Mpp}, the conditional intensity function additionally includes the conditional mark probability $p^*(m \mid t)$: $\lambda^*(t, m) = \lambda^*(t) p^*(m \mid t)$. The likelihood of $ \mathcal{D}$ observed in the interval $[0,T]$ is given by\looseness-1
\begin{equation*}
    \label{eq:mpp_lik}
    p(\mathcal{D} \mid \lambda^*(t, m)) = \left( \prod_{i=1}^N \lambda^*(t_i) p^*(m_i \mid t_i) \right) \exp( -\Lambda ),
\end{equation*}
where $\Lambda$ denotes the integral term $\Lambda = \int_{[0,T]} \lambda^*(\tau) \D\tau$.

\subsection{Causal Inference}
\label{sec:causalinference}

Causal effects are widely studied in the language of potential outcomes \citep[\textsc{Po},][]{neyman1923applications,rubin1978bayesian} or structural causal models \citep[\textsc{Scm},][]{pearl2009causality}. In this section, we briefly introduce both frameworks, as we use \textsc{Po}s to define causal queries formally and show their identifiability, while using \textsc{Scm}s to emphasize the distinction between probability distributions induced by different types of target causal queries.

\subsubsection{Potential Outcomes}

Let $A$ and $Y$ denote a treatment and an outcome variable. In the \textsc{Po} framework, the primitive building block is the potential outcome $Y[\tilde{a}]$, which is a random variable that represents the value of the outcome $Y$ under a treatment intervention $[A=\tilde{a}]$. The treatment intervention $[\tilde{a}]$ corresponds to the \textit{do-operator} $\doOperator(A=\tilde{a})$ in the \textsc{Scm} framework: $P(Y[\tilde{a}]) \equiv P(Y \mid \doOperator(A=\tilde{a}))$ \citep{pearl2009causality}.

\subsubsection{Observational, Interventional and Counterfactual Distributions}

\newcommand*{\DAGParent}{\operatorname{pa}}
For a set of variables $\mathbf{X} = \{X_k\}_{k=1}^K$, an \textsc{Scm} $\mathcal{M} = (\mathbf{S}, p(\mathbf{N}))$ consists of (i) a set of structural assignments $\mathbf{S} = \{X_k := f_k(\DAGParent(X_k), N_k)\}_{k=1}^K$, where $\DAGParent(\cdot)$ returns the parents of a given variable, and (ii) a distribution over noise variables $\mathbf{N} \sim p(\mathbf{N})$. A causal graph $\mathcal{G}$ is obtained by representing each variable $X_k$ as a node and drawing edges to $X_k$ from its parents $\DAGParent(X_k)$.
The \textsc{Scm} defines a generative process for $\mathbf{X}$ by sampling $\mathbf{N}$ from $p(\mathbf{N})$ and performing ancestral sampling over structural assignments $\mathbf{S}$, inducing an observational distribution $\mathbf{X} \sim p(\mathbf{X})$.
Each data point $\mathbf{x}^{(n)} = \{x_k^{(n)}\}_{k=1}^K$ in the observed data $ \mathcal{D} = \{\mathbf{x}^{(n)}\}_{n=1}^N$ is considered an i.i.d.~sample from the observational distribution.\looseness-1

An \emph{intervention} $\doOperator(X_k = \tilde{x}_k)$ is performed by removing the assignment $f_k(\cdot, \cdot)$ and setting $X_k$ to $\tilde{x}_k$. This yields an interventional \textsc{Scm} $\mathcal{M}_{\text{int}} = (\mathbf{S}_{\text{int}}, p(\mathbf{N}))$, inducing an interventional distribution: $p(\mathbf{X} \mid \doOperator(X_k = \tilde{x}_k)) \equiv p(\mathbf{X}[\tilde{x}_k])$.

In the context of an observed data point $\mathbf{x}^{(n)}$, we define a \emph{counterfactual} \textsc{Scm} by plugging in the individual's posterior noise distribution $p(\mathbf{N} \mid \mathbf{x}^{(n)})$ in place of the general (prior) noise distribution $p(\mathbf{N})$ and then updating the relevant structural assignment: $\mathcal{M}_{\text{cf}} = (\mathbf{S}_{\text{int}}, p(\mathbf{N} \mid \mathbf{x}^{(n)}))$. The resulting joint distribution $p(\mathbf{X} \mid \mathbf{x}^{(n)}, \doOperator(X_k = \tilde{x}_k)) \equiv p(\mathbf{X}[\tilde{x}_k] \mid \mathbf{x}^{(n)})$ is called the counterfactual distribution.

We categorize causal queries by the required joint distribution to answer them: (i) an interventional query requires access to the interventional distribution and (ii) a counterfactual query requires access to the counterfactual distribution. The key difference between the two queries is that a counterfactual query requires the posterior distribution of noise variables $p(\mathbf{N} \mid \mathbf{x}^{(n)})$ before performing the intervention, while the interventional query uses the noise prior $p(\mathbf{N})$.

\section{Problem Definition}
Consider an observational data set $\mathcal{D}$:
\begin{align}
    \label{eq:dataset}
    \mathcal{D} &= \big\{ \underbrace{\pi_{[0,T]}}_{\text{policy label}}, \underbrace{ \{(t_i, m_i)\}_{i=1}^{N_a} }_{\text{treatments } \mathbf{a}}, \underbrace{ \{(t_j, y_j)\}_{j=1}^{N_o} }_{\text{outcomes } \mathbf{o}} \big\},
\end{align}
in a period $[0,T]$. For notational ease, the data set is defined for a single patient. Our model can be trivially generalized to multiple patients, under exchangeability.

\begin{figure}[t]
\resizebox{\columnwidth}{!}{
\tikzset{mynode/.style={obs, minimum size=1.25cm}}
\tikzset{policy/.style={mynode, fill=red!10}}
\tikzset{action/.style={mynode, fill=blue!10}}
\tikzset{observation/.style={mynode, fill=yellow!10}}
\tikz{
    \node[text width=3cm] at (1.0, 0.25) {\huge{$\ldots$}};
    \node[observation] at (-7.2, 0.0)   (o1) {\huge$o_1$};
    \node[action] at (-5.55, -0.5)   (a1) {\huge$a_1$};
    \node[observation] at (-3.2, 1.4)   (o2) {\huge$o_2$};
    \node[action] at (3.25, -0.5)   (a5) {\huge$a_i$};
    \node[observation] at (4.2, 1.6)   (o5) {\huge$o_j$};
    \node[action] at (6.1, -0.5)   (a6) {\LARGE$a_{i+1}$};
    \node[observation] at (7.5, 0.6)   (o6) {\LARGE$o_{j+1}$};
    \node[policy] at (-0.15, -2.8)   (pi) {\huge$\pi$};
\edge {o1} {a1}
\edge {o1,a1} {o2}
\edge {a5,a6,o5} {o6}
\edge {a5,o5} {a6}
\edge {a5} {o5}
\edge {pi} {a1,a5,a6}
\begin{scope}[on background layer]
    \draw (0, 0) node[inner sep=0]{\includegraphics[width=1.0\textwidth]{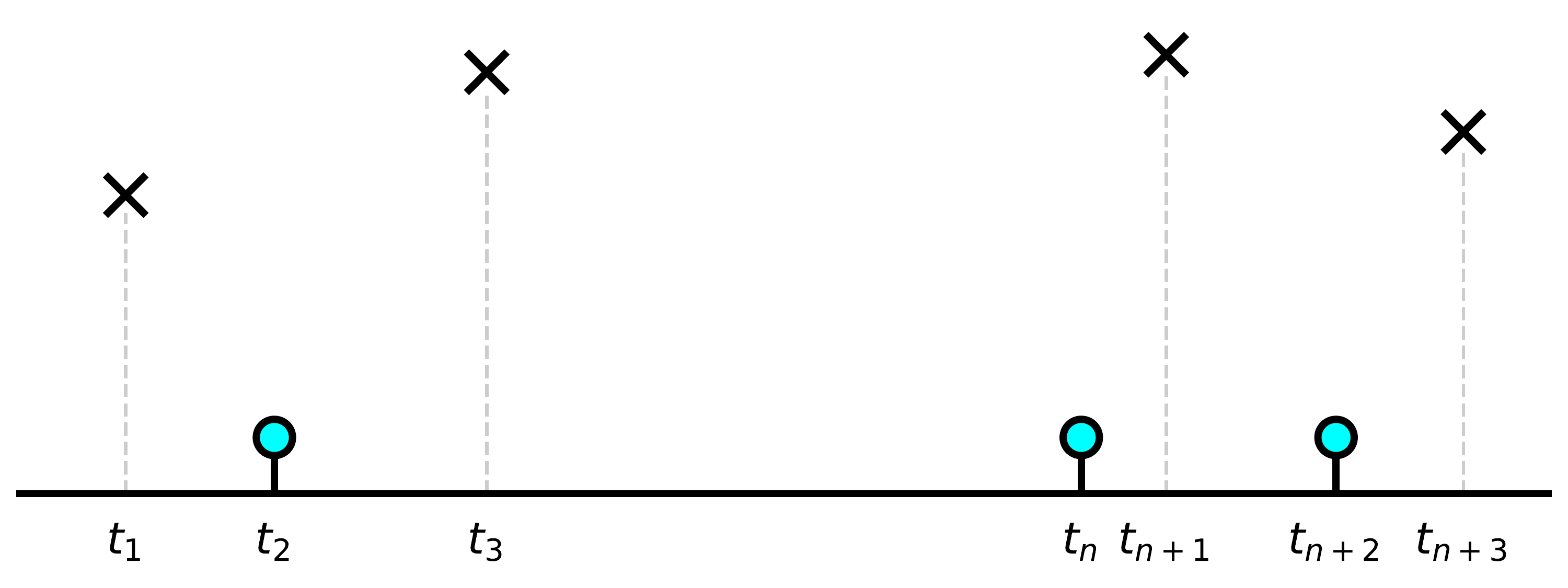}};
    \draw (0, 0) node[inner sep=0, minimum height=7.0cm,minimum width=1.0\textwidth,fill=white,opacity=0.1]{};
 \end{scope}
}
}
\caption{The causal graph $\mathcal{G}$ assumes a sequential treatment--outcome setup in continuous time, where past treatments and outcomes have a causal effect on future treatments and outcomes. The causal effect of the policy $\pi$ on outcomes is fully mediated through treatments, i.e.~there are no direct edges between the policy $\pi$ and outcomes $\mathbf{o}$.}
\label{fig:causal_graph}
\end{figure}
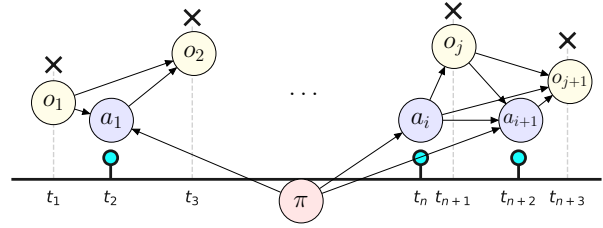

A policy is a pair $(\pi_{[0,T]}, \lambda^*_{\pi}(t,m))$ that defines when and how treatments are applied, and is effectively determined by its treatment intensity function $\lambda^*_{\pi}(t,m)$. The policy label $\pi_{[0,T]}$ is assumed fixed for the period $[0,T]$ stated in the subscript, and it is assumed observed, while the corresponding intensity $\lambda^*_{\pi}(t,m)$ is unobserved. We assume that patients belong to different groups, and each group is characterized by a shared policy (treatment intensity). Background knowledge of having, e.g., different hospitals, countries, or environments in the observed data set implies distinct policies. In the absence of such information, each patient can be assumed to have their own policy.

A treatment tuple $a_i=(t_i,m_i)$ consists of an arrival time $t_i \in [0,T]$ and a treatment mark (dosage) $m_i \in \mathbb{R}$. An outcome tuple $o_j=(t_j, y_j)$ consists of a measurement time $t_j \in [0,T]$ and an outcome value $y_j \in \mathbb{R}$. Treatment times $\mathbf{t_a}=\{t_i\}_{i=1}^{N_a}$ and outcome times $\mathbf{t_o}=\{t_j\}_{j=1}^{N_o}$ are irregularly sampled points on the interval $[0,T]$. The history $\mathcal{H}_{\leq t} = \{\pi_{\leq t}, \mathbf{a}_{\leq t}, \mathbf{o}_{\leq t}\}$ contains the information about the past policy $ \pi_{\leq t}$, past treatments $\mathbf{a}_{\leq t} = \{(t_i, m_i): t_{i} \leq t\}$ and past outcomes $\mathbf{o}_{\leq t} = \{(t_{j}, y_j): t_{j} \leq t\}$.

We observe a continuous-time process $\mathbf{Y}_{\leq T} = \{y(\tau): \tau \leq T\}$ as outcome tuples $\mathbf{o}$ measured at times $\mathbf{t_o} = \{t_j\}_{j=1}^{N_o}$. To answer causal queries, we model the potential outcome trajectory $\mathbf{Y}_{>\tilde{\tau}}[\tilde{\pi}_{>\tilde{\tau}}]$, under an intervened policy specified by $\tilde{\pi}_{>\tilde{\tau}}$, where the subscript emphasizes the period when a policy intervention takes place. When the intervention time $\tilde{\tau}$ is set to the end of the observation period $\tilde{\tau}=T$, we call the estimation task a \textit{policy intervention}, as its computation requires access to the interventional distribution (Fig. 1(b)):
\begin{align}
\label{eq:target_intervention}
    P(\mathbf{Y}_{>T}[\tilde{\pi}_{>T}] \mid \mathcal{H}_{\leq T}).
\end{align}
Also, we can set the intervention time $\tilde{\tau}$ to the start of the observation period $\tilde{\tau}=0$ and consider a hypothetical scenario under an alternative treatment policy specified by $\tilde{\pi}_{[0,T]}$. We call this estimation task a \textit{policy counterfactual}, as its computation requires access to the counterfactual distribution:
\begin{align}
\label{eq:target_counterfactual}
    P(\mathbf{Y}_{[0,T]}[\tilde{\pi}_{[0,T]}] \mid \mathcal{H}_{\leq T}).
\end{align}
The difference between a policy intervention and a policy counterfactual is illustrated in Figs.~\ref{fig:compare}~(b) and~(c).

\section{Causal Assumptions and Identifiability}

In this section, we first show that a policy intervention query, formalized in \cref{eq:target_intervention}, is equivalent to the potential outcome trajectory of a sequence of stochastic (conditional) interventions on treatments. Then, we show the latter is identifiable under causal assumptions, i.e., it can be estimated solely based on statistical terms \citep{pearl2009causality}.

\newcommand*{\obsPolicy}{\pi}
\newcommand*{\intPolicy}{\tilde{\pi}_{>T}}

We assume a sequential treatment--outcome setup where past treatments and outcomes may affect future events \citep{robins1986new,robins1987addendum}. We add the policy label $\pi$ to this setup, which has a direct causal effect on sequential treatments $\mathbf{a}$. The interplay between a policy label $\pi$, treatments $\mathbf{a}$, and outcomes $\mathbf{o}$ is illustrated in the causal graph $\mathcal{G}$ in \cref{fig:causal_graph}.
We make the standard causal assumptions of \emph{consistency} (Assumption 1) and \emph{no unobserved confounders} (\textsc{Nuc}, Assumption 2), and furthermore the continuous-time \textsc{Nuc} \citep[Assumption 3,][see formal definitions in \cref{sec:causal-assump}]{schulam2017reliable}. In words, the continuous-time \textsc{Nuc} states that treatments are assigned stochastically in continuous time, conditioned on the past history. In addition, we assume that the causal effect of the policy label $\pi$ on outcomes $\mathbf{o}$ is fully mediated through treatments $\mathbf{a}$, such that the causal graph $\mathcal{G}$ excludes direct edges from the policy label $\pi$ to the outcomes $\mathbf{o}$, stated formally as:

\textbf{Assumption 4: Fully-mediated policy effect.} Conditioned on the past history $\mathcal{H}_{\leq\tau}$, the causal effect of the policy specified by $\tilde{\pi}_{>\tau}$ on the outcome trajectory $\mathbf{Y}_{>\tau}$ is fully mediated through sequential treatments $\tilde{\mathbf{a}}_{>\tau}$: 
$$P(Y_{>\tau}[\tilde{\pi}_{>\tau}, \tilde{\mathbf{a}}_{>\tau}] \mid \mathcal{H}_{\leq \tau}) = P(Y_{>\tau}[\tilde{\mathbf{a}}_{>\tau}] \mid \mathcal{H}_{\leq \tau}).$$

Using Assumptions \{1,2,3,4\}, we inspect the potential outcome trajectory $\mathbf{Y}_{>T}[\intPolicy]$ at a discrete set of ordered query times $\mathbf{q}=\{q_1, \ldots, q_m\}$ (all $q_k > T$): $\mathbf{Y_q}[\tilde{\pi}_{>T}] = \{y(q_k)[\intPolicy]: q_k \in \mathbf{q}\}$ \citep{schulam2017reliable}. We factorize the outcome query $\mathbf{Y_q}[\intPolicy]$ in temporal order, and develop Theorem 1 below to reformulate it as the sequential estimation of treatments and outcomes:

\textbf{Theorem 1.} Under Assumptions \{1,2,3,4\}, the potential outcome query $\mathbf{Y_q}[\tilde{\pi}_{>T}]$ under a policy $\tilde{\pi}_{>T}$ is equivalent to the potential outcome query under a sequence of stochastic (conditional) interventions on treatments $\tilde{\mathbf{a}}_{>T}$:
\vspace{-1ex}
\begin{align}
    \label{eq:po_factorized}
    &P(\mathbf{Y_q}[\tilde{\pi}_{>T}] \mid \mathcal{H}_{\leq T}) = \sum_{\tilde{\mathbf{a}}_{>T}} \prod_{k=0}^{m-1} \\
    &\qquad\! P(\tilde{\mathbf{a}}_{[q_k, q_{k+1})}[\tilde{\pi}_{>T}] | \mathcal{H}_{\leq q_k}) \, P(Y_{q_{k+1}}[\tilde{\mathbf{a}}_{[q_k, q_{k+1})}] | \mathcal{H}_{\leq q_k}) , \nonumber
\end{align}
where $q_0 = T$, and $\tilde{\mathbf{a}}_{[q_{k}, q_{k+1})}$ denotes treatments in the interval $[q_{k}, q_{k+1})$ without outcome observations between consecutive query times. Here, the (whole) treatment sequence $\tilde{\mathbf{a}}_{>T}$ follows a distribution $p(\tilde{\mathbf{a}}_{>T}[\tilde{\pi}_{>T}])$, induced by the intervened policy label $\tilde{\pi}_{>T}$. The proof of Theorem 1 is given in \cref{sec:causal-iden}. Accordingly, the potential outcomes $\tilde{\mathbf{a}}_{[q_k, q_{k+1})}[\tilde{\pi}_{>T}]$ and $Y_{q_{k+1}}[\tilde{\mathbf{a}}_{[q_k, q_{k+1})}]$ can be identified under Assumptions \{1,2,3\}, using the following conditionals, both of which can be estimated with a statistical model \citep{schulam2017reliable,seedat2022continuous}:
\begin{align}
    \label{eq:stats_quantities}
    &P(\mathbf{Y_q}[\tilde{\pi}_{>T}] \mid \mathcal{H}_{\leq T}) = \sum_{\tilde{\mathbf{a}}_{>T}} \prod_{k=0}^{m-1} \\
    &\qquad\!\underbrace{P(\tilde{\mathbf{a}}_{[q_k, q_{k+1})} | \tilde{\pi}_{>T}, \mathcal{H}_{\leq q_k})}_{\text{Treatment Term}} \underbrace{P(Y_{q_{k+1}} | \tilde{\mathbf{a}}_{[q_k, q_{k+1})}, \mathcal{H}_{\leq q_k})}_{\text{Outcome Term}} . \nonumber
\end{align}
The identifiability result by \citet{schulam2017reliable} was limited to a fixed sequence of treatments. Theorem 1 and the identifiability result in \cref{eq:stats_quantities} generalize that result to a stochastic (conditional) intervention on a sequence of treatments, and one can recover the fixed sequence of treatments by assuming that the target density $p_{\tilde{\pi}}(\tilde{\mathbf{a}}_{>T})$ puts all probability on a single sequence of treatments. The identifiability of a policy counterfactual is more challenging and discussed in \cref{sec:inference}, since it requires functional assumptions related to the model definition (\cref{sec:joint-model}).

\section{Joint Treatment--Outcome Model}
\label{sec:joint-model}

To estimate both statistical terms in \cref{eq:stats_quantities}, (i) Treatment Term and (ii) Outcome Term, from observational data, we propose a joint treatment--outcome model, combining a marked point process and a conditional Gaussian process.

We encode conditional independence statements in the graph $\mathcal{G}$ by defining two dependent \textsc{Mpp}s with conditional intensity functions: (i) treatment intensity for a treatment $a=(t,m)$: $\lambda_{\pi}^*(t, m) = \lambda_{\pi}^*(t) p^*(m \mid t)$ and (ii) outcome intensity for an outcome $o=(t,y)$: $\lambda_o^*(t, y) = \lambda_o^*(t) p^*(y \mid t)$. The past history $\mathcal{H}_{<t} = \{ \pi_{<t}, \mathbf{a}_{<t}, \mathbf{o}_{<t} \}$ containing the information about the policy, past actions, and past outcomes is a valid history for both conditional intensity functions $\lambda_{\pi}^*(t, m)$ and $\lambda_o^*(t, y)$. The joint distribution for the observational data set $\mathcal{D}$ in \cref{eq:dataset} can be written in terms of treatment and outcome intensity functions:
\begin{equation*}
    p(\mathcal{D}) = \prod_{i=1}^I \lambda_{\pi}^*(t_i,m_i)
    \prod_{j=1}^J \lambda_o^*(t_j,y_j) \exp(-\Lambda),
\end{equation*}
with the integral term $\Lambda = \int_{[0,T]} \big(\lambda_{\pi}^*(\tau) + \lambda_{o}^*(\tau)\big) \D\tau$.

We further assume that the measurement times of the outcomes $\mathbf{t_o} = \{ t_j \}_{j=1}^{N_o}$ are given, which is valid for example when the data are collected through automated patient monitoring in healthcare. This assumption is equivalent to setting the outcome time intensity to an indicator function $\mathbbm{1}_{\mathbf{t_o}}(t)$:\footnote{The indicator function is: $\mathbbm{1}_{\mathbf{t_o}}(t) = \begin{cases}
      1, & \text{if}\ t \in \mathbf{t_o} \\
      0, & \text{otherwise}
    \end{cases}$}
$\lambda_o^*(t, y) = \mathbbm{1}_{\mathbf{t_o}}(t) p^*(y \mid t) = p^*(y \mid t)|_{t \in \mathbf{t_o}}$. Then, the joint distribution becomes
\begin{multline}
    p(\mathcal{D}) = \prod_{i=1}^I \underbrace{\lambda_{\pi}^*(t_i)  p^*_{\pi}(m_i \mid t_i)}_{\text{Treatment Intensity}}
    \prod_{j=1}^J \underbrace{p^*(y_j \mid t_j)}_{\text{Outcome Model}}|_{t_j \in \mathbf{t_o}} \\
    \times \exp(-\textstyle\int_{[0,T]} \lambda_{\pi}^*(\tau) \D\tau) .
    \label{eq:joint_intensity}
\end{multline}

\subsection{Treatment Intensity}
\label{sub:action}

We consider the treatment time intensity $\lambda^*_{\pi}(\tau)$ as the output of a square transformation of a latent function $g^*_{\pi}(\tau)$, which we model as the sum of a constant baseline $\beta_0$ and three time-dependent functions with \textsc{Gp} priors, $g_b, g^*_a, g^*_o \sim \mathcal{GP}$.
We assume the patients who follow the same treatment policy $\pi$ share the same treatment model, i.e., the same hyperparameters, but the intensity curves may differ because past treatments and outcomes are not the same.

The latent-state function $g_b(\tau)$ models the history-independent intensity. The regressive component $g^*_a(\tau)$ models the dependence on past treatments by taking the last $Q_a$ treatments as input and the regressive component $g^*_o(\tau)$ models the dependence on past outcomes by taking the last $Q_o$ outcomes as input. The treatment intensity $\lambda^*_{\pi}(\tau)$ is \looseness-1 %
\begin{equation}
    \label{eq:treatment_intensity}
    \lambda^*_{\pi}(\tau) = \big( \underbrace{\beta_0}_{\substack{\text{Poisson} \\ \text{Baseline}}} + \underbrace{g_b(\tau)}_{\substack{\textsc{Nhpp} \\ \text{Baseline}}} + \underbrace{g^*_a(\tau; \mathbf{a})}_{\substack{\text{Treatment} \\ \text{Effect}}} + \underbrace{g^*_o(\tau; \mathbf{o})}_{\substack{\text{Outcome} \\ \text{Effect}}} \big)^2 .
\end{equation}
The resulting treatment intensity extends the model proposed by \citet{liu2019nonparametric} with (i) additional baseline components $\beta_0$ and $g_b(\tau)$ and (ii) the functional dependence on marks.
Without regressive components $g^*_a(\tau)$ and $g^*_o(\tau)$, the treatment model becomes a non-homogeneous Poisson process (\textsc{Nhpp}), whose instantaneous intensity is independent of past events: $\lambda(\tau) = \smash{(\beta_0 + g_b(\tau))^2}$. If we further exclude the time-varying baseline $g_b(\tau)$, the model becomes a simple Poisson process with constant intensity. We model the treatment dosages (marks) by a \textsc{Gp}-prior: $m(\tau) \sim \mathcal{GP}$. The model definition is detailed in \cref{sec:treatment-details}.

\subsection{Outcome Model}

We model the outcome trajectory $\mathbf{Y} = \{y(\tau): \tau \in \mathbb{R}_{\geq 0}\}$ over time $\tau$ by a conditional \textsc{Gp} model, combining three independent components: (i) a baseline progression, (ii) a treatment response curve, and (iii) a noise variable \citep{schulam2017reliable,cheng2020patient,zhang2020errors}: %
\begin{equation}
    \label{eq:outcome}
    y(\tau) = \underbrace{f_b(\tau)}_{\text{Baseline}} + \underbrace{f_a(\tau; \mathbf{a})}_{\text{Treatment Response}} + \underbrace{\epsilon(\tau)}_{\text{Noise}},
\end{equation} %
with independent Gaussian noise $\epsilon(\tau) \sim \mathcal{N}(0,\sigma^2_{\epsilon})$.

\textbf{Baseline Progression.} This is modeled by a \textsc{Gp} prior on $f_b(\tau) \sim \mathcal{GP}$. \citet{cheng2020patient} proposed a sum of a squared exponential (\textsc{Se}) and a periodic kernel to model recurring patterns in heart rates. In our experiments, we model the baseline progression of blood glucose as a sum of a constant and a periodic kernel to capture daily blood glucose profiles of non-diabetic patients fluctuating around a constant baseline \citep{ashrafi2021computational}. We assume that the outcome baseline $f_b$ is patient-specific.

\begin{figure}[t]
    \centering
    \includegraphics[width=\linewidth]{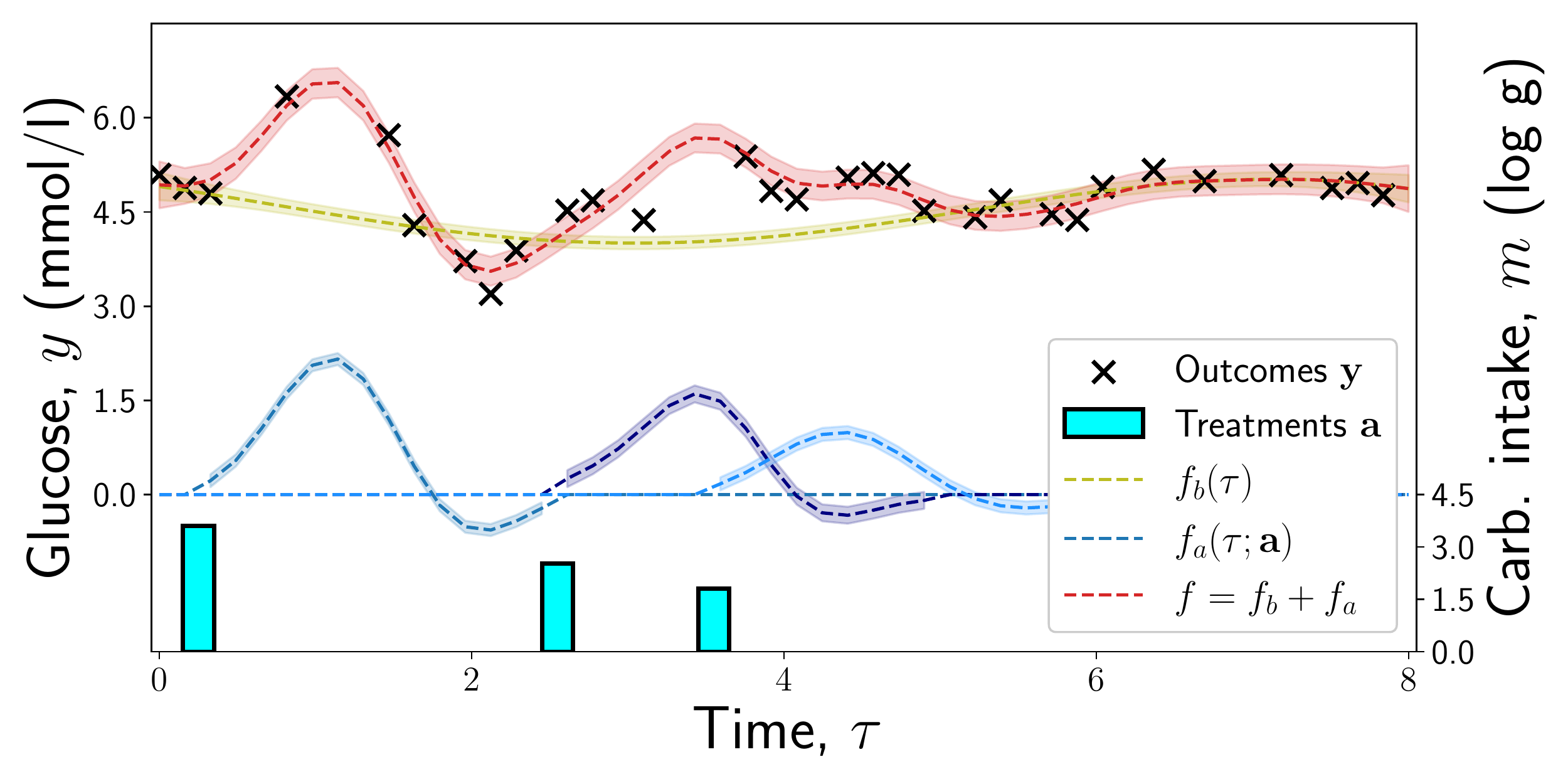}
    \hack{\vspace{-1ex}}
    \caption{The outcome model $f$ ({\color{red!80!white} red}) is the sum of baseline $f_b$ ({\color{olive} yellow}) and total response $f_a$, which adds up single treatment responses of different magnitude (3 {\color{blue!80!white} blue} lines). \looseness-1}
    \label{fig:treatment_effect}
    \hack{\vspace{-1ex}}
\end{figure}

\textbf{Treatment Response Model.} We model the treatment responses as additive, similar to \citet{cheng2020patient}: 
\begin{align}
    \label{eq:treatment_effect}
    f_a(\tau; \mathbf{a}) &= \sum_{a_i=(t_i,m_i) \in \mathbf{a}} f_m(m_i) f_t(\tau; t_i),
\end{align}
where the time-dependent response function $f_t \sim \mathcal{GP}$ represents the response `shape' and is shared across patients, and $f_m(m_i)$ is a patient-specific linear function that scales the magnitude of the response for the given treatment dosage value $m_i \in \mathbb{R}$, as commonly done in the literature \citep{cheng2020patient,zhang2020errors}. 
This means the effects of nearby treatments simply sum up as shown in \cref{fig:treatment_effect}.
The scaling function $f_m$ has patient-specific parameters that have hierarchical priors, to share information between patients.
For the response shape function $f_t$ we use an \textsc{Se} kernel, as it provides sufficient performance for our use-case (\citet{cheng2020patient} also propose an extension with a latent force model \citep{alvarez2009latent}).
The model definition is detailed in \cref{sec:outcome-details}.

\section{Inference}
\label{sec:inference}

\paragraph{Policy Intervention.} To model the policy intervention, we first fit the joint model to the observational data using variational inference (see \cref{sec:learning}), which gives us the (observational) SCM $\mathcal{M} = (\mathbf{S}, p(\mathbf{N}))$. An intervention $[\tilde{\pi}]$ on the policy results in an interventional \textsc{Scm}: $\mathcal{M}_{\text{int}} = (\mathbf{S}_{\text{int}}, p(\mathbf{N}))$. Following Theorem 1, this intervention is equivalent to a stochastic intervention on the sequence of treatments, where the treatments now follow the new policy $\tilde{\pi}$. In practice, this is modeled by replacing the observed treatment intensity $\lambda^*_{\pi}(\cdot)$ with the new interventional intensity $\lambda^*_{\tilde{\pi}}(\cdot)$, while keeping the outcome model the same.%

\paragraph{Policy Counterfactual.} The policy counterfactual is calculated in exactly the same way as the policy intervention, except that the generic noise term $p(\mathbf{N})$ is replaced by individual-specific noise $p(\mathbf{N} \mid \mathcal{D}^{(v)})$ that has been inferred from the observations for that individual $(v)$: 
$\mathcal{M}_{\text{cf}} = (\mathbf{S}_{\text{int}}, p(\mathbf{N} \mid \mathcal{D}^{(v)}))$.
In our model, the noise consists of two components, treatment and outcome noise: $\mathbf{N} = \{\mathcal{E}_a, \mathcal{E}_o\}$. The outcome noise $\mathcal{E}_o$ is defined in \cref{eq:outcome} and its posterior is available in closed form from the \textsc{Gp}.
The treatment noise $\mathcal{E}_a$ comprises noise associated with the sampling process of the treatments. In practice, we estimate the noise and condition the counterfactual sampling of the treatments with these estimates using a novel variant of the algorithm proposed by \citet{noorbakhsh2021counterfactual}, which we have extended to history-dependent point processes (see \cref{sssec:cf_algo}). 

\paragraph{Identifiability of the policy counterfactual.} The counterfactual trajectories involve two components: treatments and outcomes. The counterfactual \emph{treatments} are identifiable, as our algorithm satisfies the monotonicity assumption \citep{pearl2009causality, noorbakhsh2021counterfactual}. On the other hand, the counterfactual \emph{outcomes} are not guaranteed to be identifiable, because this would require identifiability of the baseline and treatment response separately, which is not guaranteed as they both are modeled non-parametrically.
To address this, we use priors that encourage the baseline (slow-changing) and the response functions (fast-changing) to learn different aspects of the data, and we empirically show that in practice the response function $f_a(\tau)$ (\cref{eq:treatment_effect}) and hence the counterfactual trajectory can be identified with good accuracy. The counterfactual algorithm and identifiability are further detailed in \cref{sssec:cf-id}.

\section{Experiments}

In this section, we validate that our model can estimate interventional and counterfactual outcome trajectories under policy interventions. The empirical validation is composed of two parts. First, on a \emph{real-world observational dataset}, we show the proposed model can learn clinically meaningful treatment response curves and treatment intensities that can handle time-varying confounding. Second, we evaluate our model on two causal inference tasks: (i) the policy intervention and (ii) the policy counterfactual. To evaluate the performance on interventional and counterfactual predictions, we set up a \emph{realistic semi-synthetic simulation} scenario, as the true causal effects are unknown for real-world observational data sets. To ensure a realistic simulation scenario, we use a subset of learned models from the real-world study as the ground-truth data generators. The study is reproducible and our implementation in GPflow \citep{GPflow2020multioutput} can be found at \url{https://github.com/caglar-hizli/modeling-policy-interventions}.

\subsection{Learning Treatment Intensities and Response Curves on Real-World Observational Data}
\label{subsec:exp-real-world}

\begin{figure*}[t]
    \hspace{-1cm}\includegraphics[width=1.1\linewidth]{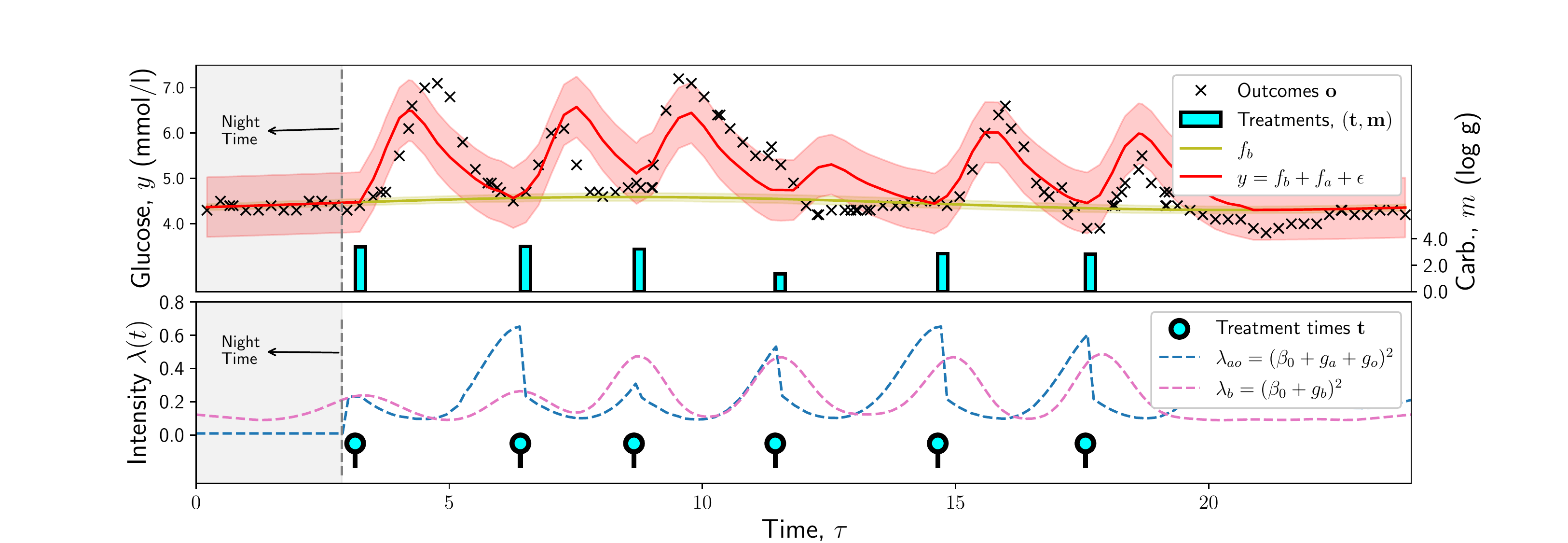}
    \caption{Estimated treatment--outcome model on the real-world meal--glucose data. \textbf{Top:} Glucose (outcome) measurements (black crosses), inferred baseline ({\color{olive}yellow}), and predicted glucose levels obtained by adding treatment responses, the baseline, and the noise ({\color{red}red}). The confidence intervals around the means refer to $\pm1$ standard deviation. We see that the increase in blood glucose is correctly predicted by the outcome model. \textbf{Bottom:} Treatment times and estimated intensities for the best model $\lambda_{ao}^*$ and the simple baseline $\lambda_b$. The baseline treatment model has a large meal intensity at times of the day when the patient usually eats. In addition, in the best treatment model, $\lambda_{ao}^*$, we see that the intensity of a new meal immediately after a previous meal decreases either directly through $g^*_a$ or indirectly through the increase in blood glucose represented by $g^*_o$. The intensities of the other models are shown in \cref{subsec:real_world_model}.}
    \label{fig:treatment_model_fit}
\end{figure*}

We first demonstrate that our model can learn clinically meaningful treatment intensities and response curves from a real-world observational dataset on physiological dynamics of blood glucose \citep{zhang2020errors,wyatt2021postprandial}. The experimental setup is further detailed in \cref{ssec:real-world}.

\textbf{Real-world dataset.} The dataset consists of treatment--outcome measurements of 14 non-diabetic patients over a 3-day period \citep{zhang2020errors}, corresponding to meals (treatments) and blood glucose measurements (outcomes). Patients record their meals in a meal diary and their blood glucose is measured by a monitoring device at regular time intervals. Meal ingredients are transformed into five nutrients: sugar, starch, protein, fiber, and fat. As meal dosage (mark), we preprocess nutrient covariates and use a single carbohydrate intake covariate equal to the sum of sugar and starch values, following \citet{ashrafi2021computational}.

\textbf{Causal Assumptions.} In the real-world dataset, the causal assumptions do \textbf{not} (exactly) hold and are not statistically testable. This is why (i) in this section, we perform only a \textit{statistical} analysis to show how our joint model works, and (ii) in \cref{ssec:semi}, we perform a \textit{causal} analysis on a semi-synthetic simulation study.

Assumption 2 states that an interventional meal has non-zero probability, which is likely to hold since we assume each patient follows their own diet (policy) and one patient's diet includes reasonable meal events that would have non-zero probability for the others. Assumption 3 does \emph{not} hold, since there can be unobserved processes, e.g., physical activity, which influence both meal times and daily glucose profiles.
However, the influence of confounding variables on blood glucose, apart from meals, is expected to be minor for non-diabetic individuals, as blood glucose is typically steady between meals \citep{ashrafi2021computational}.
Assumption 4 states that the policy effect is fully mediated through meals. 
For example, a policy that satisfies this is a dietary policy that dictates how the patient eats and in that way affects the blood glucose, but does not affect blood glucose in any other direct or indirect way. In \cref{ssec:semi}, we perform a causal analysis under such dietary policy interventions. As an example that violates this assumption, we could consider a more generic lifestyle intervention that includes aspects other than the diet, such as physical activity.

\textbf{Model.} The sequential meal--blood glucose data of a patient is modelled jointly, as detailed in Section \ref{sec:joint-model}. To show how the treatment intensity $\lambda^*(\tau)$ handles the time-varying confounding due to previous treatments and outcomes, we define five treatment intensities $\{\lambda_b, \lambda^*_{ba}, \lambda^*_{bo}, \lambda^*_{ao}, \lambda^*_{bao}\}$ for comparison.
In the most complex model $\lambda^*_{bao}$, the intensity of treatments depends on the baseline ($b$), previous treatments ($a$ for `action'), and previous outcomes ($o$); the simpler models have only some of these components as specified in their respective subscripts, see the functional forms in \cref{tab:testll}. 

\begin{table}
  \caption{Comparison of models for real-world data. The last column shows the test log-likelihood (\textsc{Tll}, higher is better; mean $\pm$ standard deviation) for the different models. Models are trained on the first 2 days of meal--glucose data. Day 3 is used to compute the \textsc{Tll}. %
  }
  \label{tab:testll}
  \centering
  \addtolength{\tabcolsep}{-3pt}
  \begin{tabular}{ccc}
    \toprule
    \textsc{Int\rlap{ensity}} & \textsc{Components} & \textsc{Tll} $\uparrow$ \\
    \midrule
    $\lambda_{b}$ & $\big(\beta_0 + g_{b}(\tau) \big)^2$ & $-13.3 \pm 0.5$ \\
    $\lambda_{ba}^*$ & $\big(\beta_0 + g_b(\tau) + g^*_a(\tau; \mathbf{a}) \big)^2$ & $-12.2 \pm 0.3$ \\
    $\lambda_{bo}^*$ & $\big(\beta_0 + g_b(\tau) + g^*_o(\tau; \mathbf{o}) \big)^2$ & $-11.0 \pm 0.3$ \\
    $\lambda_{ao}^*$ & $\big(\beta_0 + g^*_a(\tau; \mathbf{a}) + g^*_o(\tau; \mathbf{o}) \big)^2$ & $-10.7 \pm 0.4$ \\
    $\lambda_{bao}^*$ & $\big(\beta_0 + g_b(\tau) + g^*_a(\tau; \mathbf{a}) + g^*_o(\tau; \mathbf{o}) \big)^2$ & $-12.5 \pm 0.4$ \\
    \bottomrule
  \end{tabular}
\end{table}

\textbf{Results.} We show an example fit of the joint model for one patient in \cref{fig:treatment_model_fit}. The treatment- and outcome-dependent intensity $\lambda^*_{ao}$ (blue dashed line) correctly estimates that the probability of a meal event decreases (i) right after a new meal or (ii) with increasing blood glucose. This empirical finding suggests that functions $g^*_a$ and $g^*_o$ are useful in handling time-varying confounding due to past meal and glucose values.

We compare the different models in terms of the test log likelihood (\textsc{Tll}) for 14 patients in Table \ref{tab:testll}. We see that the simple history-independent intensity $\lambda_b(\tau)$ produces the lowest \textsc{Tll}. The treatment- and outcome-dependent intensity $\lambda_{ao}^*(\tau)$ produces the highest \textsc{Tll} mean, and slightly better results than the more complex intensity $\lambda_{bao}^*(\tau)$.

\colorlet{mylightblue}{blue!70!white}
\newcommand*{\citetablesup}[1]{{\color{mylightblue}$^{\textsc{(#1)}}$}}
\colorlet{mydarkblue}{blue!70!black}
\newcommand*{\ours}[1]{\textbf{\color{mydarkblue} \textsc{#1}}}
\newcommand*{\scm}[1]{$_{\text{#1}}$}
\begin{table*}[t]
  \footnotesize
  \caption{Causal tasks on semi-synthetic data: The \emph{observed} policy is $\pi_{[0,1d]} = \pi_A$, and we report mean squared error (\textsc{Mse}, lower is better) for observational $[\tilde{\pi}_{>1d} = \pi_A]$ and interventional $[\tilde{\pi}_{>1d} = {\color{red}\pi_B}]$ queries as well as for the policy counterfactual $[\tilde{\pi}_{\color{red}[0,1d]} = {\color{red}\pi_B}]$. The best results are \textbf{bolded}. Methods developed by us are highlighted in {\color{mydarkblue} \textbf{blue}}.}
  \label{tab:intervention}
  \label{tab:counterfactual}
  \centering
  \newcommand*{\nhpp}{\textsc{Nhpp}\citetablesup{L15}}
  \newcommand*{\gammap}{Gamma\citetablesup{H21}}
  \newcommand*{\paramSchulam}{Param.\citetablesup{S17}}
  \newcommand*{\paramHua}{Param.\citetablesup{H21}}
  \sisetup{detect-weight, separate-uncertainty=true, add-integer-zero=false}
  \begin{tabular}{lllS[table-format=1.2+-1.2]S[table-format=1.2+-1.2]lS[table-format=1.2+-1.2]}
    \toprule
    & & \multicolumn{3}{c}{\textsc{Query} (\textsc{Mse} $\downarrow$)} & \multicolumn{2}{c}{\textsc{Query} (\textsc{Mse} $\downarrow$)} \\
    \cmidrule(lr){3-5}
    \cmidrule(lr){6-7}
    \multirow{2}{*}{\parbox{1cm}{\textsc{Treatment Model}}} & \multirow{2}{*}{\parbox{1cm}{\textsc{Response Model}}} & 
    \multirow{2}{*}{$\mathcal{M}_{\textsc{Scm}}$} & {\textsc{Observational}} & 
    {\textsc{Interventional}} &
    \multirow{2}{*}{$\mathcal{M}_{\textsc{Scm}}$} & {\textsc{Counterfactual}}
    \\
     & & & {$[\tilde{\pi}_{>1d} = \pi_{A}]$} & {$[\tilde{\pi}_{>1d} = {\color{red}\pi_{B}}]$} & & {$[\tilde{\pi}_{\color{red}[0,1d]} = {\color{red}\pi_{B}}]$} \\
    \midrule
    \gammap      & \paramHua     & H21\scm{int}          &     .50+-.02 &     .50 \pm .02 & H21\scm{cf}          &     .49+-.02 \\
    \gammap      & \ours{Gp}     & \textsc{Ab1}\scm{int} &     .30+-.02 &     .30 \pm .01 & \textsc{Ab1}\scm{cf} &     .21+-.01 \\
    \nhpp        & \paramSchulam & \textsc{Ab2}\scm{int} &     .97+-.03 &    1.00 \pm .03 & \textsc{Ab2}\scm{cf} &     .59+-.02 \\
    \nhpp        & \ours{Gp}     & \textsc{Ab3}\scm{int} &     .34+-.02 &     .32 \pm .01 & \textsc{Ab3}\scm{cf} &     .23+-.01 \\
    \ours{Gp-pp} & \paramHua     & \textsc{Ab4}\scm{int} &     .46+-.01 &     .46 \pm .01 & \textsc{Ab4}\scm{cf} &     .45+-.02 \\
    \ours{Gp-pp} & \paramSchulam & \textsc{Ab5}\scm{int} &     .96+-.03 &    1.00 \pm .03 & \textsc{Ab5}\scm{cf} &     .56+-.02 \\
    \ours{Gp-pp} & \ours{Gp}     & \ours{Our}\scm{obs}   & \bf .19+-.01 &     .52 \pm .02 & \ours{Our}\scm{obs}  &     .82+-.03 \\
    \ours{Gp-pp} & \ours{Gp}     & \ours{Our}\scm{int}   & \bf .19+-.01 & \bf .20 \pm .01 & \ours{Our}\scm{int}  &     .71+-.03 \\
    \ours{Gp-pp} & \ours{Gp}     & \ours{Our}\scm{cf}    & \bf .19+-.01 & \bf .20 \pm .01 & \ours{Our}\scm{cf}   & \bf .14+-.01 \\
    \bottomrule
  \end{tabular}
\end{table*}

\subsection{Causal Tasks on Semi-Synthetic Data}
\label{ssec:semi}

In this section, we set up a semi-synthetic simulation study to demonstrate the ability of the proposed model to answer interventional and counterfactual queries resulting from an intervention on the treatment policy. Due to space constraints, the experimental setup is detailed in \cref{ssec:semi-synth}.

\textbf{Simulator.} We obtain the ground-truth simulator by fitting our joint model to the meal--glucose data set, as discussed in Section \ref{subsec:exp-real-world}. To be able to assess the sequential causal inference tasks, we have to include both sources of time-varying confounding. Therefore, we select as the ground-truth meal simulator the intensity that depends on both meal and glucose histories: $\lambda^{*}_{\pi} \sim \lambda^*_{ao}$.

\newcommand*{\piA}{{\color{blue}\pi_A}}
\newcommand*{\piB}{{\color{red}\pi_B}}
\textbf{Semi-synthetic dataset.} The ground-truth simulator is used to simulate samples from observational, interventional and counterfactual distributions of each patient $(v)$. Simulated patients are divided into (i) two policy groups $\{\piA, \piB\}$ representing different diets and (ii) three patient groups with distinct baseline and response functions, enabling individualization among patients. The meal simulators of two diets correspond to learned intensity functions of two real-world patients. The glucose simulators of three groups are learned from three real-world patients.

\newcommand*{\oneday}{\SI{1}{\day}}
For the observational data set, we simulate 1-day long meal--glucose time-series for 50 patients: $\mathcal{D} = \{\mathcal{D}^{(v)}\}_{v=1}^{50}$. For the interventional data set, we sample the next day of each patient under the specified policy intervention $[\tilde{\pi}_{>1d}]$. For the counterfactual data set, we condition on the observed data $\mathcal{D}^{(v)}$ for each patient and sample a hypothetical first day under a policy intervention $[\tilde{\pi}_{[0,1d]}]$, using the noise posteriors. We train all models on the observational dataset, and use interventional and counterfactual datasets for testing.

\textbf{Benchmarks.} First, we compare the proposed model with a parametric joint model \citep[\textsc{H21},][]{hua2021personalized}, which uses a renewal process with a parametric Gamma-shaped trigger function as the treatment model, and a linear mixed effects function as the outcome model. Next, we create ablations with different treatment and outcome model combinations, since other existing methods do not model treatments. With ablations having different treatment intensities, we measure the impact of having a flexible treatment intensity that can handle complex time-varying confounding scenarios. With ablations having different outcome models, we measure the impact of predicting the glucose trajectory well. The ablation \textsc{Ab1} combines the Gamma-based treatment intensity with our \textsc{Gp}-response model. Ablations \textsc{Ab2} and \textsc{Ab3} combine a history-independent treatment model \citep[\textsc{Nhpp},][\citetablesup{L15}]{lloyd2015variational} with (i) a parametric response model proposed by \citet[\citetablesup{S17}]{schulam2017reliable} and (ii) our \textsc{Gp}-response model. Ablations \textsc{Ab4} and \textsc{Ab5} combine our flexible treatment intensity (\textsc{Gp-pp}) with two parametric response models proposed by \citet[\citetablesup{H21}]{hua2021personalized} and \citet[\citetablesup{S17}]{schulam2017reliable}.

The columns $\mathcal{M}_{\textsc{Scm}}$ in \cref{tab:intervention} specify the combination of (i) a joint model identifier $\mathcal{M}$, e.g., \textsc{Our}, and (ii) the target distribution of the \textsc{Scm} stated as a subscript: $(\cdot)$\scm{obs} for observational, $(\cdot)$\scm{int} for interventional, and $(\cdot)$\scm{cf} for counterfactual. Similar to \citet{schulam2017reliable}, we use our own observational model (\textsc{Our}$_\text{obs}$) as another baseline in the interventional task. In the counterfactual task, we use both our observational (\textsc{Our}$_\text{obs}$) and interventional model (\textsc{Our}$_\text{int}$) as baselines. Our counterfactual model is denoted by \textsc{Our}$_\text{cf}$. To give our baseline and ablation models the most strength, we use their interventional version, ($\cdot$)\scm{int}, for estimating the interventional query and their counterfactual version, $(\cdot)$\scm{cf}, for estimating the counterfactual query.

\textbf{Metric.} We report the mean squared error (\textsc{Mse}) between ground-truth and estimated glucose trajectories under the interventional and counterfactual distributions, over all patients and all measurement times. To obtain estimated trajectories comparable to the ground-truth trajectories, we fix the noise variables of the point process (meal) sampling algorithm between the ground-truth simulator and each estimated model. Each experiment is repeated 10 times. We report mean and standard deviation of the \textsc{Mse} metric.

\textbf{Policy Intervention.} We estimate the interventional query: What will happen if patient $(v)$ with the diet $\pi^{(v)} = \piA$ will continue for the next day in the future, adopting another patient's diet $\piB$? Following \cref{eq:target_intervention}, the causal query for this task is formalized as the interventional query: \[ P(\mathbf{Y}^{(v)}_{>1d}[\tilde{\pi}_{>1d} = \piB] \mid \pi_{[0,1d]}=\piA, \mathcal{H}^{(v)}_{\leq 1d}). \]
\textsc{Mse} results are shown in Table \ref{tab:intervention}. We see that the interventional model \textsc{Our}$_{\text{int}}$ is able to estimate observational and interventional trajectories with low error when the intervention policy is (i) the same as the observed policy $[\tilde{\pi}_{>1d} = \piA]$ and (ii) different from the observed policy $[\tilde{\pi}_{>1d} = \piB]$, while the observational model \textsc{Our}$_{\text{obs}}$ fails in the latter case. This is because \textsc{Our}$_{\text{obs}}$ can only answer a statistical query of the following form: Observing the history of patient $(v)$, how will the patient state progress for the next day, $P(\mathbf{Y}^{(v)}_{>1d} \mid \pi_{[0,1d]}=\piA, \mathcal{H}^{(v)}_{\leq 1d})$? Furthermore, our empirical findings suggest that simplistic models with parametric glucose responses (Param.\citetablesup{S17,H21}) perform badly, even when they target the interventional distribution.

\textbf{Policy Counterfactual.} We estimate the counterfactual query: What \emph{would have} happened if patient $(v)$ with the diet $\pi^{(v)} = \piA$ had followed another patient's diet $\piB$ in the observation period $\color{red}[0,1d]$? Following \cref{eq:target_counterfactual}, the causal query for this task is formalized as the counterfactual \[ P(\mathbf{Y}^{(v)}_{\leq 1d}[\tilde{\pi}_{\color{red}[0,1d]} = \piB] \mid \pi_{[0,1d]}=\piA, \mathcal{H}^{(v)}_{\leq 1d}) . \] 
\textsc{Mse} results are shown in the last column of Table \ref{tab:counterfactual}. We see that the interventional model \textsc{Our}$_{\text{int}}$ fails to estimate counterfactual trajectories, as it does not take into account the noise posterior of each patient. On the other hand, the counterfactual model \textsc{Our}$_{\text{cf}}$ accurately samples counterfactual trajectories and has the smallest error of the methods considered.

\section{Discussion}

To study what happens if the (possibly implicit) treatment policy of one individual (hospital, unit, country, \dots) is or had been adopted by another individual, we proposed a model that jointly considers sequences of treatments and outcomes of each individual. Theoretically, we showed that an intervention on a treatment policy is equivalent to a sequence of stochastic interventions on treatments, whose potential outcomes can be estimated from observational data with the joint model. In a real-world scenario, we demonstrated that our non-parametric model can be learned from observational sequences of treatments and outcomes, and it can handle time-varying confounding in continuous time. In a semi-synthetic experiment, we demonstrated that the joint model can answer causal queries about the interventional and counterfactual distributions of the outcome after an intervention on the treatment policy.

\subsection{Limitations and Future Work}

\paragraph{Modeling Limitations.} The main limitation of the proposed outcome model is scalability, as it is not trivial to use inducing point approximations for the current `time-marked' treatment response function. This could be improved in future work, e.g., by clever inducing point selections. Furthermore, we considered treatment responses to be additive, which does not allow for modeling interactions of nearby treatments. This could be improved by integrating more sophisticated kernels into the treatment response function.

The thinning algorithm for sampling the point process is sequential and known to be slow, which can raise a challenge in using our method for downstream tasks such as optimizing the policy by a model-based reinforcement learning algorithm. It could be possible to adapt an inverse sampling method with $\mathcal{O}(1)$ time complexity \citep{rasmussen2011temporal}, but this would require a completely different treatment model definition as in \citet{schur2020fast}, whose applicability would need to be studied separately.

\paragraph{Causal Assumptions.}
For real-world observational datasets, our causal assumptions do not necessarily hold, in which case our method should not be used for \emph{causal} analysis (the model may still be useful for statistical analysis).
For example, the violation of the \textsc{Nuc} assumption would result in an error in the causal effect estimates due to the unblocked back-door paths, whose magnitude depends on the size of the violation.
An interesting future direction is to extend existing sensitivity analysis methods \citep{robins2000sensitivity} in order to estimate the size of the error in causal effects of continuous-time treatment sequences.

\section*{Acknowledgements}

The authors would like to thank \c{C}a\u{g}atay Y{\i}ld{\i}z for fruitful discussions. This work was supported by the Academy of Finland (Flagship programme: Finnish Center for Artificial Intelligence FCAI, and grants 336033, 352986) and EU (H2020 grant 101016775 and NextGenerationEU).

\bibliography{ms}
\bibliographystyle{ms}

\newpage
\appendix
\onecolumn

\section{Further Related Work}
\label{sec:related_work}

\begin{table}
  \caption{Taxonomy of the related work, with respect to treatment model type, outcome model type, continuity of time-steps and their target distributions: observational (Obs.), interventional (Int.) and counterfactual (Cf.).}
  \label{tab:related_work}
  \centering
  \begin{tabular}{lllllll}
    \toprule
    & & & \multicolumn{3}{c}{Target Distribution} & \\
    \cmidrule(r){4-6}
    Treatment & Response & Continuous- &  &  &  &  \\
    Model & Model & time & Obs. & Int. & Cf. & Reference \\
    \midrule
    {\color{red} \xmark} & Parametric & {\color{teal} \cmark} & {\color{teal} \cmark} & {\color{teal} \cmark} & {\color{red} \xmark} & \citet{schulam2017reliable}  \\
    {\color{red} \xmark} & \textsc{Rnn} & {\color{red} \xmark} & {\color{teal} \cmark} & {\color{teal} \cmark} & {\color{red} \xmark} & \citet{lim2018forecasting} \\
    {\color{red} \xmark} & \textsc{Rnn} & {\color{red} \xmark} & {\color{teal} \cmark} & {\color{teal} \cmark} & {\color{red} \xmark} & \citet{bica2020estimating} \\
    {\color{red} \xmark} & Neural \textsc{Cde} & {\color{teal} \cmark} & {\color{teal} \cmark} & {\color{teal} \cmark} & {\color{red} \xmark} & \citet{seedat2022continuous}  \\
    {\color{red} \xmark} & \textsc{Nhpp} & {\color{teal} \cmark} & {\color{teal} \cmark} & {\color{teal} \cmark} & {\color{teal} \cmark} & \citet{noorbakhsh2021counterfactual} \\
    Gamma & Parametric & {\color{teal} \cmark} & {\color{teal} \cmark} & {\color{red} \xmark} & {\color{red} \xmark} & \citet{hua2021personalized} \\
    \textsc{Gp-pp} & Non-parametric & {\color{teal} \cmark} & {\color{teal} \cmark} & {\color{teal} \cmark} & {\color{teal} \cmark} & Our Work \\
    \bottomrule
  \end{tabular}
\end{table}

\paragraph{Treatment Responses for Sequential Treatment--Outcome Data.} 

Estimating sequential treatment effects has been studied by a large number of works \citep{schulam2017reliable, xu2016bayesian, soleimani2017treatment, lim2018forecasting, bica2020estimating, zhang2020errors, cheng2020patient, seedat2022continuous}. Recurrent marginal structural networks \citep[\textsc{Rmsn},][]{lim2018forecasting} extend marginal structural models \citep[\textsc{Msm},][]{robins2000marginal} using a recurrent neural network (\textsc{Rnn}) architecture to estimate propensity weights and treatment effects over time. The counterfactual recurrent network \citep[\textsc{Crn},][]{bica2020estimating} also uses an \textsc{Rnn} architecture, but extends the previous work by using domain adversarial training instead of estimating the propensity weights explicitly. Both works focus on a \emph{discrete-time} sequential treatment--outcome setup.\looseness-1

For \emph{continuous-time} treatments and outcomes, a number of works have proposed generative models over the outcome trajectory \citep{schulam2015framework, xu2016bayesian, soleimani2017treatment, zhang2020errors, cheng2020patient}, in the direction of the g-computation method \citep{robins1986new, robins1987addendum}. These works model the outcome trajectory as a sum of (i) a counterfactual baseline that captures the no-treatment case and (ii) an additive treatment response function. Most of these methods use a Bayesian non-parametric baseline and a parametric treatment response \citep{schulam2015framework, xu2016bayesian, soleimani2017treatment, zhang2020errors}. In our work, we combine a non-parametric baseline with a more flexible, non-parametric treatment response function, similar to \citet{cheng2020patient}. In addition to these continuous-time methods, \citet{seedat2022continuous} estimate continuous-time treatment responses, by combining (i) neural controlled differential equations (Neural \textsc{Cde}) to model the latent outcome trajectory, and (ii) domain adversarial training to deal with the time-varying confounding. However, none of these methods model treatments, so they require a fixed sequence of treatment interventions as \emph{input} at test time. Besides, they do not consider counterfactuals of observed time-series.

\paragraph{Treatment (Event) Models in Continuous Time.} Continuous-time treatments can be modeled by temporal point processes (\textsc{Tpp}), which have been investigated from a causal inference perspective by a large body of work \citep{lok2008statistical, schulam2017reliable, gao2021causal, aalen2020time, ryalen2020causal, hua2021personalized}. \citet{lok2008statistical} estimates the causal effect of interventions on continuous-time treatments, by extending structural nested models \citep{robins1992estimation} using a martingale approach. The methods proposed by \citet{aalen2020time} and \citet{ryalen2020causal} consider interventions affecting continuous-time events on a survival outcome, where the \textsc{Tpp} terminates after a single survival/death event. \citet{gao2021causal} investigate the average treatment effect between pairs of event variables of a multivariate point process. Similar to our work, \citet{schulam2017reliable} estimate interventional outcome (mark) trajectories affected by continuous-time treatments; however, they assume a fixed sequence of treatment interventions as input. Instead, we model continuous-time treatments jointly with the outcome trajectories. This enables our model to make predictions under more realistic, alternative scenarios.

From a modeling perspective, the model proposed in \citet{hua2021personalized} is the closest to our work, as they propose a Bayesian joint model for the sequences of treatments and outcomes. However, their goal is to find an optimal treatment policy for continuous-time treatments, while our goal is to estimate interventional and counterfactual outcome trajectories resulting from policy interventions. Besides, they use parametric models for both treatment and outcome models, heavily-tailored for their kidney transplantation application, which do not generalize trivially to other problem setups.

\paragraph{Counterfactual Treatments.} Counterfactual reasoning has recently raised interest in the machine learning literature, where it has been used for evaluating and explaining model predictions \citep{oberst2019counterfactual, tsirtsis2020decisions, tsirtsis2021counterfactual, abid2022meaningfully, aalen2020time, ryalen2020causal, noorbakhsh2021counterfactual}. Most of these work focus on discrete-time treatment--outcome setups \citep{oberst2019counterfactual, tsirtsis2020decisions, tsirtsis2021counterfactual, abid2022meaningfully}. \citet{aalen2020time} and \citet{ryalen2020causal} consider counterfactuals of a continuous-time survival outcome as discussed in the previous paragraph. From the perspective of counterfactual treatments, the closest to our work is counterfactual \textsc{Tpp}s proposed in \citet{noorbakhsh2021counterfactual}. However, \citet{noorbakhsh2021counterfactual} consider continuous-time event occurrences as target outcomes, while our main target is an outcome (mark) trajectory that is causally affected by a treatment policy intervention through continuous-time treatment events. Besides, the counterfactual sampling algorithm proposed in \citet{noorbakhsh2021counterfactual} is limited to non-homogeneous Poisson processes (\textsc{Nhpp}) (independent events) or their Hawkes process variants since they choose Lewis' thinning algorithm \citep{lewis1979simulation} as the generative process. This limits the capacity of the algorithm to handle time-varying confounding in a sequential treatment--outcome setup. Therefore, we extend their counterfactual sampling algorithm to history-dependent point processes, by choosing Ogata's thinning algorithm \citep{ogata1981lewis} as the generative process.

A taxonomy of the related work is presented in Table \ref{tab:related_work}, which classifies the closest of the aforementioned works in terms of the treatment model type, the outcome model type, the continuity of time-steps and their target distributions: observational (Obs.), interventional (Int.) and counterfactual (Cf.).

\section{Causal Assumptions}
\label{sec:causal-assump}

Assume a structural causal model (\textsc{Scm}) with an intervention variable $A \in \mathcal{A}$, outcome variable $Y \in \mathbb{R}$ and all other variables $Z \in \mathcal{Z}$.

\subsection{Intervention Types}

In practice, an intervention on a variable $A$ removes the structural equation $A := f_a(\DAGParent(A), U_a)$. Intervention types differ in terms of the way they set the intervention value $do(A)$.

\textbf{Atomic intervention.} Sets $A$ to a static value $\tilde{a}$. Under identifiability conditions, the effect of the intervention on the outcome variable is as follows:
    $$ p(Y \mid do(A=\tilde{a})) = \sum_{z} p(Y \mid A=\tilde{a}, Z=z) p(Z=z). $$
\textbf{Conditional intervention.} Sets $A$ by a deterministic function $\tilde{h}: \mathcal{Z} \rightarrow \mathcal{A}$. Under identifiability conditions, the effect of the intervention on the outcome variable is as follows:
    \begin{align*}
        p(Y \mid do(A=\tilde{h}(z))) &= \sum_z p(Y \mid do(A=\tilde{h}(z)), Z=z) p(Z=z \mid do(A=\tilde{h}(z))) \\
        &= \sum_z p(Y \mid a, z)|_{a=\tilde{h}(z)} p(z).
    \end{align*}
\textbf{Stochastic (conditional) intervention.} Sets $A$ by a stochastic relationship $\tilde{p}(A \mid Z)$. Given $Z$, the intervention $do(A=\tilde{a})$ occurs with probability $\tilde{p}(A \mid Z)$. Therefore, the effect of an atomic intervention $do(A=\tilde{a})$ is averaged over all possible $\tilde{a} \sim \tilde{p}(A | Z)$:
    \begin{align*}
        p(Y \mid &do(A \sim \tilde{p}(A \mid Z))) \\ 
        &= \sum_{\tilde{a}} \sum_{z} p(Y \mid do(A=\tilde{a}), z=z) \tilde{p}(A=\tilde{a} \mid Z=z) p(Z=z) \\
        &= \sum_{\tilde{a}} \sum_{z} p(Y \mid \tilde{a}, z) \tilde{p}(\tilde{a} \mid z) p(z).
    \end{align*}

\subsection{Formal Definitions of Causal Assumptions}

Consistency, no-unmeasured confounding (NUC) and positivity are standard causal assumptions required for the identifiability of a static causal effect \citep{pearl2009causality, hernan2010causal, schulam2017reliable, bica2020estimating}.

\textbf{Assumption 1: Consistency.} The potential outcome $Y[a]$ under an action $a \in \mathcal{A}$ is consistent with its factual outcome, provided that the outcome $Y$ is observed under the action $A=a$: $P(Y[a] \mid A=a) = P(Y \mid A=a)$.

\textbf{Assumption S1: Positivity (Overlap).} For a well-defined causal effect, the intervention must be have a non-zero probability: $P(A=a \mid Z) > 0$.

\textbf{Assumption S2: No-Unmeasured Confounding (NUC).} The potential outcome $Y[a]$ is independent of the action $A$ conditioned on $Z$, if the action assignment is done at random given $Z$: $Y[a] \independent A \mid Z$.

For continuous-time interventions, we require a continuous-time version of the positivity assumption (Assumption 2). In addition, \citet{schulam2017reliable} propose the continuous-time NUC assumption (Assumption 3), for the continuous-time sequential treatment--outcome setup. Furthermore, we assume that a policy intervention affects the outcome trajectory only through the sequence of treatments, i.e., there is no direct effect from the policy variable to the outcomes (Assumption 4).

\textbf{Assumption 2: Continuous-Time Positivity.} The conditional treatment intensity $\lambda_{\pi}^*(t, m)$ is non-zero for all intervention times $t$ and all intervention marks $m$, given any intervention history $\mathcal{H}_{<t}$: $\lambda_{\pi}^*(t, m) > 0, \forall t \in \mathbb{R}_{\geq 0}, \forall m \in \mathbb{R}$.

\textbf{Assumption 3: Continuous-time NUC.} Let sequential treatments $\mathbf{a}$ occur at a discrete set of time points on a given interval, characterized by the conditional intensity function $\lambda_{\pi}^*(t, m)$. The conditional treatment intensity $\lambda_{\pi}^*(t, m)$ of a treatment $a = (t, m)$, is independent of the potential outcome trajectory $\mathbf{Y}_{>t}[\tilde{\mathbf{a}}]$, conditioned on the past history $\mathcal{H}_{< t}$, $\forall t \in \mathbb{R}_{\geq 0}$\footnote{In fact, the history $\mathcal{H}_{< t}$ can be extended to include an outcome variable that occurs at time $t$, if it exists: $\mathcal{H}_{\leq t} \cup \{o_t\}$, assuming instantaneous effects from outcomes to treatments: $o \rightarrow a$. However, it would clutter the notation and in practice, the probability of an outcome $o=(t,y)$ to occur at time $t$ is $0$. Therefore, assuming no instantaneous effects between variables can be made without any practical effects.}.

\textbf{Assumption 4: Fully-mediated policy effect.} Conditioned on the past history $\mathcal{H}_{<\tau}$, the causal effect of the policy $\pi_{>\tau}$ on the outcome trajectory $\mathbf{Y}_{>\tau}$ is fully mediated through sequential treatments $\mathbf{a}_{>\tau}$: 
$$P(Y_{>\tau}[\tilde{\pi}_{>\tau}, \tilde{\mathbf{a}}_{>\tau}] \mid \mathcal{H}_{\leq \tau}) = P(Y_{>\tau}[\tilde{\mathbf{a}}_{>\tau}] \mid \mathcal{H}_{\leq \tau}).$$

\section{Causal Identifiability}
\label{sec:causal-iden}

We evaluate the potential outcome trajectory $\mathbf{Y}_{>T}[\tilde{\pi}_{>T}]$ at a discrete set of ordered query points $\mathbf{q}=\{q_1, ..., q_m: q_i > T, \forall i \in 1,...,m\}$: $\mathbf{Y_q}[\tilde{\pi}_{>T}] = \{y(q_k)[\tilde{\pi}_{>T}]: q_k \in \mathbf{q}\}$, similar to \citet{schulam2017reliable}. We denote the outcome value $y(q_k)$ at time $q_k$ by $Y_{q_k} = y(q_k)$. Under the \textsc{Nuc} Assumption (Assumption S2), the assignment of the target policy label $\tilde{\pi}$ is independent of the potential outcome query $\mathbf{Y_q}[\tilde{\pi}_{>T}]$: 
\begin{align}
\label{eq:po_query}
P(\mathbf{Y_q}[\tilde{\pi}_{>T}] \mid \mathcal{H}_{\leq T}) = P(\mathbf{Y_q}[\tilde{\pi}_{>T}] \mid \tilde{\pi}_{>T}, \mathcal{H}_{\leq T})
\end{align}

The causal graph $\mathcal{G}$ suggests that future sequential treatments $\mathbf{a}_{>T}$ act as mediators for the causal effect of the policy intervention $[\tilde{\pi}_{>T}]$ on the outcome query $\mathbf{Y_q}$. To consider this mediation effect, we include future sequential treatments $\mathbf{a}_{>T}$ to \cref{eq:po_query}: 
\begin{align}
\label{eq:po_query_w_action}
P(\mathbf{Y_q}[\tilde{\pi}_{>T}] \mid \tilde{\pi}_{>T}, \mathcal{H}_{\leq T}) = \sum_{\mathbf{a}_{>T}} P(\mathbf{Y_q}[\tilde{\pi}_{>T}], \mathbf{a}_{>T} \mid \tilde{\pi}_{>T}, \mathcal{H}_{\leq T}).
\end{align}
Now, let us consider an intermediate outcome $Y_{q_k}$ at query time $q_k$. The intermediate outcome $Y_{q_k}$ acts as a common cause for some future actions $\mathbf{a}_{[q_k,q_{k+1})}$ and a future outcome $Y_{q_{k+1}}$. However, it is not possible to adjust for $Y_{q_k}$ as it simultaneously acts as a mediator for a different set of causal paths from previous actions $\mathbf{a}_{<q_k}$ to a future outcome $Y_{q_{k+1}}$. This phenomenon is known as time-varying confounding \citep{robins1986new, pearl1995probabilistic}. To eliminate the time-varying confounding effect, we factorize the counterfactual query $\mathbf{Y_q}[\tilde{\pi}_{>T}]$ and sequential actions $\mathbf{a}_{>T}$ in \cref{eq:po_query_w_action} in time-order:
\begin{align}
    \label{eq:g_estimation1}
    P(\mathbf{Y_q}[\tilde{\pi}_{>T}] \mid \mathcal{H}_{\leq T})
    = \sum_{\mathbf{a}} \prod_{k=1}^m \Big( &P(Y_{q_k}[\tilde{\pi}_{>T}] \mid \mathbf{a}_{<q_k}, \mathbf{Y}_{\leq q_{k-1}}[\tilde{\pi}_{>T}], \tilde{\pi}_{>T}, \mathcal{H}_{\leq T}) \nonumber \\
    &P(\mathbf{a}_{[q_{k-1}, q_k)} \mid \mathbf{a}_{<q_{k-1}}, \mathbf{Y}_{\leq q_{k-1}}[\tilde{\pi}_{>T}], \tilde{\pi}_{>T}, \mathcal{H}_{\leq T}) \Big),
\end{align}
similar to the g-estimation formula \citep{robins1986new}. By consistency assumption (Assumption 1), the potential outcome $\mathbf{Y}_{\leq q_{k-1}}[\tilde{\pi}_{>T}]$ is equal to the factual outcome when we observe (condition on) the policy specified by $\tilde{\pi}_{>T}$: 
\begin{align}
    \label{eq:g_estimation2}
    P(\mathbf{Y_q}[\tilde{\pi}_{>T}] \mid \mathcal{H}_{\leq T})
    = \sum_{\mathbf{a}} \prod_{k=1}^m \Big(P(Y_{q_k}[\tilde{\pi}_{>T}] \mid \tilde{\pi}_{>T}, \underbrace{\mathbf{a}_{<q_k}, \mathbf{Y}_{\leq q_{k-1}}, \mathcal{H}_{\leq T}}_{\mathcal{H}_{q_{k-1}}})
    P(\mathbf{a}_{[q_{k-1}, q_k)} \mid \tilde{\pi}_{>T}, \underbrace{\mathbf{a}_{<q_{k-1}}, \mathbf{Y}_{\leq q_{k-1}}, \mathcal{H}_{\leq T}}_{\mathcal{H}_{q_{k-1}}}) \Big).
\end{align}

For both the outcome variable $Y_{q_k}$ and actions $\mathbf{a}_{[q_{k-1}, q_k)}$, the set of variables $\{\mathbf{a}_{<q_{k-1}}, \mathbf{Y}_{\leq q_{k-1}}, \mathcal{H}_{\leq T}\}$ is a valid history until the query point $q_{k-1}$: $\mathcal{H}_{q_{k-1}} = \mathcal{H}_{\leq q_{k-1}} \setminus \{a_{q_{k-1}}\} = \{\mathbf{a}_{<q_{k-1}}, \mathbf{Y}_{\leq q_{k-1}}, \mathcal{H}_{\leq T}\}$. Hence, \cref{eq:g_estimation2} can be simplified as follows:
\begin{align}
\label{eq:factors}
    P(\mathbf{Y_q}[\tilde{\pi}_{>T}] \mid \mathcal{H}_{\leq T})
    = \sum_{\mathbf{a}} \prod_{k=1}^m \Big( \underbrace{P(Y_{q_k}[\tilde{\pi}_{>T}] \mid \mathbf{a}_{[q_{k-1}, q_k)}, \tilde{\pi}_{>T}, \mathcal{H}_{q_{k-1}})}_{\text{Outcome Factor}}
    \underbrace{P(\mathbf{a}_{[q_{k-1}, q_k)} \mid \tilde{\pi}_{>T}, \mathcal{H}_{q_{k-1}})}_{\text{Treatment Factor}} \Big).
\end{align}

Under Assumptions $\{1, S2\}$, the treatment factor term in \cref{eq:factors} is equivalent to the potential outcome $\mathbf{a}_{[q_{k-1}, q_k)}[\tilde{\pi}_{>T}]$:
\begin{align}
\label{eq:treatment_factor_idn}
    P(\mathbf{a}_{[q_{k-1}, q_k)} \mid \tilde{\pi}_{>T}, \mathcal{H}_{q_{k-1}}) 
    &= P(\mathbf{a}_{[q_{k-1}, q_k)}[\tilde{\pi}_{>T}] \mid \tilde{\pi}_{>T}, \mathcal{H}_{q_{k-1}}) &\text{A.1} \nonumber \\
    &= P(\mathbf{a}_{[q_{k-1}, q_k)}[\tilde{\pi}_{>T}] \mid \mathcal{H}_{q_{k-1}}). &\text{A.S2}
\end{align}

Conditioned on the past history $\mathcal{H}_{q_{k-1}}$, the causal effect of the policy intervention $[\tilde{\pi}_{>T}]$ on the outcome value $Y_{q_k}$ is fully mediated through sequential treatments $\mathbf{a}_{[q_{k-1}, q_k)}$. Using Assumptions $\{1,2,3,4\}$, the outcome factor term in \cref{eq:factors} is equivalent to the potential outcome $Y_{q_k}[\tilde{\mathbf{a}}_{[q_{k-1}, q_k)}]$:
\begin{align}
\label{eq:outcome_factor_idn}
    \qquad\qquad P(Y_{q_k}[\tilde{\pi}_{>T}] \mid& \mathbf{a}_{[q_{k-1}, q_k)}, \tilde{\pi}_{>T}, \mathcal{H}_{q_{k-1}}) &\text{ } \notag \\ 
    &=P(Y_{q_k} \mid \mathbf{a}_{[q_{k-1}, q_k)}, \tilde{\pi}_{>T}, \mathcal{H}_{q_{k-1}}) &\text{A.1} \notag \\
    &=P(Y_{q_k} \mid \mathbf{a}_{[q_{k-1}, q_k)}, \mathcal{H}_{q_{k-1}}) &\text{A.4} \notag \\
    &= P(Y_{q_k}[\mathbf{a}_{[q_{k-1}, q_k)}] \mid \mathbf{a}_{[q_{k-1}, q_k)}, \mathcal{H}_{q_{k-1}}) &\text{A.1} \notag \\
    &= P(Y_{q_k}[\mathbf{a}_{[q_{k-1}, q_k)}] \mid \mathcal{H}_{q_{k-1}}) . &\text{A.3}
\end{align}

Plugging the potential outcome representations of the treatment factor (\cref{eq:treatment_factor_idn}) and of the outcome factor (\cref{eq:outcome_factor_idn}) into \cref{eq:factors}, the potential query can be considered as a sequence of stochastic (conditional) interventions on treatment variables:
\begin{align}
    P(\mathbf{Y_q}[\tilde{\pi}_{>T}] \mid \mathcal{H}_{\leq T})
    = \sum_{\tilde{\mathbf{a}}} \prod_{k=1}^m P(Y_{q_k}[\tilde{\mathbf{a}}_{[q_{k-1}, q_k)}] \mid \mathcal{H}_{q_{k-1}}) P(\tilde{\mathbf{a}}_{[q_{k-1}, q_k)}[\tilde{\pi}_{>T}] \mid \mathcal{H}_{q_{k-1}}).
\end{align}
Under Assumptions $\{1,2,3\}$, a sequence of interventions on treatments is identified \citep{schulam2017reliable,seedat2022continuous}:
\begin{align}
    \label{eq:target_statistical}
    P(\mathbf{Y_q}[\tilde{\pi}_{>T}] \mid \mathcal{H}_{\leq T}) = \sum_{\tilde{\mathbf{a}}} \prod_{k=1}^m P(Y_{q_k} \mid \tilde{\mathbf{a}}_{[q_{k-1}, q_k)}, \mathcal{H}_{q_{k-1}}) P(\tilde{\mathbf{a}}_{[q_{k-1}, q_k)} \mid \tilde{\pi}_{>T}, \mathcal{H}_{q_{k-1}}).
\end{align}

\clearpage
\section{Nonparametric Modeling approaches for \textsc{Tpp}s}

\begin{figure}[h]
    \centering
    \begin{subfigure}{.45\textwidth}
        \centering
        \includegraphics[width=0.95\linewidth]{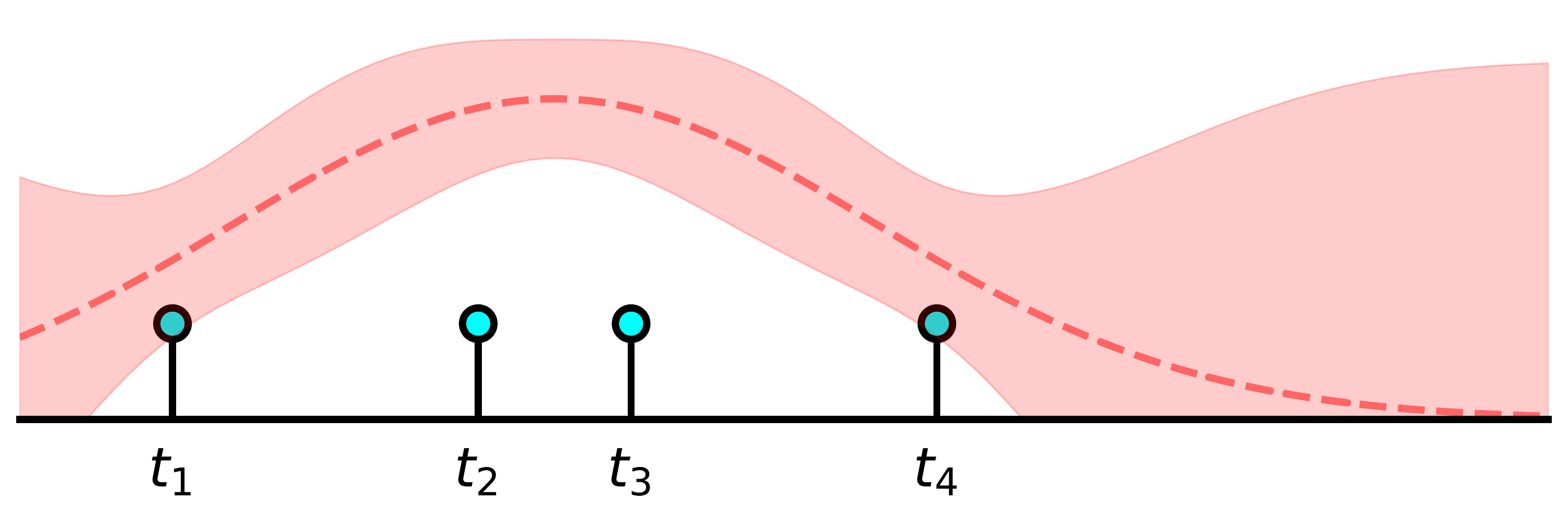}
        \caption{Latent-state point processes.}
        \label{fig:latentpp}
     \end{subfigure}
     \begin{subfigure}{.45\textwidth}
        \centering
        \includegraphics[width=0.95\textwidth]{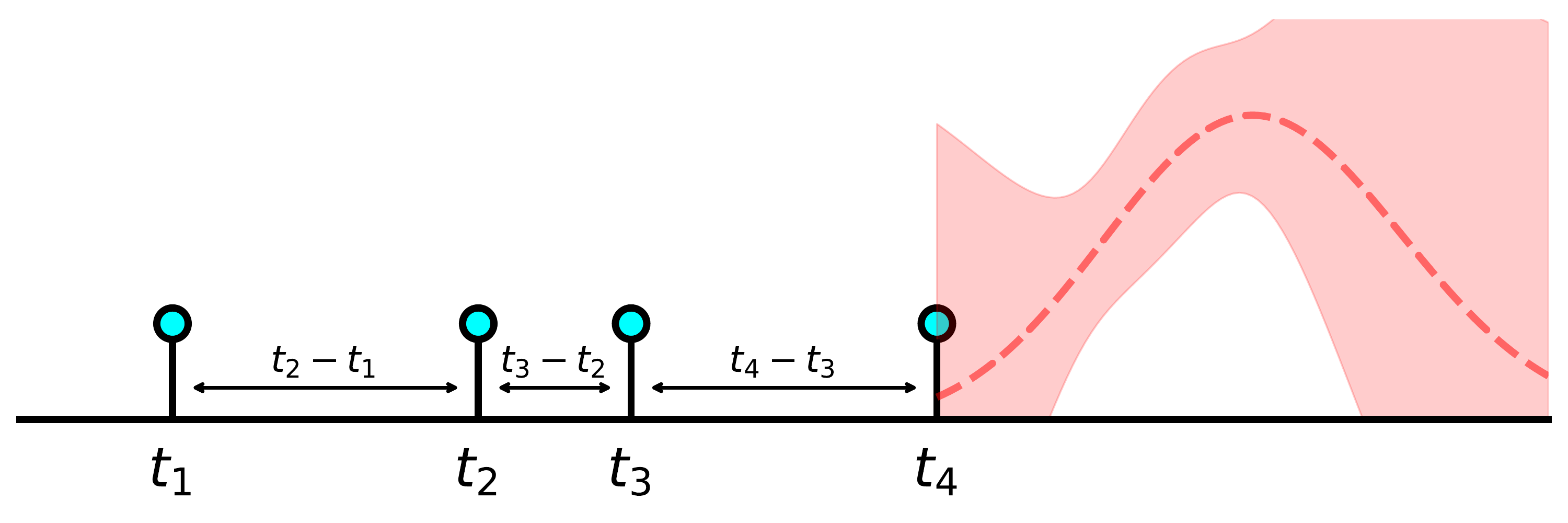}
        \caption{Regressive point processes.}
        \label{fig:regressivepp}
     \end{subfigure}
    \caption{\textbf{(a)} An example of the conditional intensity function $\lambda$ for a latent-state point process, which takes absolute times as input. \textbf{(b)} An example of the conditional intensity function $\lambda$ for a regressive point process, which takes relative times as input.}
    \label{fig:tpp}
\end{figure}

\textsc{Tpp}s can be modeled by two nonparametric approaches: (i) latent-state point processes \cite{moller1998log,adams2009tractable,lloyd2015variational} and (ii) regressive point processes \cite{hawkes2018hawkes,liu2019nonparametric}. The difference between the two approaches is shown in Figure \ref{fig:tpp}.

\subsection{Latent-state point processes}

Latent-state point processes, e.g. a log/sigmoidal Gaussian Cox process \citep{moller1998log,adams2009tractable} and a variational Bayes point process \citep[\textsc{Vbpp},][]{lloyd2015variational}, model the temporal dependence of the intensity indirectly, in the form of smoothness assumptions encoded by a latent (\textsc{Gp}) prior \citep{williams2006gaussian}. For example, the \textsc{Vbpp} model defines a latent \textsc{Gp} prior $f \sim \mathcal{GP}$ with a squared-exponential (\textsc{Se}) kernel over a period $[0,T]$. The latent function $f$ is passed through a square transformation to obtain the conditional intensity function $\lambda(\tau)$. 
The latent function $f: \mathbb{R}_{\geq 0} \rightarrow \mathbb{R}$ takes an absolute time $\tau \in [0,T]$ as input and outputs the latent rate for the same time point. The choice of the \textsc{Se} kernel assumes a smooth latent function, which implicitly enforces some temporal dependence of nearby intensity values. For example, the intensity values for points $\{t_i\}_{i=1}^4$ in \cref{fig:latentpp} is given by:
\begin{align*}
\lambda^*
\left(
\begin{bmatrix}
t_1 \\
t_2 \\
t_3 \\
t_4
\end{bmatrix}
\right) =
f \left(
\begin{bmatrix}
t_1 \\
t_2 \\
t_3 \\
t_4
\end{bmatrix}
\right)^2.
\end{align*}

\subsection{Regressive Point Processes}

Unlike a latent-state formulation, the regressive point processes model the dependence on past events explicitly, by using relative times as input instead of absolute times \citep{hawkes2018hawkes,liu2019nonparametric}. \citet{liu2019nonparametric} propose the conditional \textsc{Gp} regressive point process (\textsc{Cgprpp}), which defines a latent function $f \sim \mathcal{GP}$ that takes time differences of the last $D$ events relative to a future time point $\tau$ as input: $f: \mathbb{R}_{\geq 0}^D \rightarrow \mathbb{R}$. They define $\tau$ to be a query time and $r: \mathbb{R}_{\geq 0} \rightarrow \{\mathbb{R}_{\geq 0} \cup \infty \}^D$ be a function that returns the relative times of the last $D$ events: $r(\tau)=(\Delta t_1, \ldots, \Delta t_D)$, where $\Delta\tau_d$ denotes the relative time between $\tau$ and $d^{\text{th}}$ last event: $\Delta\tau_d = \tau - t_{d}$ for $d \in 1, \ldots, D$:
$$
r(\tau) = \{\Delta\tau_d\}_{d=1}^{D} = (\tau - t_{1}, \ldots, \tau - t_D).
$$
If $d^{\text{th}}$ last event does not exist, the function $r(\tau)$ returns the placeholder value $\infty$ for the $d^{\text{th}}$ index.

\textsc{Cgprpp} regresses relative times $\{\Delta\tau_d\}_{d=1}^D$ on the latent intensity $f(\tau)$, which is passed through a square transformation to obtain the conditional intensity value $\lambda^*(\tau)$. 
As an example, the intensity values for points $\{t_i\}_{i=1}^4$ in Figure \ref{fig:regressivepp}, with $D=2$, is given by:
\begin{align*}
\lambda^*
\left(
\begin{bmatrix}
t_1 \\
t_2 \\
t_3 \\
t_4
\end{bmatrix}
\right) =
f \left(
\begin{bmatrix}
\infty & \infty \\
t_2 - t_1 & \infty \\
t_3 - t_2 & t_3 - t_1 \\
t_4 - t_3 & t_4 - t_2
\end{bmatrix}
\right)^2.
\end{align*}

\section{Treatment Model Details}
\label{sec:treatment-details}

\subsection{Model Definition}

Similar to the \textsc{Cgprpp} model proposed by \citet{liu2019nonparametric}, we model the treatment time intensity $\lambda^*(\tau)$ as the square transformation of a latent function, which is a sum of a constant scalar $\beta_0$ and three time-dependent functions with \textsc{Gp} priors, $g_b, g^*_a, g^*_o \sim \mathcal{GP}$. The latent-state function $g_b: \mathbb{R}_{\geq 0} \rightarrow \mathbb{R}$ models the baseline intensity. The regressive component $g^*_a: \{\mathbb{R}_{\geq 0} \times \mathbb{R}\}^{Q_a} \rightarrow \mathbb{R}$ models the dependence on past treatments by taking the last $Q_a$ treatment pairs (treatment time and dosage) as input. Similarly, the regressive component $g^*_o: \{\mathbb{R}_{\geq 0} \times \mathbb{R}\}^{Q_o} \rightarrow \mathbb{R}$ models the dependence on past outcomes by taking the last $Q_o$ outcome pairs (measurement time and value) as input. 

The treatment time intensity $\lambda^*(\tau)$ is defined as follows:
\begin{equation*}
    \lambda^*(\tau) = \big( \underbrace{\beta_0}_{\substack{\text{PP} \\ \text{baseline}}} + \underbrace{g_b(\tau)}_{\substack{\text{NHPP} \\ \text{Baseline}}} + \underbrace{g^*_a(\tau)}_{\substack{\text{Treatment} \\ \text{Effect}}} + \underbrace{g^*_o(\tau)}_{\substack{\text{Outcome} \\ \text{Effect}}} \big)^2.
\end{equation*}

\subsection{Kernel Definition}
\label{subsec:treatment_kernel}

Regressive components $g^*_a$ and $g^*_o$ explicitly model how the intensity depends on the past history. They take relative times to the last $Q_a$ treatments and $Q_o$ outcomes as input, while the baseline function $g_b$ takes the absolute time. For example, let $\tau$ be a query time. Let $(t_{a_1}, \ldots, t_{a_{Q_a}})$ be the arrival times of last $Q_a$ treatments before time $\tau$. Similarly, let $(t_{o_1}, \ldots, t_{o_{Q_o}})$ be the list of arrival times of the last $Q_o$ outcomes, before time $\tau$. Then, three function components has the following $1+2*Q_a+2*Q_o$ dimensional vector as input:
$$
\tau, \underbrace{(\tau - t_{a_1}, m_1), \ldots}_{\text{last } Q_a \text{ Treatments}}, \underbrace{(\tau - t_{o_1}, y_1), \ldots}_{\text{last } Q_o \text{ Outcomes}}
$$

To represent the information regarding the last $Q_a+Q_o$ events, we define retrieval functions:
\begin{itemize}
    \item Let $r_{k_i,t}: \mathbb{R}_{\geq 0} \rightarrow \mathbb{R}_{\geq 0}$ denote a retrieval function that takes time $\tau$ as input and outputs the time of the $i^{\text{th}}$ last event of type $k \in \{a, o\}$. For example, the retrieval function $r_{a_1,t}$ outputs the time of the last action occurred before time $\tau$: $r_{a_1,t}(\tau) = t_{a_1}$. Similar to $r_{k_i,t}$, let $\Delta_{k_i,t}: \mathbb{R}_{\geq 0} \rightarrow \mathbb{R}_{\geq 0}$ be a retrieval function that returns the relative time: $\Delta_{a_1,t}(\tau) = \tau - t_{a_1}$.
    \item Let $r_{k_i,m}: \mathbb{R}_{\geq 0} \rightarrow \mathbb{R}$ denote a retrieval function that takes time $\tau$ as input and returns the mark $m \in \mathbb{R}$ of $i^{\text{th}}$ last tuple of type $k \in \{a, o\}$.
    \item Let $r_{k_i}: \mathbb{R}_{\geq 0} \rightarrow \mathbb{R}_{\geq 0} \times \mathbb{R}$ denote a retrieval function that outputs both the relative time and the mark of the $i^{\text{th}}$ last event of type $k \in \{a, o\}$. For example, the retrieval function $r_{a_1}$ outputs the relative time and the mark of the last action occurred before time $\tau$: $r_{a_1}(\tau) = (\tau - t_{a_1}, m_1)$.
    \item For a query time $\tau$, the retrieval function $r(\tau): \mathbb{R}_{\geq 0} \rightarrow \mathcal{X}$ returns the overall input vector:
    $$
    r(\tau) = \{\tau, \underbrace{r_{a_{1}}(\tau), \ldots, r_{a_{Q_a}}(\tau)}_{\text{last } Q_a \text{ Treatments}}, \underbrace{r_{o_{1}}(\tau), \ldots, r_{o_{Q_o}}(\tau)}_{\text{last } Q_o \text{ Outcomes}}\},
    $$
    where $\mathcal{X}$ denotes the input domain $\mathcal{X} = \mathbb{R}_{\geq 0} \times \{\mathbb{R}_{\geq 0} \times \mathbb{R}\}^{Q_a} \times \{\mathbb{R}_{\geq 0} \times \mathbb{R}\}^{Q_o}$.
\end{itemize}

We represent the unavailable past information by an identifier value $r_{k_i} = (\infty, \infty)$, similar to \cite{liu2019nonparametric}. To set covariance values concerning the unavailable information to $0$, an indicator function $\mathbbm{1}_k: \mathbb{R}_{\geq 0} \times \mathbb{R} \rightarrow \{0, 1\}$ is defined: 
\begin{align*}
    \mathbbm{1}_k[r_{k_i}(\tau)] = \begin{cases}
      1, & \text{if}\ r_{k_i,t} < \infty, r_{k_i,m} < \infty \\
      0, & \text{otherwise}
    \end{cases}
\end{align*}
Using the indicator function, the kernel function is defined as follows:
\begin{align*}
    K(v, v') = K_{b}(\tau, \tau')
    + \sum_{k \in \{a,o\}} \sum_{i=1}^{Q_k} \mathbbm{1}[r_{k_i}(\tau)] \mathbbm{1}[r_{k_i}(\tau')] K_{k_i}(r_{k_i}(\tau), r_{k_i}(\tau')),
\end{align*}
where $K_{b}$ is an \textsc{Se} kernel, $v=r(\tau)$ and $ v'=r(\tau')$ are input vectors, and the single-event kernel function $K_{k_i}: \mathbb{R}_{\geq 0} \times \mathbb{R} \rightarrow \mathbb{R}$ is a sum of \textsc{Se} kernels acting on relative time and mark dimensions independently:
\begin{align*}
    K_{k_i}(r_{k_i}(\tau), r_{k_i}(\tau')))
    &= \gamma_i \underbrace{\exp \left(-\frac{(\Delta_{k_i,t}(\tau) - \Delta_{k_i,t}(\tau'))^2}{\sigma_{i,t}} \right)}_{\text{Relative Time Kernel } K_t} \underbrace{\exp \left(-\frac{(r_{k_i,m}(\tau) - r_{k_i,m}(\tau'))^2}{\sigma_{i,m}} \right)}_{\text{Mark Kernel } K_m} \\
    &= \gamma_i K_t(\Delta_{k_i,t}(\tau), \Delta_{k_i,t}(\tau')) K_m(r_{k_i,m}(\tau), r_{k_i,m}(\tau'))
\end{align*}
The overall kernel is a sum of the baseline \textsc{Se} kernel and $Q_a+Q_o$ two-dimensional single-event kernels $K_k$, each of which is an \textsc{Se} kernel on two dimensions: relative time and mark.

\subsection{Likelihood}

The likelihood for a set of treatments $\mathcal{D} = \{(t_n, m_n)\}_{n=1}^N$ can be written in terms of the conditional intensity function $\lambda^*(t,m) = \lambda^*(t) p^*(m \mid t)$:
\begin{align}
\label{eq:likelihood_pp}
    p(\mathcal{D} \mid \lambda^*(\cdot))  
    &= \prod_n \lambda^*(t_n) p^*(m_n \mid t_n) \times \exp \Big\{ -\int_{\mathcal{T}} \lambda^*(\tau) d\tau \Big\}
\end{align}

The mark intensity $p^*(m_n \mid t_n)$ factorizes and is modeled by a \textsc{Gp} prior independent from the time intensity $\lambda^*(t)$. The inference for the mark intensity $p^*(m_n \mid t_n)$ is straightforward. Therefore, in the following, we derive the inference objective for the time intensity $\lambda^*(t)$.

\subsubsection{Multiple Observations}

We assume multiple observations $\mathcal{D} = \{\mathcal{D}_1, ..., \mathcal{D}_O\}_{o=1}^O$ are conditionally independent given the treatment intensity $\lambda^*$, e.g., daily meals for a patient for $O$ days, where we consider each day as conditionally independent. Then, the likelihood factorizes over multiple observations:
\begin{align*}
    p(\mathcal{D}_1, ..., \mathcal{D}_O \mid \lambda^*(\cdot)) &= \prod_o p(\mathcal{D}_o \mid \lambda^*(\cdot)).
\end{align*}
where the term $p(\mathcal{D}_o \mid \lambda^*(\cdot))$ follows \cref{eq:likelihood_pp}.

\subsection{Variational Inference}
\label{ssec:vi}

In variational inference, a lower bound (\textsc{Elbo}), denoted by $\mathcal{L}$, of the log likelihood $\log p(\mathcal{D})$ is maximized, which is equivalent to minimizing the \textsc{KL} divergence between the variational distribution $q(f)$ and the true posterior $p(f \mid \mathcal{D})$:
\begin{align*}
    \log p(\mathcal{D}) &= \mathcal{L} + \text{KL}[q(f) \mid \mid p(f \mid \mathcal{D})],
\end{align*}
where the lower bound $\mathcal{L}$ is given by:
\begin{align*}
    \mathcal{L} &= \EX_q[\log p(\mathcal{D} \mid f)] - \text{KL}[q(f) \mid \mid p(f)] \\
    &= \mathcal{L}_{\mathcal{D}} - \text{KL}[q(f) \mid \mid p(f)].
\end{align*}

\subsubsection{Inducing Points}

Inducing point approximations are commonly used for scalable inference in \textsc{Gp}s. Let $\mathbf{Z} = \{z_m\}_{m=1}^M$ be a set of inducing points. Their function evaluations are collected in a set of inducing variables $\mathbf{u} = f(Z) \in \mathbb{R}^M$, where each $u_m = f(z_m)$. Conditioned on inducing variables $\mathbf{u}$, we assume that the variational conditional distribution $q(\mathbf{f} \mid \mathbf{u})$ is equal to the true conditional $p(\mathbf{f} \mid \mathbf{u})$. Then, the variational distribution $q(\mathbf{f, u})$ can be written as follows:
\begin{align*}
    q(\mathbf{f}, \mathbf{u}) &= p(\mathbf{f} \mid \mathbf{u}) q(\mathbf{u}),
\end{align*}
where $q(\mathbf{u}) \sim N(\mathbf{u}; \mathbf{m}, \mathbf{S})$. Generally, we integrate out $\mathbf{u}$:
\begin{align*}
    q(\mathbf{f}) &= \int_{\mathcal{U}} p(\mathbf{f} \mid \mathbf{u}) q(\mathbf{u}) d\mathbf{u} = \mathcal{GP}(f; \Tilde{\mu}, \Tilde{\Sigma}), \\
    \Tilde{\mu}(\mathbf{x}) &= \mathbf{k_{xz}} \mathbf{K}^{-1}_{\mathbf{zz}} \mathbf{m}, \\
    \Tilde{\Sigma}(\mathbf{x}, \mathbf{x'}) &= \mathbf{K_{xx'}} - \mathbf{k_{xz}} \mathbf{K}^{-1}_{\mathbf{zz}} \mathbf{k_{zx}} + \mathbf{k_{xz}} \mathbf{K}^{-1}_{\mathbf{zz}} \mathbf{S} \mathbf{K}^{-1}_{\mathbf{zz}} \mathbf{k_{zx'}}.
\end{align*}

\subsubsection{Evidence Lower Bound (\textsc{Elbo})}
\label{subsec:elbo}

The lower bound $\mathcal{L}$ is:
\begin{align*}
    \mathcal{L} &= \mathcal{L}_{\mathcal{D}} - \text{KL}[q(f) \mid \mid p(f)].
\end{align*}
Similar to \citet{lloyd2015variational,matthews2016sparse,john2018large}, we compute the KL term $\text{KL}[q(f) \mid \mid p(f)]$ by computing the KL divergence at inducing points $\text{KL}[q(\mathbf{u}) \mid \mid p(\mathbf{u})]$:
\begin{align*}
    \mathcal{L} &= \mathcal{L}_{\mathcal{D}} - \text{KL}[q(\mathbf{u}) \mid \mid p(\mathbf{u})]
\end{align*}
We now consider the log-likelihood expectation term $\mathcal{L}_\mathcal{D}=\EX_q[\log p(\mathcal{D} \mid f)]$:
\begin{align*}
    \mathcal{L}_{\mathcal{D}} &= \sum_o \textcolor{violet}{\sum_n \EX_q[\log \lambda^*(t^{(o)}_n)]} - \sum_o \textcolor{blue}{\EX_q\Big[ \int_{\mathcal{T}} \lambda^*(t) dt \Big]} \\
    &= \sum_o \textcolor{violet}{\mathcal{L}^{(o)}_n} - \sum_o \textcolor{blue}{\mathcal{L}^{(o)}_t}
\end{align*}
First, we start with the data term $\textcolor{violet}{\mathcal{L}^{(o)}_n}$ for a single observed point process realization, indexed by $(o)$:
\begin{align*} 
    \textcolor{violet}{\mathcal{L}_n^{(o)}} &= \sum_n \EX_q[\log \lambda^*(t_n)] \\
    &= \sum_n \EX_q[\log (f(r(t_n)) + \beta)^2],
\end{align*}
As \citet{john2018large}, we use a change of variables $f = f_n + \beta$ and exploit the computational trick used in \citet{lloyd2015variational} to compute the one-dimensional integral $\EX_q[\log f_n^2]$:
\begin{align*} 
    \EX_q[\log f^2] &= \int_{-\infty}^{\infty} \log (f^2) \mathcal{N}(f; \Tilde{\mu}, \Tilde{\sigma}^2) df \\
    &= -\Tilde{G} \Big( -\frac{\Tilde{\mu}^2}{2\Tilde{\sigma}^2} \Big) + \log \Big( \frac{\Tilde{\sigma}^2}{2} \Big) - C
\end{align*}
where $C$ is the Euler-Mascheroni constant and $\Tilde{G}$ is defined by the confluent hyper-geometric function, which is approximated by a look-up table for faster computation.

To compute the integral term $\textcolor{blue}{\mathcal{L}_t^{(o)}}$, we divide the observation period into $N+1$ intervals with end points $(0, t_1, \ldots, t_{N}, T)$, where $\{t_i\}_{i=1}^N$ includes all event points that have an effect on the conditional intensity value. In each interval $[t_{n}, t_{n+1}]$, the conditional intensity values $\{\lambda^*(\tau): \tau \in [t_{n}, t_{n+1}]\}$ can be estimated using the past history $\mathcal{H}_{<t_n}$. Using a shorthand notation for $\mathbf{r} = r(t) \in \mathcal{X}$, we write the integral term as a sum of three integrals:
\begin{align}
\label{eq:integral_term}
    \textcolor{blue}{\mathcal{L}_t^{(o)}} &= \EX_q\Big[ \int_{\mathcal{T}} \lambda^*(\tau) d\tau \Big], \nonumber \\
    &= \EX_q\Big[ \int_{\mathcal{T}} (f(r(\tau)) + \beta)^2 d\tau \Big], \nonumber \\
    &= \Big[ \int_{\mathcal{T}} \Big( \EX_q[f^2(\mathbf{r})]  + 2 \beta \EX_q[f(\mathbf{r})] + \beta^2 \Big)  d\tau \Big], \nonumber \\
    &= \Big[ \int_{\mathcal{T}} \EX_q[f(\mathbf{r})]^2 d\tau + \int_{\mathcal{T}} \text{Var}_q[f(\mathbf{r})] d\tau + 2 \beta \int_{\mathcal{T}} \EX_q[f(\mathbf{r})] d\tau + \beta^2 |\mathcal{T}| \Big], \nonumber \\
    &= \sum_{n=1}^{N+1} \Big[ \int_{t_{n-1}}^{t_n} \EX_q[f(\mathbf{r})]^2 d\tau + \int_{t_{n-1}}^{t_n} \text{Var}_q[f(\mathbf{r})] d\tau + 2 \beta \int_{t_{n-1}}^{t_n} \EX_q[f(\mathbf{r})] d\tau + \beta^2 |\mathcal{T}| \Big].
\end{align}
We can compute each integral term as follows: 
\begin{align*}
    \int_{t_{n-1}}^{t_n} \EX_q[f(r(\tau))]^2 d\tau &= \mathbf{m^T K^{-1}_{zz} \Psi_n K^{-1}_{zz} m}, \\
    \int_{t_{n-1}}^{t_n} \text{Var}_q[f(\tau)] d\tau &= \sum_{d=1}^D \gamma_d \int_{t_{n-1}}^{t_n} \mathbbm{1}[h_d(t)] dt + \text{Tr}(\mathbf{K^{-1}_{zz} \Psi_n}) + \text{Tr}(\mathbf{K^{-1}_{zz} S K^{-1}_{zz} \Psi_n}),
    \\
    \int_{t_{n-1}}^{t_n} \EX_q[f(\tau)] d\tau &= \mathbf{\Phi_n^T K^{-1}_{zz} m}, \\
    \mathbf{\Psi_n}(\mathbf{z}, \mathbf{z'}) &= \int_{t_{n-1}}^{t_n} K(\mathbf{z}, r(\tau)) K(r(\tau), \mathbf{z'}) d\tau, \\
    \mathbf{\Phi_n}(\mathbf{z}) &= \int_{t_{n-1}}^{t_n} K(\mathbf{z}, r(\tau)) d\tau.
\end{align*}
Then, we plug in all integral terms into \cref{eq:integral_term}:
\begin{align*}
    \textcolor{blue}{\mathcal{L}^{(o)}_t} &= \mathbf{m^T K^{-1}_{zz} \Psi_n K^{-1}_{zz} m} \\
    &+ \sum_{d=1}^D \gamma_d \int_{t_{n-1}}^{t_n} \mathbbm{1}[h_d(t)] dt + \text{Tr}(\mathbf{K^{-1}_{zz} \Psi_n}) + \text{Tr}(\mathbf{K^{-1}_{zz} S K^{-1}_{zz} \Psi_n}) \\
    &+ 2 \beta \mathbf{\Phi_n^T K^{-1}_{zz} m} + \beta^2 |\mathcal{T}_o|.
\end{align*}
The phi vector $\mathbf{\Phi_n}$ and the psi matrix $\mathbf{\Psi_n}$ have closed form solutions, obtained by evaluating the integrals for the sum of \textsc{Se} kernels. The difference to \citet{liu2019nonparametric} is the inclusion of the \textsc{Se} mark kernel terms $k_m(\cdot, \cdot)$.
\begin{align*}
    \Phi_n(z) = \sum_{q=1}^{Q_a+Q_o} &\mathbbm{1}[z_q] \mathbbm{1}[r_q(t_n)] \gamma_q \frac{\sqrt{\pi \alpha_{q,t}}}{\sqrt{2}} 
    \exp \left(\frac{(m_{z_q} - r_{q,m}(t_n))^2}{\alpha_{q,m}} \right) \\ 
    & \left[ \text{erf}\left( \frac{t_n - r_{q,t}(t_n) - t_{z_q}}{\sqrt{2 \alpha_{q,t}}} \right) - \text{erf}\left( \frac{t_{n-1} - r_{q,t}(t_n) - t_{z_q}}{\sqrt{2 \alpha_{q,t}}} \right) \right],
\end{align*}
\begin{align*}
    \Psi_n(z, z') = \sum_{i,j=1}^{Q_a+Q_o} &\mathbbm{1}[z_i] \mathbbm{1}[z_j] \mathbbm{1}[r_i(t_n)] \mathbbm{1}[r_j(t_n)] \gamma_i \gamma_j \frac{\sqrt{\pi \alpha_{i,t} \alpha_{j,t}}}{\sqrt{2(\alpha_{i,t} + \alpha_{j,t})}} \\
    & \exp \left(\frac{(m_{z_i} - r_{i,m}(t_n))^2}{\alpha_{i,m}} \right) \exp \left(\frac{(m_{z_j} - r_{j,m}(t_n))^2}{\alpha_{j,m}} \right) \\
    & \exp \left( -\frac{(t_{z_i} + r_{i,t}(t_n) - t'_{z'_j} - r_{j,t}(t_n))^2}{2(\alpha_{i,t} + \alpha_{j,t})} \right) \\
    & \Bigg[ \text{erf}\left( \frac{ \alpha_i(t_n - r_{j,t}(t_n) - t'_{z'_j}) + \alpha_j(t_n - r_{i,t}(t_n) - t_{z_i}) }{\sqrt{2 \alpha_{i,t}\alpha_{j,t} (\alpha_{i,t} + \alpha_{j,t})}} \right) \\
    & - \text{erf}\left( \frac{ \alpha_i(t_{n-1} - r_{j,t}(t_n) - t'_{z'_j}) + \alpha_j(t_{n-1} - r_{i,t}(t_n) - t_{z_i}) }{\sqrt{2 \alpha_{i,t}\alpha_{j,t} (\alpha_{i,t} + \alpha_{j,t})}} \right) \Bigg].
\end{align*}

\section{Outcome Model Details}
\label{sec:outcome-details}

We model the outcome trajectory $\mathbf{Y} = \{y(\tau): \tau \in \mathbb{R}_{\geq 0}\}$ by a conditional \textsc{Gp} model, combining three independent function components: (i) a baseline progression, (ii) a treatment response function and (iii) a noise variable \citep{schulam2017reliable,zhang2020errors}:
\begin{align}
    y(\tau) &= \underbrace{f_b(\tau)}_{\text{Baseline}} + \underbrace{f_a(\tau; \mathbf{a})}_{\text{Treatment Response}} + \underbrace{\epsilon(\tau)}_{\text{Noise}},
\end{align}

\textbf{Baseline progression.} The baseline progression $f_b(\tau)$ over time $\tau$ can be modeled by a \textsc{Gp} prior. The kernel function is chosen depending on the application context. For example, a baseline function $f_b(\tau) \sim \mathcal{GP}(0, K_{se}+K_{per})$, with a zero mean and a kernel function equal to the sum of a squared exponential (SE) and a periodic (PER) kernel was proposed to model blood pressure and the heart rate measurements in \citet{cheng2020patient}. A combination of two non-stationary kernel functions was chosen as the baseline kernel function to model creatinine measurements in \citet{schulam2017reliable}. In \citet{xu2016bayesian}, the baseline kernel function was chosen as the sum of a linear and an exponential kernel. In our experiments, we model the baseline progression of blood glucose as a sum of a constant and a periodic kernel to capture daily blood glucose profiles of non-diabetic patients fluctuating around a constant baseline \citep{ashrafi2021computational}.

\textbf{Treatment Response Model.} We define the treatment response function $f_a(\tau;\mathbf{a})$ as a Gaussian process: $f_a(\tau;\mathbf{a}) \sim \mathcal{GP}$, similar to \citet{cheng2020patient}. While \citet{cheng2020patient} propose an extension using a latent force model \citep{alvarez2009latent}, we use an \textsc{Se} kernel, as it produces sufficient performance for our use case. For each treatment, we consider the treatment response as additive:
\begin{align*}
    f_a(\tau; \mathbf{a}) &= \sum_{a_i=(t_i,m_i) \in \mathbf{a}} f_m(m_i) f_t(\tau; t_i),
\end{align*}
where the scaling function $f_m: \mathbb{R} \rightarrow \mathbb{R}$ is a function of the treatment mark $m_i \in \mathbb{R}$. Each treatment type is assumed to have a distinct pair of a scaling function $f_m$ and a time-dependent response function $f_t$. The scaling function $f^{(v)}_m(m_i)$ is a linear function that captures the individualized treatment response magnitude for each patient $(v)$: 
$$
f^{(v)}_m(m_i) = \beta^{(v)}_0 + \beta^{(v)}_1 m_i,
$$
where a hierarchical Gaussian prior is placed on parameters $\{\beta^{(v)}_0, \beta^{(v)}_1\}$: $\beta^{(v)}_0 \sim \mathcal{N}(\beta_0, \sigma^2_0)$, $\beta^{(v)}_1 \sim \mathcal{N}(\beta_1, \sigma^2_1)$.

The time-dependent function $f_t(\tau;\mathbf{a})$ is defined as a piecewise function that is equal to $0$ outside treatment response intervals $\mathcal{T}_{t_i} = [t_i, t_i+T_i]$ and $\mathcal{T}_{t'_{i}} = [t'_i, t'_i+T'_i]$:
\begin{align*}
    f_t(\tau; t_i) = \begin{cases}
      \mathcal{GP}(0, k_{f_t, f_t'}(\tau, \tau'; t_i, t_i')), & \text{if}\  \tau \in \mathcal{T}_{t_i}, \tau' \in \mathcal{T}_{t'_{i}} \\
      0, & \text{otherwise}
    \end{cases},
\end{align*}
where the treatment response interval $T_i$ is set by domain knowledge. The kernel function works with non-negative relative times $\Delta \tau_i = \textsc{Relu}(\tau - t_i)$, rather than the absolute time $\tau$:
\begin{align*}
    k_{f_t, f_t'}(\tau, \tau'; t_i, t_i') = \exp\left(-\frac{\left(\Delta \tau_i - \Delta \tau_i')\right)^2}{l_t^2}\right).
\end{align*}
The adoption of the piecewise definition and relative times is enforcing a `causal' kernel for the treatment response, as in `causal' \textsc{Gp} proposed by \citet{cunningham2012gaussian}. Here, the term `causal' is used differently than in the causality discussion in previous sections and has more of a common sense meaning implying the time direction.

\section{Learning}
\label{sec:learning}

For the treatment model, we maximize the \textsc{Elbo} $\mathcal{L}$ with respect to variational parameters $\theta_{a,z}=\{\mathbf{m_z}, \mathbf{S_z}\}$ and kernel hyperparameters $\theta_{a,h}=\{\gamma_q,\alpha_q\}_{q=0}^{Q_a+Q_o}$ (see \cref{ssec:vi} for details). For the outcome model, we maximize the marginal likelihood $p(\mathcal{D} \mid \theta_{o,h})$ with respect to model hyperparameters $\theta_{o,h} = \{b_b^{(v)}, \ell_b^{(v)}, \alpha_b^{(v)}\}_{v=1}^{N_\text{patient}} \cup \{ \ell_t, \beta_0, \beta_1, \{\beta_0^{(r)}, \beta_1^{(v)} \}_{v=1}^{N_\text{patient}}\}$, where $\{b_b^{(v)}, \alpha_b^{(v)}, \ell_b^{(v)}\}$ denote the parameters of the individual-specific baseline function and $\{\beta_0,\beta_1, \beta_0^{(v)},\beta_1^{(v)}\}$ denote the parameters of the linear, hierarchical model for the scale function $f^{(v)}_m(\cdot)$.

\section{Inference}

\subsection{Test Log-Likelihood (\textsc{Tll})}
\label{subsec:tll}

The outcome model is a \textsc{Gp} prior with an independent Gaussian noise. The posterior of the latent function $\mathbf{f_q}$ at query times $\mathbf{q}$ follows a multivariate normal distribution and its mean and covariance functions have closed form solutions: $\mathbf{f_q} \mid \mathbf{y} \sim \mathcal{N}(\tilde{\mu}_q, \tilde{\Sigma}_q)$, with $\hat{\mu}_q=\mathbf{k_*^T} (\mathbf{K}+\sigma^2_{\epsilon}\mathbf{I})^{-1}\mathbf{y}$ and $\hat{\Sigma}_q=\mathbf{K_{**}} - \mathbf{k_*^T}(\mathbf{K}+\sigma^2_{\epsilon}\mathbf{I})^{-1} \mathbf{k_*}$. The test log-likelihood has a closed form solution and can be computed by integrating out the posterior $\mathbf{f_q} \mid \mathbf{y}$:
\begin{align*}
    \log p(\mathcal{D}_\text{test} \mid \mathcal{D}) &= \log \int_{\mathbf{f_q}} \log p(\mathcal{D}_\text{test} \mid \mathbf{f_q}) p(\mathbf{f_q} \mid \mathcal{D}) d\mathbf{f_q}.
\end{align*}
For the treatment model, the marginal log-likelihood is intractable. A lower-bound on the test log-likelihood $\log p(\mathcal{D}_\text{test} \mid \mathcal{D})$ can be computed as follows:
\begin{align*}
    \log p(\mathcal{D}_\text{test} \mid \mathcal{D}) &= \log \int_{g^*} \log p(\mathcal{D}_\text{test} \mid g^*) p(g^* \mid \mathcal{D}) dg^*,  \\
    &= \log \EX_{p(g^* \mid \mathcal{D})}\left[ p(\mathcal{D}_\text{test} \mid g^*) \right], \\
    &\approx \log \EX_{q(g^*)}\left[ p(\mathcal{D}_\text{test} \mid g^*) \right], \\
    &\geq \EX_{q(g^*)}\left[ \log p(\mathcal{D}_\text{test} \mid g^*) \right],
\end{align*}
where we approximate the true posterior $p(g^* \mid \mathcal{D})$ by the variational distribution $q(g^*) = p(g^* \mid u) q(u)$. The term $\EX_{q(g^*)}\left[ \log p(\mathcal{D}_\text{test} \mid g^*) \right]$ can be computed as in the steps detailed in Appendix \ref{subsec:elbo}, as it has the same form as the expected log likelihood component of the \textsc{Elbo}.

\begin{figure}[h]
  \centering
  \begin{minipage}{1.0\linewidth}
    \begin{algorithm}[H]
        \caption{Ogata's Thinning algorithm}
        \label{alg:ogata}
        \textbf{Input:} Start $T_1$, End $T_2$, Interval function $l(\cdot)$, Conditional intensity $\lambda^*(\cdot)$. \\
        \textbf{Output:} Point process sample $\mathcal{T} = \{t_1, \ldots, t_n\}$in the interval $[T_1, T_2]$.
        \begin{algorithmic}[1]
        \Function{Sample-ogata}{$T_1, T_2, l, \lambda^*$}
            \State $\tau = T_1, \mathcal{T} = \emptyset$. \Comment{Initialize}
            \While{$\tau < T_2$}
                \State $\lambda_{ub} = \sup_{s \in [\tau, \tau+l(\tau)]} \lambda^*(s)$.  \Comment{Upper-bound.}
                \State $u_{ub}, u_a  \sim \mathcal{U}(0,1)$.  \Comment{Sample noise variables.}
                \State $t_i = -1/\lambda_{ub} * \log(u_{ub})$  \Comment{Draw the inter-arrival time, $t_i \sim \operatorname{Exp}(\lambda_{ub})$.}
                \If {$t_i \leq l(\tau)$} \Comment{Candidate in the interval.}
                    \If {$u_a \leq \lambda^*(\tau+t_i) / \lambda_{ub}$} \Comment{Keep with probability $\lambda^*/\lambda_{ub}$}
                        \State $\mathcal{T} = \mathcal{T} \cup \{\tau+t_i\}$. \Comment{Point accepted.}
                    \EndIf
                    \State $\tau = \tau + t_i$. \Comment{Continue from the candidate point.}
                \Else  \Comment{No candidate point in the interval.}
                    \State $\tau = \tau + l(\tau)$. \Comment{Continue from the end of the interval.}
                \EndIf
            \EndWhile
            \State \Return $\mathcal{T} = \{t_1, \ldots, t_n\}$.
        \EndFunction
        \end{algorithmic}
    \end{algorithm}
  \end{minipage}
\end{figure}

\subsection{Counterfactual Query}
\label{subsec:cf_query}

For a policy counterfactual, we estimate the potential outcome of the past trajectory $P(\mathbf{Y}_{\leq T}[\tilde{\pi}_{\leq T}] \mid \mathcal{H}_{\leq T}, \pi_{\leq T})$, under a new treatment policy specified by $\tilde{\pi}_{\leq T}$ different than the observed policy specified by $\pi_{\leq T}$. To answer a counterfactual query, we require access to the posterior distribution of noise variables $\mathbf{N}$ in addition to the counterfactual intensity $\lambda^*_{cf}$.

In our model, the noise consists of two components: treatment and outcome noise, $\mathbf{N} = \{\mathcal{E}_a, \mathcal{E}_o\}$. The outcome noise $\mathcal{E}_o$ is defined in \cref{eq:outcome} and it is equal to the difference between the outcome measurement $y(\tau)$ and the latent function $f(\tau)$. Hence, its posterior is available in closed form from the \textsc{Gp} posterior:
$$\epsilon(\tau) \mid \mathbf{y} = y(\tau) - f(\tau) \mid \mathbf{y}.$$
In our experiments, we use the mean of the noise posterior as a point estimate:
$$\EX \big[\epsilon(\tau) \mid \mathbf{y}\big] = y(\tau) - \EX \big[f(\tau) \mid \mathbf{y} \big].$$

The treatment noise $\mathcal{E}_a$ comprises noise associated with the sampling process of the treatments. In practice, we extend the counterfactual sampling algorithm for non-homogeneous Poisson processes provided in \citet{noorbakhsh2021counterfactual} to history-dependent point processes where future events depend on past events. The key difference in algorithms is that the counterfactual sampling algorithm provided in \citet{noorbakhsh2021counterfactual} is based on Lewis' thinning algorithm \citep{lewis1979simulation}, while our algorithm is based on Ogata's thinning algorithm \citep{ogata1981lewis}. Both thinning algorithms follow a general recipe: they first sample candidate points from an upper-bounding intensity and then perform rejection sampling with a probability proportional to the target intensity, so that the intensity of accepted points is equivalent to the target intensity. Different than Lewis' algorithm, Ogata's algorithm considers smaller intervals where the target intensity can be upper-bounded. In the following, we first discuss Ogata's thinning algorithm and then present our counterfactual sampling algorithm. We conclude by discussing the identifiability of the counterfactual query.

\subsubsection{Ogata's Thinning Algorithm}

Consider an initial time point $s$, where we start sampling. Ogata's algorithm first selects an interval such as $[s, s+l(s)]$, where the function $l(\cdot)$ specifies the length of the interval $l(s)$ until which it is possible to upper bound the conditional intensity $\lambda^*(\tau)$ \citep{rasmussen2011temporal}. The upper bound intensity should be greater than or equal to the supremum of the conditional intensity $\lambda^*(\tau)$ in the interval $[s, s+l(s)]$:
$$\lambda_{ub} \geq \sup_{\tau \in [s, s+l(s)]} \lambda^*(\tau).$$
Using the constant upper bound intensity, one can sample a candidate point from an upper-bounding Poisson process $t_i \sim \mathcal{PP}(\lambda_{ub})$. After sampling $t_i$, the algorithm considers two options:
\begin{itemize}
    \item If $t_i > l(s)$, then there exists no candidate points in the interval $[s,s+l(s)]$ and the sampling procedure continues from $s+l(s)$.
    \item If $t_i < l(s)$, $t_i$ is kept with the probability $\lambda^*(t_i)/ \lambda_{ub}$. In practice, a noise variable $U_i$ is sampled from a uniform distribution: $u_i \sim \mathcal{U}(0, 1)$ and the point is kept if $u_i \leq \lambda^*(t_i)/\lambda_{ub}$. Regardless of the thinning decision, the procedure continues from the time point $t_i$.
\end{itemize}
The sampling process is illustrated in Algorithm \ref{alg:ogata}.

\begin{figure}[h]
  \centering
  \begin{minipage}{1.0\linewidth}
    \begin{algorithm}[H]
        \caption{Counterfactual Sampling Algorithm for \textsc{Tpp}s (Based on Ogata's)}
        \label{alg:counterfactual_tpp}
        \textbf{Input:} Start $T_1$, End $T_2$, Interval function $l(\cdot)$, Observational intensity $\lambda_{obs}^*(\cdot)$, Counterfactual intensity $\lambda_{cf}^*(\cdot)$, Observed points $\mathcal{T}_{obs}$ \\
        \textbf{Output:} Counterfactual point process sample $\mathcal{T}_{cf} = \{t_1, \ldots, t_n\}$
        \begin{algorithmic}[1]
        \Function{Sample}{$T, l, \lambda^*$}
            \State $\tau = T_1, \mathcal{T} = \emptyset$. \Comment{Initialize}
            \While{$\tau < T_2$}
                \State $\lambda_{ub} = \sup_{s \in [\tau, \tau+l(\tau)]} \{\lambda^*(s): \lambda^* \in \{\lambda^*_{obs}, \lambda^*_{cf}\}\}$. \Comment{Shared upper-bound.}
                \State $t_{rej} =$ \textsc{Sample-ogata}$(\tau, \tau+l(t), \lambda_{ub}, \lambda_{ub}-\lambda^*_{obs})$.  \Comment{Draw a rejected point (\cref{alg:ogata}).}
                \If {$t_{rej} \leq l(\tau)$ and $t_{rej}+\tau \leq T$}
                    \State $u_{rej} \sim \mathcal{U}(\lambda^*_{obs}(\tau+t_r, \mathcal{H}_{obs}), \lambda_{ub})$. \Comment{Noise posterior.}
                    \If {$u_{rej} \leq \lambda^*_{cf}(\tau+t_r)$}
                        \State $\mathcal{T}_{cf} = \mathcal{T}_{cf} \cup \{\tau+t_r\}$.
                    \EndIf
                    \State $\tau = \tau + t_r$.
                \Else  \Comment{No rejections in interval.}
                    \If {$\tau+l(\tau) \in \mathcal{T}_{obs}$}  \Comment{Check if the end point is observed.}
                        \State $t_{obs} = \tau + l(\tau)$.
                        \State $u_{obs} \sim \mathcal{U}(0, \lambda^*_{obs}(t_{obs}, \mathcal{H}_{obs}))$. \Comment{Noise posterior.}
                        \If {$u_{obs} \leq \lambda^*_{cf}(t_{obs}, \mathcal{H}_{cf})$}
                            \State $\mathcal{T}_{cf} = \mathcal{T}_{cf} \cup \{t_{obs}\}$.
                        \EndIf
                    \EndIf
                    \State $\tau = \tau + l(\tau)$.
                \EndIf
            \EndWhile
            \State \Return $\mathcal{T}_{cf} = \{t_1, \ldots, t_n\}$.
        \EndFunction
        \end{algorithmic}
    \end{algorithm}
  \end{minipage}
\end{figure}

\subsubsection{Counterfactual Sampling Algorithm for \textsc{Tpp}s}
\label{sssec:cf_algo}

To estimate the policy counterfactual, we use the counterfactual treatment intensity $\lambda^*_{cf}$ and individual-specific noise distribution $p(\mathbf{N} \mid \mathcal{D}^{(r)})$ (see \cref{sec:inference}). We define treatment noise variables $\mathcal{E}_a$ by assuming that the generative process for observed treatments follows Ogata's thinning algorithm \citep{ogata1981lewis, noorbakhsh2021counterfactual}. Since the sampling algorithm is not actually used to generate the observed data, the noise variables are latent variables of the model.

In Ogata's algorithm, there are two sources for the stochasticity: (i) the candidate proposal noise and (ii) the acceptance noise. The candidate proposal noise tells us where the candidate points are. Since we already observe the accepted points, the missing part is the rejected points. Similar to \citet{noorbakhsh2021counterfactual}, we sample the rejected points from the `rejected' intensity equal to $\lambda^*_{rej} = \lambda_{ub} - \lambda^*_{obs}$, at each interval of the Ogata's algorithm. In addition to this, a noise variable is defined as the acceptance noise at every candidate point, i.e., at all rejected and accepted points.

For accepted points, rejected points and the noise posteriors to be well defined, we choose an upper-bound intensity $\lambda_{ub}$ that is an upper bound for both the observational intensity $\lambda^*_{obs}$ and the counterfactual intensity $\lambda^*_{cf}$, for each interval $[\tau, \tau+l(\tau)]$:
$$\lambda_{ub} \geq \sup_{s \in [\tau, \tau+l(\tau)]} \{\lambda^*(s): \lambda^* \in \{\lambda^*_{obs}, \lambda^*_{cf}\}\}.$$
In practice, we select the function $l(\tau)$ so that it returns the next event after time $\tau$, which can be a future treatment or an outcome. The observation period $[0,T]$ is split into intervals with end points $(0, t_1, \ldots, t_N, T)$, so that we can compute both the observational intensity $\lambda^*_{obs}$ and the counterfactual intensity $\lambda^*_{cf}$ for all points in each interval, conditioned on their respective histories $\mathcal{H}_{obs}$ and $\mathcal{H}_{cf}$. Here, the ordered set of events $t_1, \ldots, t_N$ are equal to the sorted union of all treatments and outcomes: $t_1, \ldots, t_N = \text{\textsc{Sort}}(\mathbf{t_a} \cup \mathbf{t_o})$.

\begin{figure}[h]
    \centering
    \begin{subfigure}{.9\textwidth}
        \centering
        \includegraphics[width=0.5\linewidth]{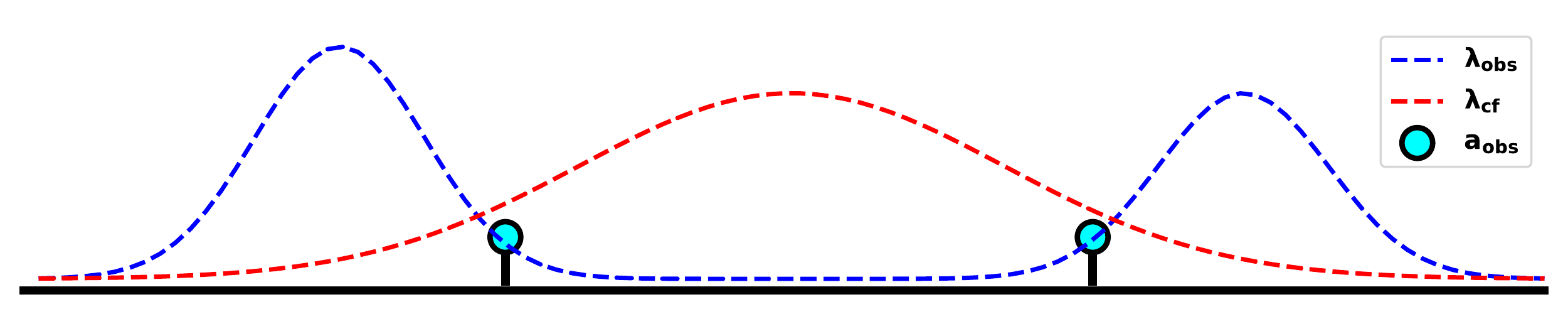}
        \caption{Initial: Observed treatments, observational and counterfactual intensity.}
     \end{subfigure}
     \begin{subfigure}{.45\textwidth}
        \centering
        \includegraphics[width=1.0\linewidth]{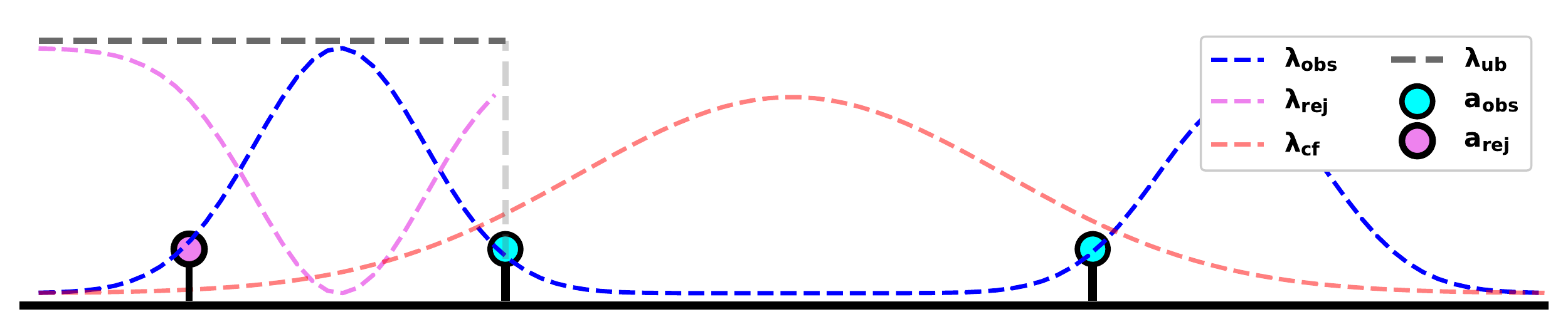}
        \caption{Step 1: Sample rejected events for the first interval.}
        \label{fig:step1}
        \includegraphics[width=1.0\linewidth]{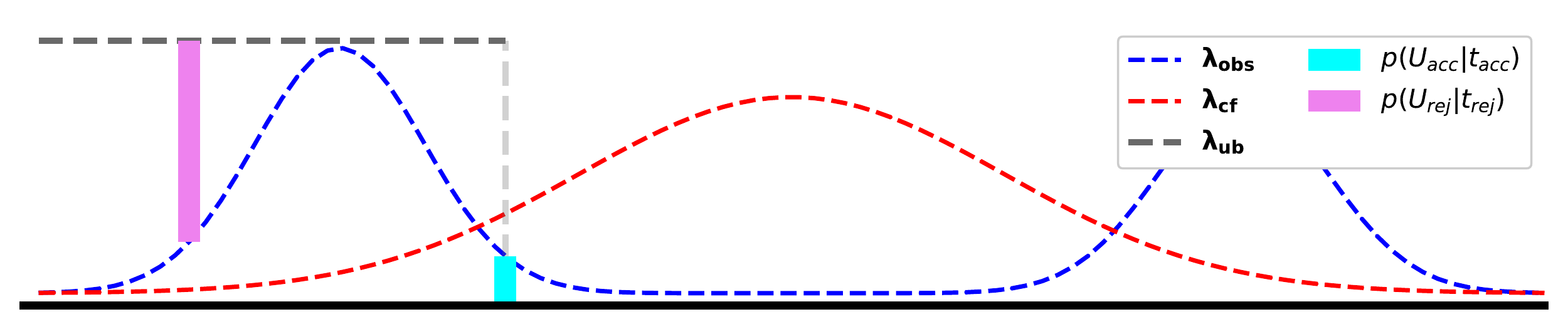}
        \caption{Step 2: Noise posteriors for the first interval.}
        \label{fig:step2}
        \includegraphics[width=1.0\linewidth]{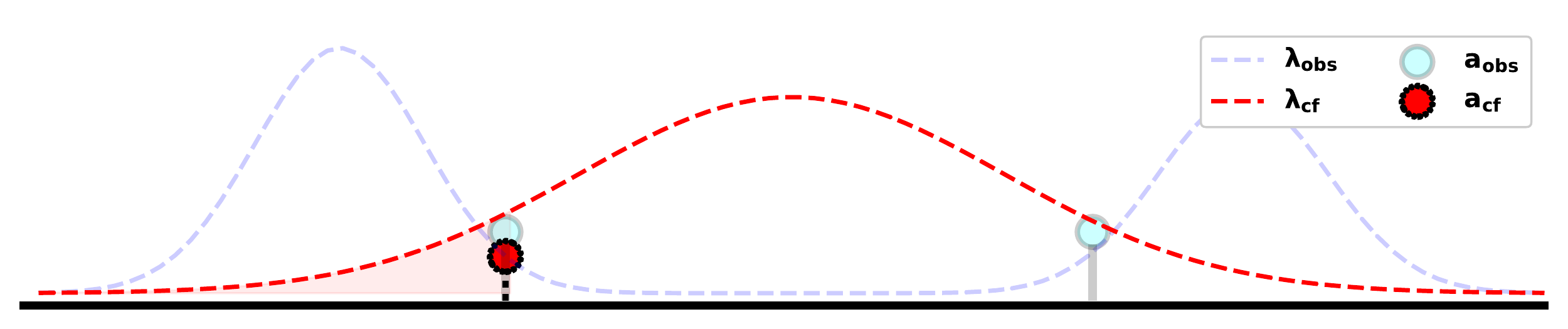}
        \caption{Step 3: Re-thin using the noise posteriors and the counterfactual intensity for the first interval.}
        \label{fig:step3}
     \end{subfigure}
     \hspace{0.25cm}
     \begin{subfigure}{.45\textwidth}
        \centering
        \includegraphics[width=1.0\linewidth]{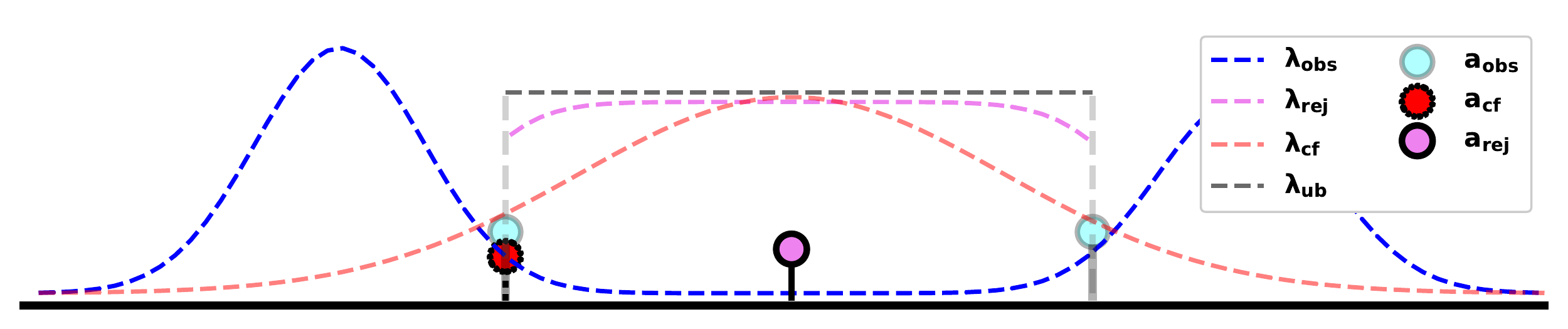}
        \caption{Step 4: Sample rejected events for the second interval.}
        \label{fig:step4}
        \includegraphics[width=1.0\linewidth]{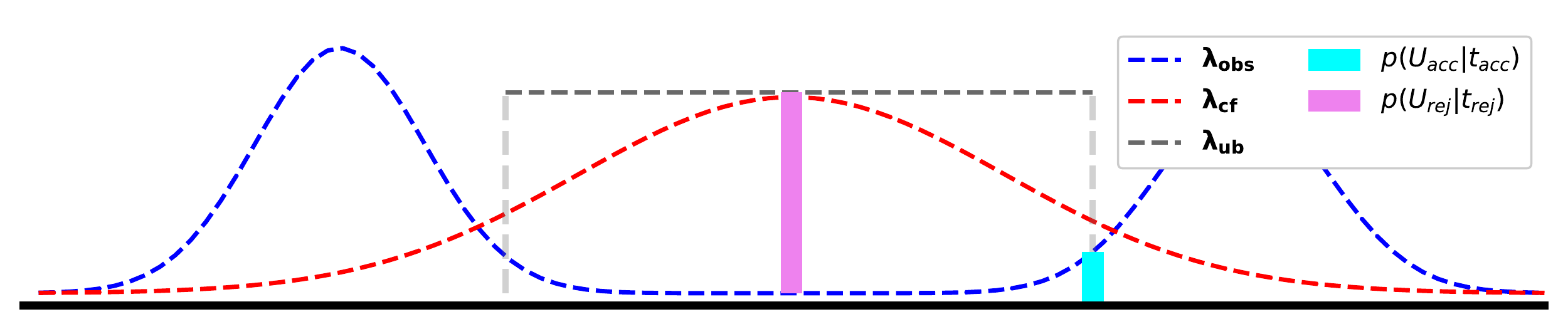}
        \caption{Step 5: Noise posteriors for the second interval.}
        \label{fig:step5}
        \includegraphics[width=1.0\linewidth]{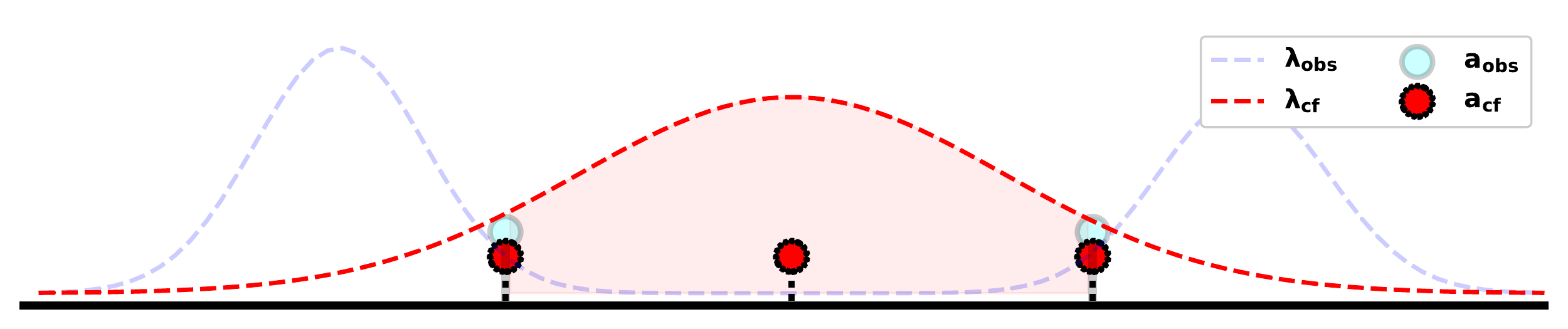}
        \caption{Step 6: Re-thin using the noise posteriors and the counterfactual intensity for the second interval.}
        \label{fig:step6}
     \end{subfigure}
     \begin{subfigure}{.9\textwidth}
        \centering
        \includegraphics[width=0.5\linewidth]{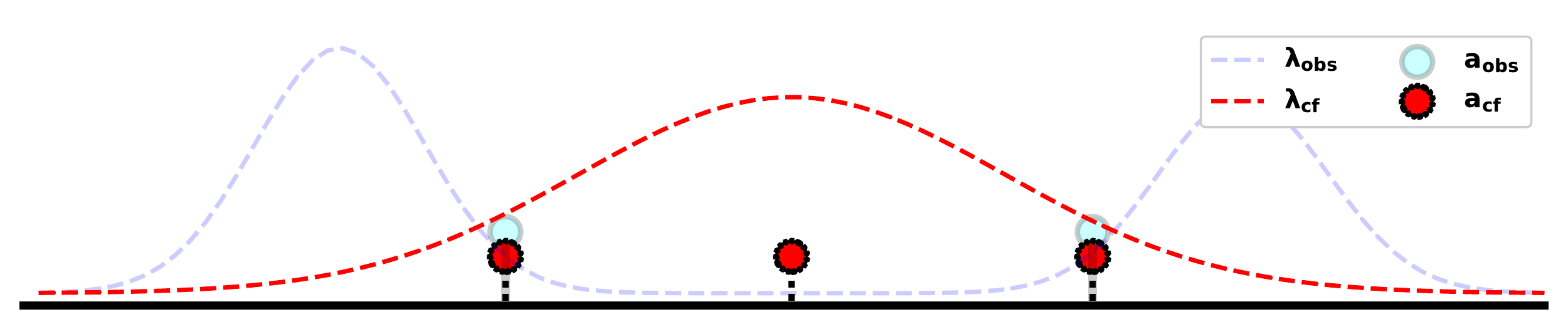}
        \caption{Final: Counterfactual treatments.}
        \label{fig:step7}
     \end{subfigure}
    \caption{The algorithm is illustrated on a toy example. \textbf{(a-h)} The observed ({\color{cyan}cyan}), rejected ({\color{pink}pink}) and counterfactual ({\color{red}red}) points are shown by filled circles. The observational ({\color{blue}blue}), rejected ({\color{pink}pink}) and counterfactual ({\color{red}red}) intensities are shown in dashed lines. The noise posteriors for the accepted points ({\color{cyan}cyan}) and the rejected points ({\color{pink}pink}) are shown by filled rectangles. \textbf{(a)} The initial setup with observed treatments, the observational and the counterfactual intensities. \textbf{(b-d)} We demonstrate the three tasks for the first interval, which is up until the first observed point. \textbf{(b-d)} The same procedure for the second interval, which is between two observed points. \textbf{(h)} The final counterfactual treatments. Notice how the algorithm chooses to keep the observed treatments where the counterfactual intensity is larger than the observed intensity.}
    \label{fig:counterfactual-sampling}
\end{figure}

The algorithm performs three main tasks, for each interval $[\tau, \tau+l(\tau)]$. This procedure is described below in words and illustrated in \cref{fig:counterfactual-sampling}.
\begin{enumerate}
    \item (\cref{fig:step1,,fig:step4}) The algorithm finds the upper-bound intensity $\lambda_{ub}$ for the interval $[\tau, \tau+l(\tau)]$ and it samples rejected events by using the difference of the upper bound intensity and the observational intensity: $\lambda^*_{rej} = \lambda_{ub} - \lambda^*_{obs}$.
    \item (\cref{fig:step2,,fig:step5}) It computes the posterior distribution of each acceptance noise variable $U_i$ in the interval $[\tau, \tau+l(\tau)]$, e.g. a rejected point $t_{rej}$ or an observed point $t_{obs}$. By inspecting the sampling process in Algorithm \ref{alg:ogata}, we can compute the noise posterior for a noise variable $U_i$ as follows:
    \begin{align*}
        p(U_i \mid t_i, \lambda_{ub}, \lambda^*_{obs}) = \begin{cases}
          \mathcal{U}(0, \lambda^*_{obs}(t_i)), & \text{if}\ t_i\ \text{is observed}, \\
          \mathcal{U}(\lambda^*_{obs}(t_i), \lambda_{ub}), & \text{if}\ t_i\ \text{is rejected}.
        \end{cases}
    \end{align*}
    \item (\cref{fig:step3,,fig:step6}) Then, it uses the counterfactual intensity and the noise posteriors to re-thin the candidate accepted and rejected points.
\end{enumerate}

As a result, once we have access to (i) observed (accepted) events, (ii) rejected events, (iii) posterior distributions of noise variables, (iv) the observational intensity and (v)  the counterfactual intensity, we can sample point process realizations from the counterfactual distribution, in a similar fashion to the forward sampling process of Ogata's algorithm. The counterfactual sampling algorithm is presented in Algorithm \ref{alg:counterfactual_tpp}. There are three key differences to the observational sampling process in Algorithm \ref{alg:ogata}: (i) in each interval, the algorithm samples hypothetical rejected events using the rejection intensity $\lambda_{ub}-\lambda^*_{obs}$ (Line 5), (ii) the counterfactual algorithm uses the noise posteriors ($p(U_i \mid t_i)$) instead of noise priors $\mathcal{U}(0, \lambda_{ub})$ (Lines 7,15) and (3) the accept/reject decisions are based on the counterfactual intensity $\lambda^*_{cf}$ instead of the observational intensity $\lambda^*_{obs}$ (Line 8,16).

\subsubsection{Identifiability of the Counterfactual Query}
\label{sssec:cf-id}

In this section, we first describe the monotonicity condition for binary \textsc{Scm}s. Then, we show that our counterfactual \textsc{Tpp} algorithm satisfies the monotonicity condition and hence the counterfactual identifiability. We conclude by discussing the identifiability of the counterfactual outcome trajectory.

Consider an \textsc{Scm} with a binary intervention variable $\pi$ and a binary target variable $A$. For the binary \textsc{Scm}, the monotonicity condition \citep{pearl2009causality, oberst2019counterfactual, noorbakhsh2021counterfactual} is a sufficient condition to render the counterfactuals of the target variable $A$ identifiable. It is stated formally as:

\textbf{Monotonicity \citep{noorbakhsh2021counterfactual}.}  The binary target variable $A$ is monotonic with respect to the binary intervention variable $\pi$, if the following condition holds for $a \neq a'$: 
\begin{align}
    \label{eq:monotonicity}
    \EX[A=a \mid do(\pi=1)] \geq \EX[A=a \mid do(\pi=0)] &\implies P(A=a' \mid A=a,\pi=0,do(\pi = 1)) = 0.
\end{align}
The monotonicity condition states that an intervention which increases the probability of an already observed event $A=a$ can not produce a counterintuitive counterfactual value $A=a'$.

Let us denote the observational regime by $\pi=0$ and the interventional regime by $\pi=1$, such that $\pi \in \{0,1\}$. Similarly, we denote their corresponding intensities by $\lambda^*_0$ and $\lambda^*_1$. For a given point at time $t$, the probability of acceptance $A$ under a regime $\pi \in \{0,1\}$ is proportional to its intensity and the upper bound, e.g., for $\pi=0$: 
$$p(A \mid t,\lambda_{ub}, do(\pi=0)) = p(A \mid t,\lambda_{ub},\pi=0) = \frac{\lambda^*_0(t)}{\lambda_{ub}}.$$

Without loss of generality, we consider a single time point $t$ for our algorithm. An input point to the algorithm at time $t$ is either an accepted ($A=1$) or a rejected ($A=0$) point:
\begin{itemize}
    \item For a rejected point $t$ with $A=0$, the probability of the point to be kept as a counterfactual point is zero, after an intervention that decreases the intensity at time $t$:
    \begin{align}
        \label{eq:cf-cond1}
        \EX[A=0 \mid do(\pi=1)] \geq \EX[A=0 \mid do(\pi=0)] \implies P(A=1 \mid A=0, \pi=0, do(\pi=1)) = 0.
    \end{align}
    This follows from the posterior distribution of the rejected point $U_{rej} \sim \mathcal{U}(\lambda^*_0(t), \lambda_{ub})$ being larger than the interventional intensity $\lambda^*_1(t) \leq \lambda^*_0(t)$.
    \item For an accepted point $t$ with $A=1$, the probability of the point to be removed from the counterfactual set is zero after an intervention that increases the intensity at time $t$:
    \begin{align}
        \label{eq:cf-cond2}
        \EX[A=1 \mid do(\pi=1)] \geq \EX[A=1 \mid do(\pi=0)] \implies P(A=0 \mid A=1, \pi=0, do(\pi=1)) = 0.
    \end{align}
    This follows from the posterior distribution of the accepted point $U_{acc} \sim \mathcal{U}(0, \lambda^*_0(t))$ being smaller than the interventional intensity $\lambda^*_1(t) \geq \lambda^*_0(t)$.
\end{itemize}
\cref{eq:cf-cond1,,eq:cf-cond2} imply that the counterfactual sampling algorithm satisfies the monotonicity condition in \cref{eq:monotonicity}, and hence the counterfactual treatments are identifiable.

The counterfactual algorithm either adds or removes a treatment to the observed treatment sequence, which produces a new or removed bump in the outcome trajectory, formalized by the treatment response function $f_a$, while the baseline function $f_b$ remains unchanged. Therefore, the counterfactual outcome trajectory is identifiable, conditioned on the identifiability of the treatment response function $f_a(\tau;\mathbf{a})$, which is not possible without further functional assumptions. In our semi-synthetic experiments, we empirically show that it is possible to identify the treatment response function using distinctive functional priors for the baseline and the treatment response, when the outcome trajectory follows our model definition. When this is the case, the counterfactual trajectory can be empirically estimated with good accuracy.

\section{Experiment Details}

\subsection{Real-World Study Details}
\label{ssec:real-world}

\subsubsection{Data Preprocessing}

The real-world data set consists of meal-blood glucose measurements for 14 non-diabetic individuals, where individuals are monitored over a 3-day period \citep{zhang2020errors}. The blood glucose is measured by a continuous monitoring device, which takes a sample approximately every 15 minutes. The meal data is collected through a meal diary, recorded daily by individuals. Later, meal records in the diary are translated into nutrient values by a look-up table.

As the meal data set is collected by individuals, it inherently contains measurement errors \citep{zhang2020errors}. To deal with the measurement errors, we take a data preprocessing step before fitting our treatment model to the meal data. For our application, there are two main measurement errors that may lead to incorrect results: (i) individuals may record their meals in parts, where a single meal is recorded as 2-3 meal occurrences, and (ii) individuals record their meal times with error. 

To deal with measurement errors of type (i), we remove redundant meals by keeping only a single meal event if two meals co-occur in a 2-hour interval. To deal with measurement errors of type (ii), we update the meal time if a meal seems to occur at a time where the blood glucose is changing rapidly. For example, if an individual reports a meal with a delay, the meal seems to occur at a time point where the blood glucose measurements have already increased rapidly and is about to decline. The recorded meal time (with error) implies an incorrect result that the meal leads to a decrease in the blood glucose, or similarly, that the meal likelihood increases with increasing blood glucose. On the contrary, the bump in the glucose level is most likely caused by the same meal in the first place. In practice, we move a meal time to the next possible time point where the blood glucose derivative is less than a threshold, $0.5$. At each time point $t_i$, we approximate the derivative of the blood glucose by averaging derivative values for two consecutive regions, $(t_{i-1}, t_i)$ and $(t_{i}, t_{i+1})$:
$$dy_i \approx \frac{1}{2}\left(\frac{(y_i - y_{i-1})}{t_i - t_{i-1}} + \frac{(y_{i+1} - y_i)}{t_{i+1} - t_i}\right).$$

\subsubsection{Joint Model Definition}
\label{subsec:real_world_model}

\paragraph{Treatment Intensity.} We assume each patient's meal habits are different, corresponding to a distinct treatment policy. Furthermore, we define five model variants $\{ \lambda_{b}, \lambda^*_{ba}, \lambda^*_{bo}, \lambda^*_{ao}, \lambda^*_{bao} \}$ for each patient (see \cref{tab:testll} for details). Each treatment model is defined and trained independently on each patient's meal--glucose measurements. 

We choose the constant baseline intensity $\beta_0$ to be equal to $0.1$. For the treatment-dependent function $g^*_a(\tau; \mathbf{a})$, we use a relative-time kernel $k_t$ only, i.e. we assume meal dosages (marks) do not have an affect on the next meal time. For the outcome-dependent function $g^*_o(\tau; \mathbf{o})$, we use a combination of a mark kernel $k_m$ and a relative-time kernel $k_t$. We choose the dimensionality parameters $\{Q_a, Q_o\}$ of the kernels as $Q_a=1, Q_o=1$, i.e., the meal intensity depends only on the last treatment and the last outcome observations. This is a simplification of complex real-world dynamics between meals and blood glucose. Yet, our goal is not to model these dynamics in the best possible way in this work, but rather to understand if our joint model can learn clinically meaningful treatment intensities, that can handle time-varying confounding.

At a query time $\tau$, the retrieval function $r(\tau)$ returns the input vector $r(\tau) = \mathbf{r} \in \mathbb{R}^{4}$, as defined in \cref{subsec:treatment_kernel}:
$$
r(\tau) = \{\underbrace{\tau}_{\substack{\text{Input for} \\ \text{Baseline}}}, \underbrace{\tau - t_i}_{\substack{\text{Input for} \\ \text{last } Q_a \\ \text{ treatment}}}, \underbrace{\tau - t_j, y_j}_{\substack{\text{Input for} \\ \text{last } Q_o \\ \text{ outcome}}}\}.
$$
The intensities $\{ \lambda_{b}, \lambda^*_{ba}, \lambda^*_{bo}, \lambda^*_{ao}, \lambda^*_{bao} \}$ use the relevant dimensions of the input. For example, the intensity $\lambda_{b}$ uses only the absolute time information (first dimension) $\tau$, which is for the baseline input.

We choose $M=20$ inducing points: $\mathbf{Z} \in \mathbb{R}^{M \times 4}$. We place the inducing points in regular intervals inside the target domain of the treatment intensity component $g_b$, $g^*_a$ or $g^*_o$. For example, the baseline component's input domain is the observation period $[0, T]$ and inducing points for the baseline dimension are placed onto $[0,T]$ regularly. Similarly, input domains of treatment-effect and outcome-effect kernels include relative times to the last $Q_a$ actions and $Q_o$ outcomes. For these two dimensions, we place inducing points between 0 and the maximum (or some quantile) time between the last $Q_a$ actions or $Q_o$ outcomes. Inducing points are assumed to be independent along each dimension similar to \citet{liu2019nonparametric}.

We choose kernel hyperparameters $\theta_{a,h}=\{\gamma_q,\ell_q\}_{q=0}^{Q_a+Q_o}$ by inspection, since our data set is small and there exists identifiability issues when one optimizes the effect of each component on the treatment effect simultaneously. As the number of meal events are small in the data set, we choose relatively small values for the variance parameters $(\gamma_b, \gamma_a, \gamma_o) = (0.1, 0.05, 0.15)$. We choose lengthscale parameters as $(\ell_b, \ell_a, \ell_{o,t}, \ell_{o,m}) = (7.0, 1.0, 100.0, 2.5)$. Each lengthscale value represents an underlying assumption based on the domain knowledge: (i) the baseline lengthscale $\ell_b = 7.0$ assumes a slow-changing smooth baseline function, (ii) the treatment-effect time lengthscale $\ell_a$ is $1.0$ as the last meal generally has an effect on the next meal in the next 2-3 hours time span, (iii) a very large outcome-effect time lengthscale $\ell_{o,t}=100$ is chosen so that the outcome-effect is constant in time for a given glucose value $y$, mimicking a simultaneous outcome-effect on the meal intensity with a large number of regular glucose measurements, (iv) the outcome-effect mark lengthscale $\ell_{o,m}=2.5$ is chosen so that the mark effect is also a slow-changing, smooth function, as the glucose level should gradually affect meal intensity.

\paragraph{Outcome Model.} We define a single hierarchical outcome model for all patients, as detailed in \cref{sec:outcome-details}. 

The baseline progression $f_b$ for each patient $(v)$ is the sum of a constant kernel and a periodic kernel whose period is set to 24 hours. The periodic kernel captures daily glucose profiles of patients. The constant kernel function has an intercept parameter $\{b_b^{(v)}\}$, initialized to $\{1.0\}$. Each periodic kernel has parameters $\{\alpha^{(v)}_{b},\ell^{(v)}_{b}\}$, initialized to $\{1.0, 1.0\}$.

The treatment response function is equal to the sum of two components: $f_a(\tau; \mathbf{a}=(\mathbf{m}, \mathbf{t})) = f_m(\mathbf{m}) f_t(\tau; \mathbf{t})$. The response shape function $f_t(\tau; \mathbf{t})$ has an \textsc{Se} kernel with hyperparameters $\{\ell^{(m)}_{t}\}$ for each treatment type $(m)$. We assume all meals are of the same type: $m=1$. We initialize the lengthscale to $0.5$: $\ell^{(m)}_{t} = \ell_{t} = 0.5$. The effective interval $\mathcal{T}_{t_i}$ for each meal $t_i$ is set to 3 hours: $\mathcal{T}_{t_i} = 3$. The treatment response scaling function $f^{(v)}_m$ specifies the amplitude of the treatment response for each patient $(v)$, using patient-specific intercept and slope parameters: $\{\beta_0^{(v)}, \beta_1^{(v)}\}$. A hierarchical model is imposed on the intercept and the slope parameters $\{\beta_0^{(v)}, \beta_1^{(v)}\}_{v=1}^{14}$ by defining a hierarchical Gaussian prior on them: $\beta_0^{(v)} \sim \mathcal{N}(\beta_0, \sigma^2_0), \beta_0^{(v)} \sim \mathcal{N}(\beta_1, \sigma^2_1), \forall v \in \{1, \ldots, 14\}$. All scaling function parameters $\{\{\beta_0^{(v)}, \beta_1^{(v)}\}_{v=1}^{14}, \beta_0, \beta_1, \sigma_0, \sigma_1\}$ are initialized to 0.1.

\paragraph{Training.} We assume each day of meal observations as conditionally independent given the conditional intensity function. We train each model on the first two days of observations and use the third day as the test set. For the outcome model, the measurements follow a Gaussian likelihood, i.e., an exact \textsc{Gp} model. We maximize the log marginal likelihood with respect to the hyperparameters $\theta_o = \{ \beta_0, \beta_1, \{b^{(v)}, \gamma^{(v)}_{b},\ell^{(v)}_{b}, \beta_0^{(v)}, \beta_1^{(v)} \}_{r=1}^{14}\}$. We fix hyperparameters $\{\sigma_0, \sigma_1\}$ at their initialized values. For the treatment model, we maximize the training objective \textsc{Elbo} with respect to the variational parameters $\theta_{a,z}=\{\mathbf{m_z}, \mathbf{S_z}\}$, while the treatment hyperparameters $\theta_{a,h}$ are fixed at their initial values.

\begin{table}[H]
  \small
  \caption{Results for the test log likelihood (\textsc{Tll}) values for treatment models with distinct meal intensities, for all patients. Models are trained on first 2-days of meal-glucose data. The last day is used for computing the \textsc{Tll}. The intensity $\lambda_{ba}^*(\tau)$ provide the highest \textsc{Tll} value for all patients.}
  \label{tab:testll_full}
  \centering
  \begin{tabular}{l|c|rrrrr}
    \toprule
     & & \multicolumn{5}{c}{\textsc{Meal Intensities}} \\
    \cmidrule(r){3-7}
    \textsc{Patient-id} & \textsc{Metric} & $\lambda_{b}^*$ & $\lambda_{ba}^*$ & $\lambda_{bo}^*$ & $\lambda_{ao}^*$ & $\lambda_{bao}^*$ \\
    \midrule
    \textsc{Patient} 1 & \multirow{14}{*}{\textsc{Tll}} 
    & $-10.38$ & $-9.89$ & $-9.84$ & $\mathbf{-7.68}$ & $-10.68$ \\
    \textsc{Patient} 2 & & $-14.37$ & $-11.39$ & $\mathbf{-10.57}$ & $\mathbf{-10.58}$ & $-11.85$ \\
    \textsc{Patient} 3 & & $-12.07$ & $-12.85$ & $\mathbf{-12.05}$ & $-12.78$ & $-12.79$ \\
    \textsc{Patient} 4 & & $-12.90$ & $-12.75$ & $\mathbf{-9.38}$ & $-9.61$ & $-11.52$ \\
    \textsc{Patient} 5 & & $-14.93$ & $-11.93$ & $\mathbf{-10.99}$ & $-11.12$ & $-12.35$ \\
    \textsc{Patient} 6 & & $-13.46$ & $-9.52$ & $\mathbf{-8.51}$ & $-8.59$ & $-9.73$ \\
    \textsc{Patient} 7 & & $-13.87$ & $-12.19$ & $-12.47$ & $\mathbf{-10.86}$ & $-13.98$ \\
    \textsc{Patient} 8 & & $-12.79$ & $-12.62$ & $-10.85$ & $\mathbf{-9.70}$ & $-12.20$ \\
    \textsc{Patient} 9 & & $-13.60$ & $-14.07$ & $-13.01$ & $\mathbf{-12.91}$ & $-13.95$ \\
    \textsc{Patient} 10 & & $-16.29$ & $-13.10$ & $\mathbf{-11.37}$ & $\mathbf{-11.39}$ & $-12.61$ \\
    \textsc{Patient} 11 & & $-13.49$ & $-12.95$ & $-12.33$ & $\mathbf{-12.10}$ & $-14.00$ \\
    \textsc{Patient} 12 & & $-15.75$ & $-13.82$ & $\mathbf{-12.07}$ & $-13.61$ & $-14.79$ \\
    \textsc{Patient} 13 & & $\mathbf{-9.71}$ & $-10.81$ & $-10.07$ & $-9.99$ & $-11.67$ \\
    \textsc{Patient} 14 & & $-12.08$ & $-12.50$ & $-10.95$ & $\mathbf{-9.13}$ & $-12.88$ \\
    \midrule
    \textsc{All} & \textsc{Tll} & $-13.26 \pm 0.49$ & $-12.17 \pm 0.36$ & $-11.03 \pm 0.34$ & $-10.72 \pm 0.46$ & $-12.50 \pm 0.37$ \\
    \bottomrule
  \end{tabular}
\end{table}

\begin{figure}[h]
    \centering
    \includegraphics[width=\linewidth]{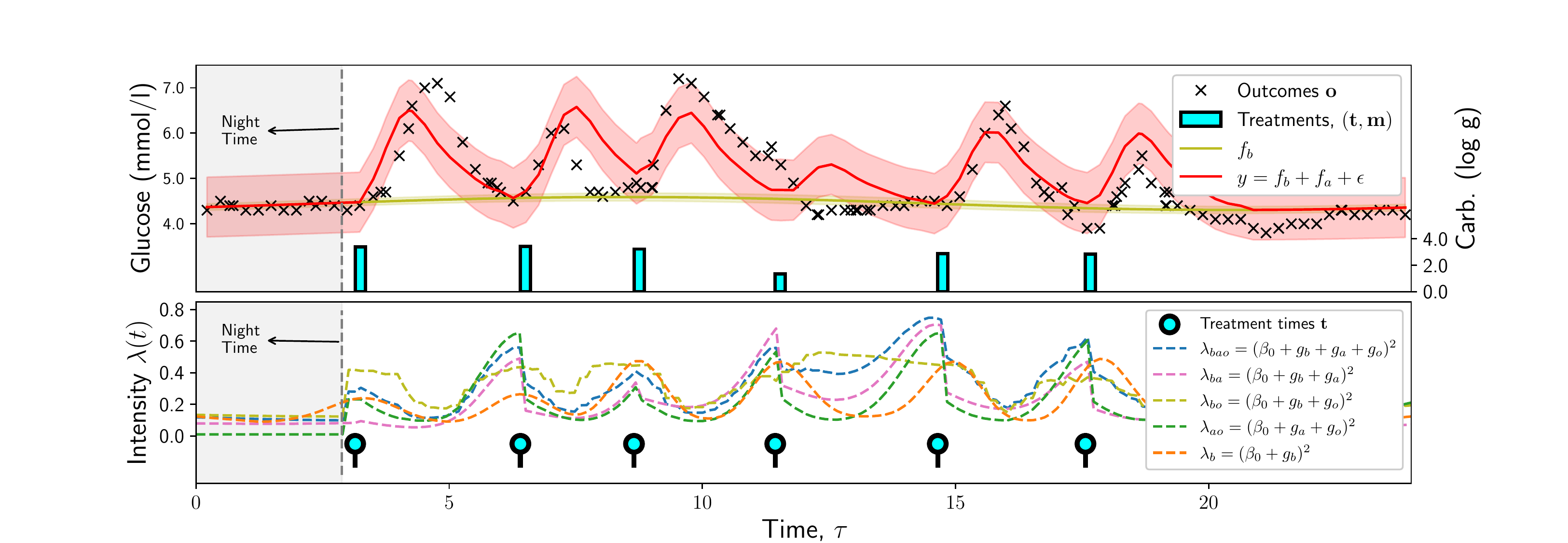}
    \caption{Estimated treatment--outcome model on the real-world meal--glucose data, similar to \cref{fig:treatment_model_fit}. \textbf{(Top)} Estimated outcome model on the glucose measurements (see \cref{fig:treatment_model_fit} for details). \textbf{(Bottom)} Estimated intensities for all five treatment models $\{ \lambda_{b}, \lambda^*_{ba}, \lambda^*_{bo}, \lambda^*_{ao}, \lambda^*_{bao}\}$ are shown. Inspecting intensities $\lambda^*_{ba}$ ({\color{pink}pink}), $\lambda^*_{ao}$ ({\color{teal}green}) and $\lambda^*_{bao}$ ({\color{blue}blue}), we see that the intensity of a new meal immediately after a previous meal decreases through $g^*_a$. Similarly, inspecting intensities $\lambda^*_{bo}$ ({\color{olive}yellow}), $\lambda^*_{ao}$ ({\color{teal}green}) and $\lambda^*_{bao}$ ({\color{blue}blue}), we see that the intensity of a new meal immediately after a previous meal decreases indirectly through the increase in blood glucose represented by $g^*_o$.}
    \label{fig:treatment_model_fit_full}
\end{figure}

\paragraph{Results.} For the treatment model, we report the test log-likelihood (\textsc{Tll}) for the third day to compare different treatment intensity variants. The computation of \textsc{Tll} is detailed in \cref{subsec:tll}. \textsc{Tll} results for all patients are shown in Table \ref{tab:testll_full}. In agreement with the mean \textsc{Tll} value, the intensity $\lambda^*_{ao}$ produce largest \textsc{Tll} values for most patients. In addition, the simplest (baseline) intensity $\lambda_{b}$ with no past treatment-outcome effect and the most complex intensity $\lambda^*_{bao}$ with all components provide low \textsc{Tll} values in general. This suggests that a treatment model independent of the past history is not able to explain the meal intensity well. Similarly, the intensity with all components seem to be overly complex for this small data set. In addition, we demonstrate the model fits for all intensities in \cref{fig:treatment_model_fit_full}. 

\begin{figure}[h]
    \centering
    \includegraphics[width=\linewidth]{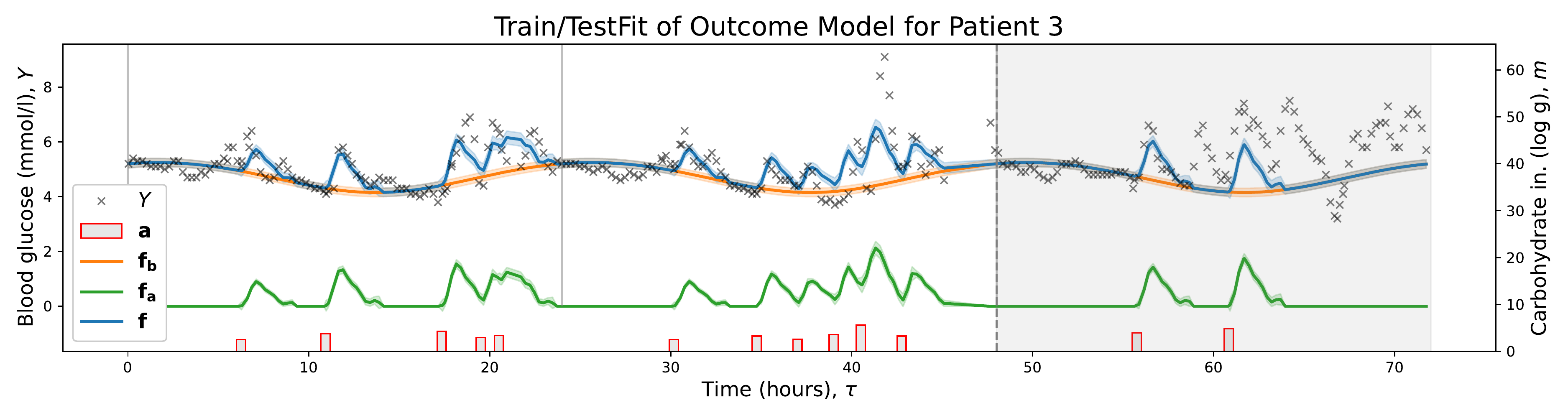}
    \includegraphics[width=\linewidth]{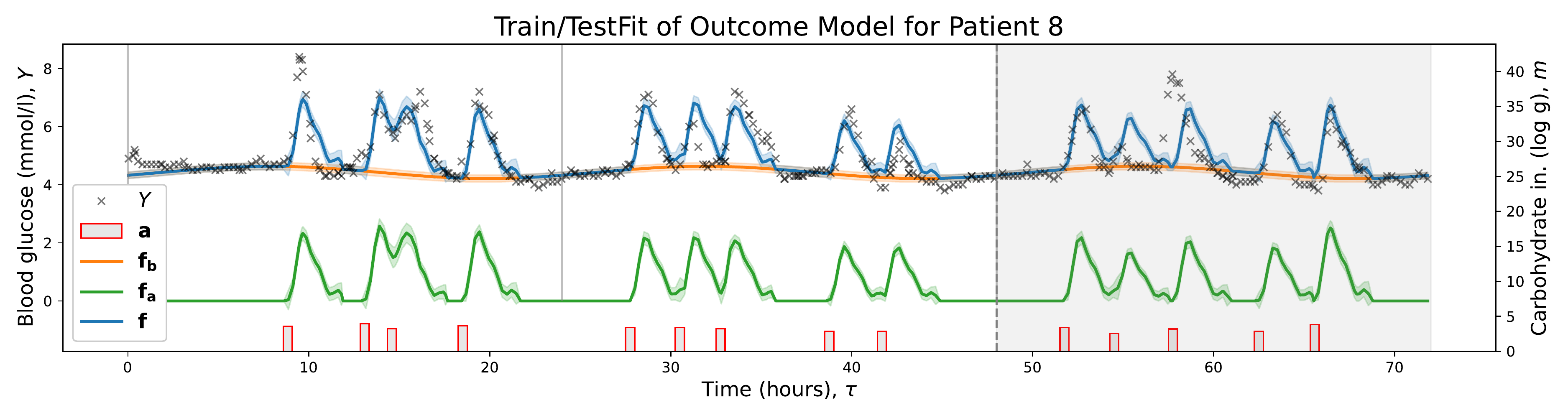}
    \includegraphics[width=\linewidth]{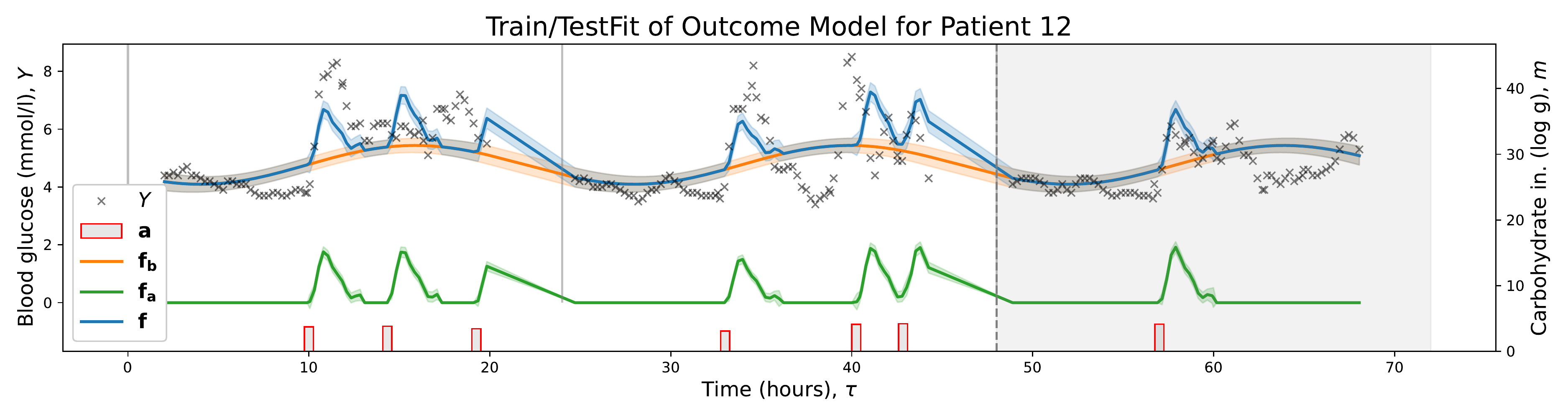}
    \caption{The learned ground-truth functions $f_b(\tau)$, $f_a(\tau)$ and $f(\tau)=f_b(\tau)+f_a(\tau)$ for Patients \{3,8,12\}. The grey-shaded area is the test period. The baseline kernels are sums of (i) a constant kernel \citep{ashrafi2021computational} and (ii) a periodic kernel (combined with a long-lengthscale \textsc{Se} kernel) whose period is fixed to 1-day in order to capture the daily blood glucose profiles. The treatment shape function is close to a Gaussian bump \protect\citep{wyatt2021postprandial}.}
    \label{fig:gt-f_outcome}
\end{figure}

For the outcome model, we examine train and test fits for patients qualitatively. We show train/test fits for a subset of patients $\{3,8,12\}$ in \cref{fig:gt-f_outcome}. The empirical findings suggest that the outcome model is able to capture clinically meaningful treatment responses that are close to bell-shaped curves reported in \citet{wyatt2021postprandial}. It is important to note that the treatment response curves are the main learning objective in continuous-time treatment--outcome setups \citep{schulam2015framework, xu2016bayesian, schulam2017reliable, soleimani2017treatment, cheng2020patient, seedat2022continuous}.

\begin{figure}[h]
    \centering
    \begin{subfigure}{.33\textwidth}
        \centering
        \includegraphics[width=\linewidth]{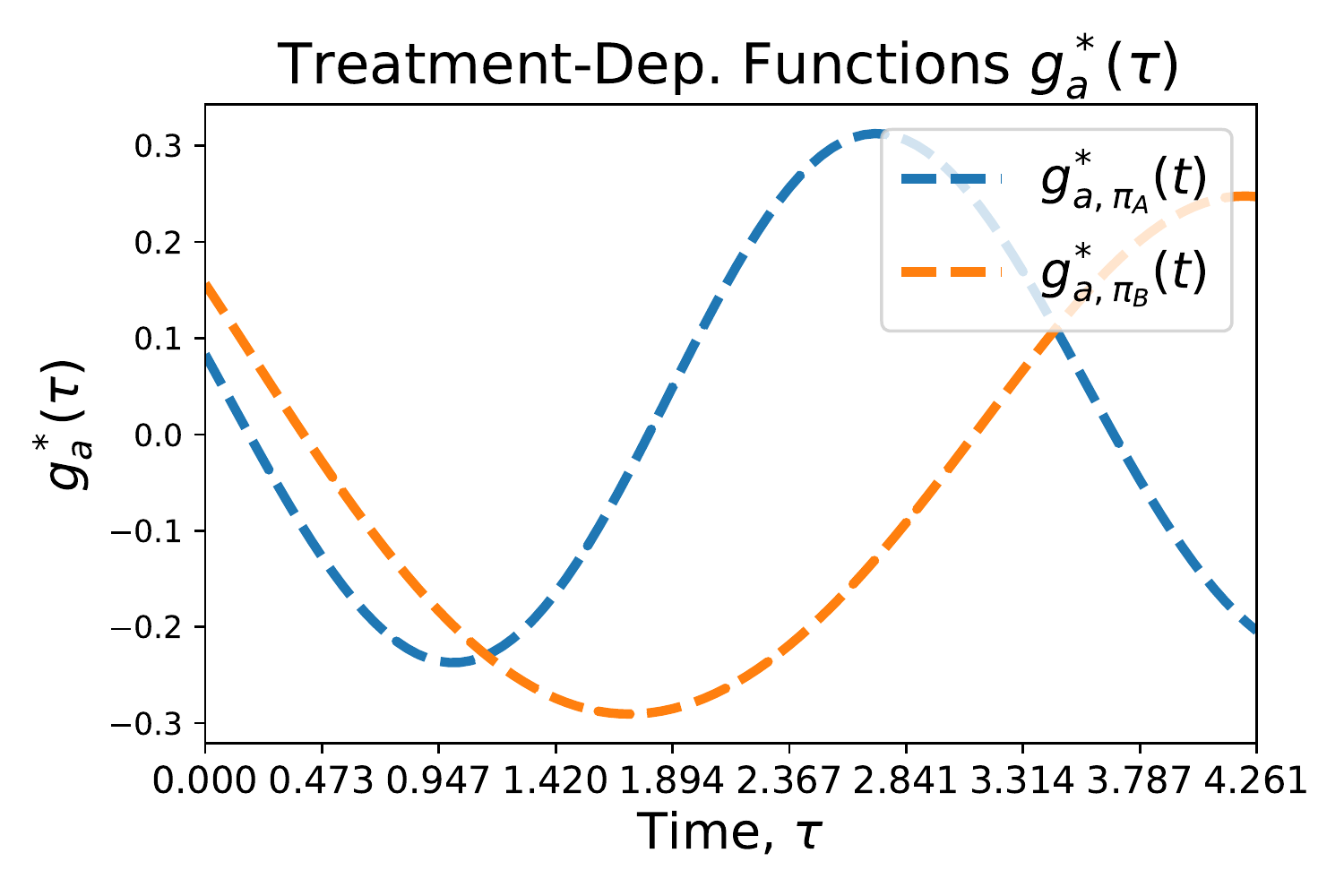}
        \caption{Dependence on last meal $g^*_a$.}
        \label{fig:lastmealdep}
     \end{subfigure}
     \begin{subfigure}{.33\textwidth}
        \centering
        \includegraphics[width=\textwidth]{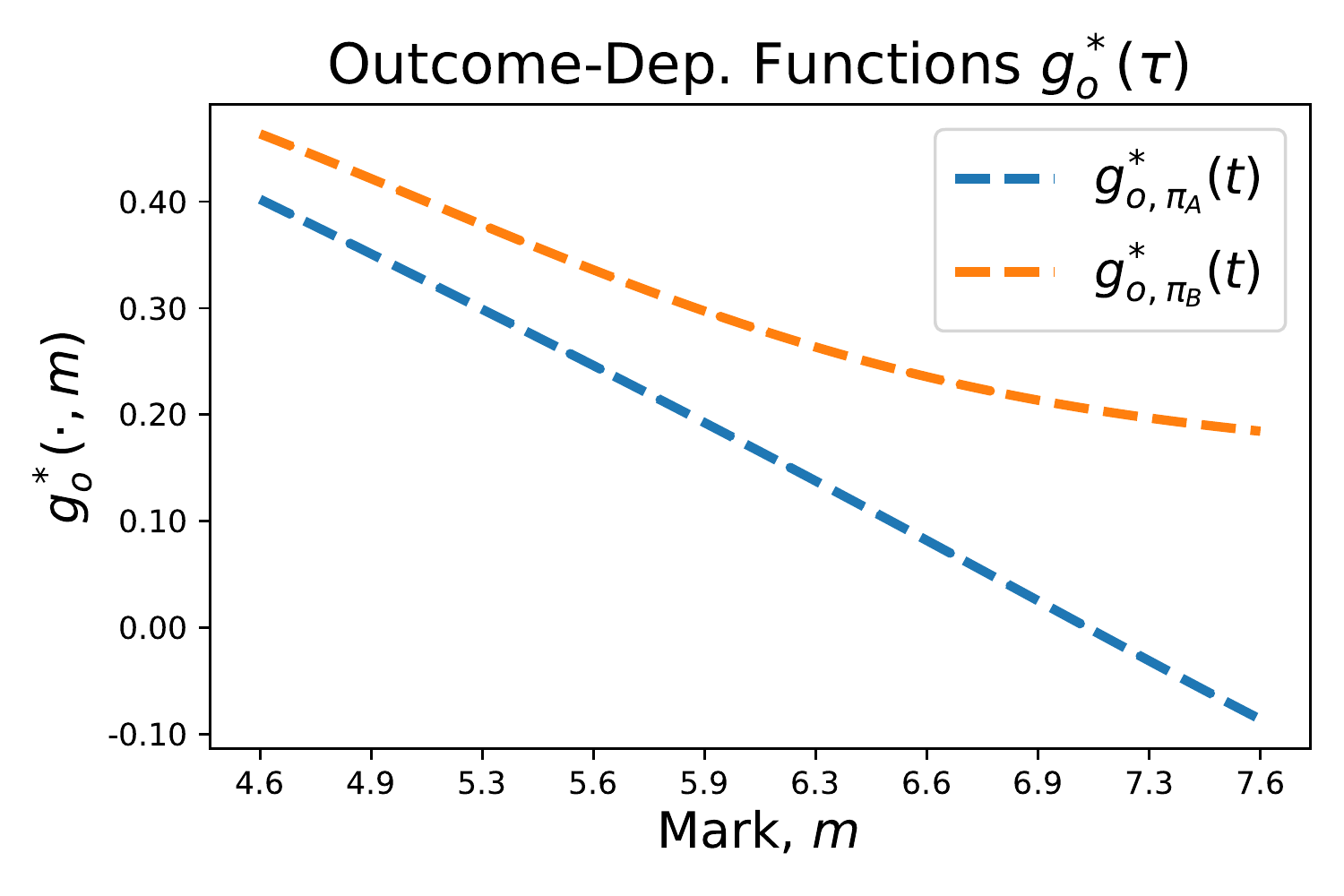}
        \caption{Dependence on last glucose value $g^*_o$.}
        \label{fig:lastoutdep}
     \end{subfigure}
     \begin{subfigure}{.33\textwidth}
        \centering
        \includegraphics[width=\textwidth]{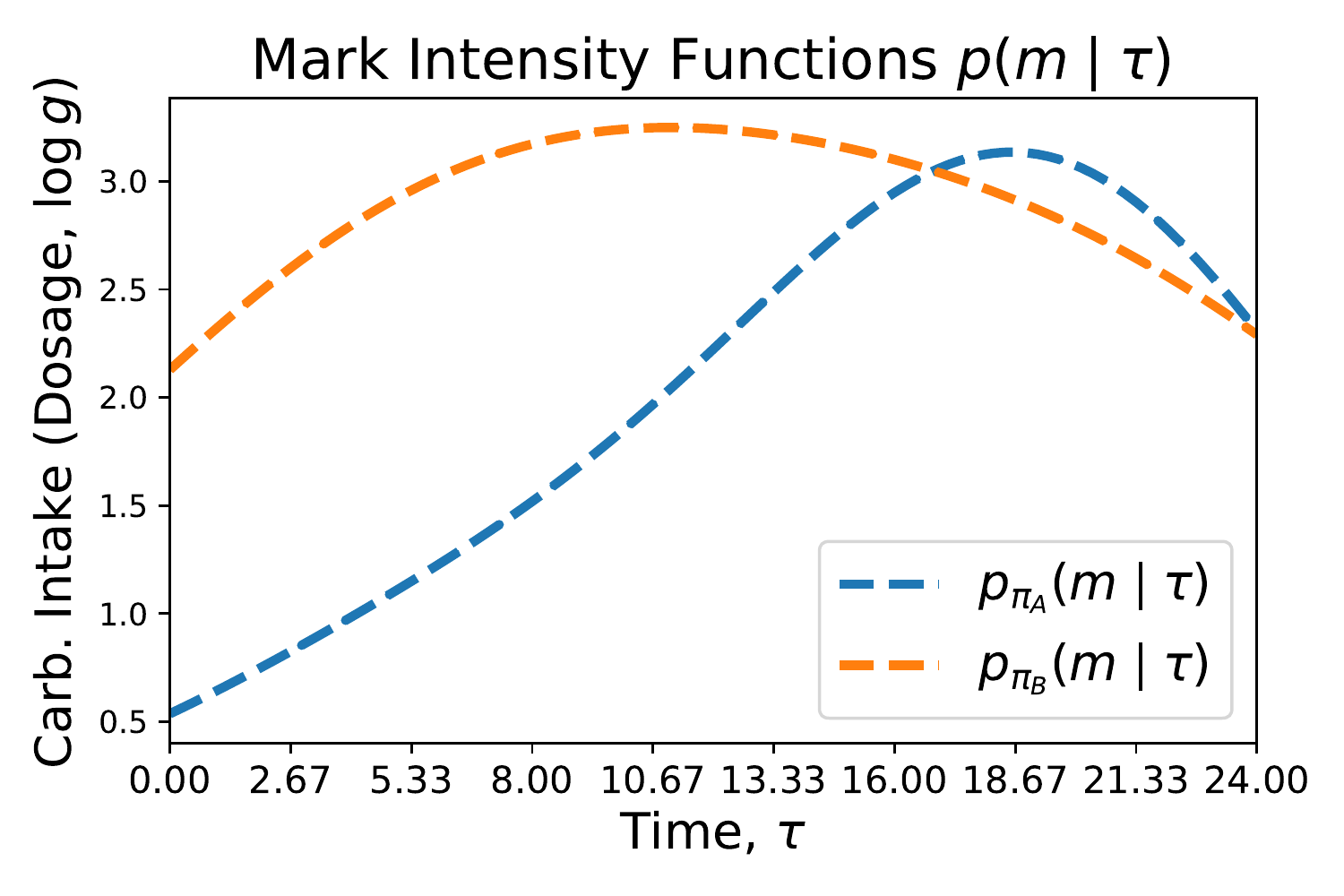}
        \caption{Carb. intake intensity $p(m\mid \tau)$.}
        \label{fig:carbint}
     \end{subfigure}
    \caption{The learned simulator functions functions $g^*_a$, $g^*_o$ and $p(m \mid \tau)$ for Patient 4 ({\color{blue}blue dashed lines}) and Patient 13 ({\color{orange}orange dashed lines}). \textbf{(a)} After a meal occurs, both intensities first decline and then increase to account for the next meal. The bump for the next meal occurs earlier for Patient 4 than Patient 13. Additionally, the magnitude of the bump is higher for Patient 4 than Patient 13. \textbf{(b)} The meal intensity of Patient 13 increases slightly more than Patient 4 for all blood glucose values. \textbf{(c)} Patient 13 is more likely to have meals with higher carbohydrate intake during early hours of the day, while Patient 4 is more likely to have higher carbohydrate intake after 5pm.}
    \label{fig:gt-gao}
\end{figure}

\subsection{Semi-Synthetic Simulation Study Details}

We perform a semi-synthetic simulation study to evaluate our model's performance on the causal tasks, as the true causal effects are unknown in a real-world observational data set.

\subsubsection{Simulator} 
\label{sssec:simulator}

\paragraph{Simulator functions.} Our ground-truth joint simulator is composed of a treatment and an outcome simulator. The ground-truth treatment simulator, has the form of a treatment- and outcome-dependent intensity $\lambda^*_{ao}$ (defined in \cref{tab:testll}) to include both sources of time-varying confounding. The ground-truth outcome simulator has the form of the hierarchical outcome model (defined in \cref{eq:outcome}). The model definitions, initialization and the training procedures are detailed in \cref{subsec:real_world_model}. 

We fit our joint model to a subset of the real-world meal--glucose data set to obtain ground-truth simulator functions $\mathcal{M}^{(v)}_{gt} = \{\lambda^{*}_{\pi_v}(\tau), f^{(v)}_b(\tau), f^{(v)}_a(\tau; \mathbf{a})\}$. In the rest of this section, we show the characteristics of these functions and what kind of observational trajectories they produce.

\paragraph{Treatment simulator.} We divide simulated patients into two policy groups $\{\pi_A, \pi_B\}$, representing distinct treatment policies of different hospitals, countries, etc. The simulator intensities of two policies correspond to learned intensity functions of two real-world patients: Patients \{4,13\}. We choose these two intensities, so that the treatment distributions induced by two policies show different characteristics, i.e. observational and interventional distributions are not similar. Otherwise, an intervention on the treatment policy might not have a visible effect on the treatment distribution. For example, consider the limiting case of having the same treatment intensity for both policies $\{\pi_A, \pi_B\}$, then the observational and the interventional distributions are identical.

Comparing simple statistics of two patients, Patient 4 has 5 meals on average per day, while Patient 13 has 4 meals on average per day, which leads to higher intensity values for Patient 4, and hence lower expected meal arrival times. Besides, the learned mark intensity functions $p(m \mid \tau)$ for both policies are distinct. The ground-truth functions $g^*_a$ $g^*_o$ and $p(m \mid \tau)$, that are learned from Patients \{4,13\} are shown in \cref{fig:gt-gao}.

\paragraph{Outcome simulator.} To enable individualization among the blood glucose dynamics of patients, we assume there are three patient groups $\{gr_0,gr_1,gr_2\}$. Patients in the same group share the baseline and treatment response functions, e.g., patients in the group $gr_1$ have the following functions: $\{f_{b,gr_1}, f_{a,gr_1}\}$. 

\begin{figure}[h]
    \centering
    \begin{subfigure}{.49\textwidth}
        \centering
        \includegraphics[width=\linewidth]{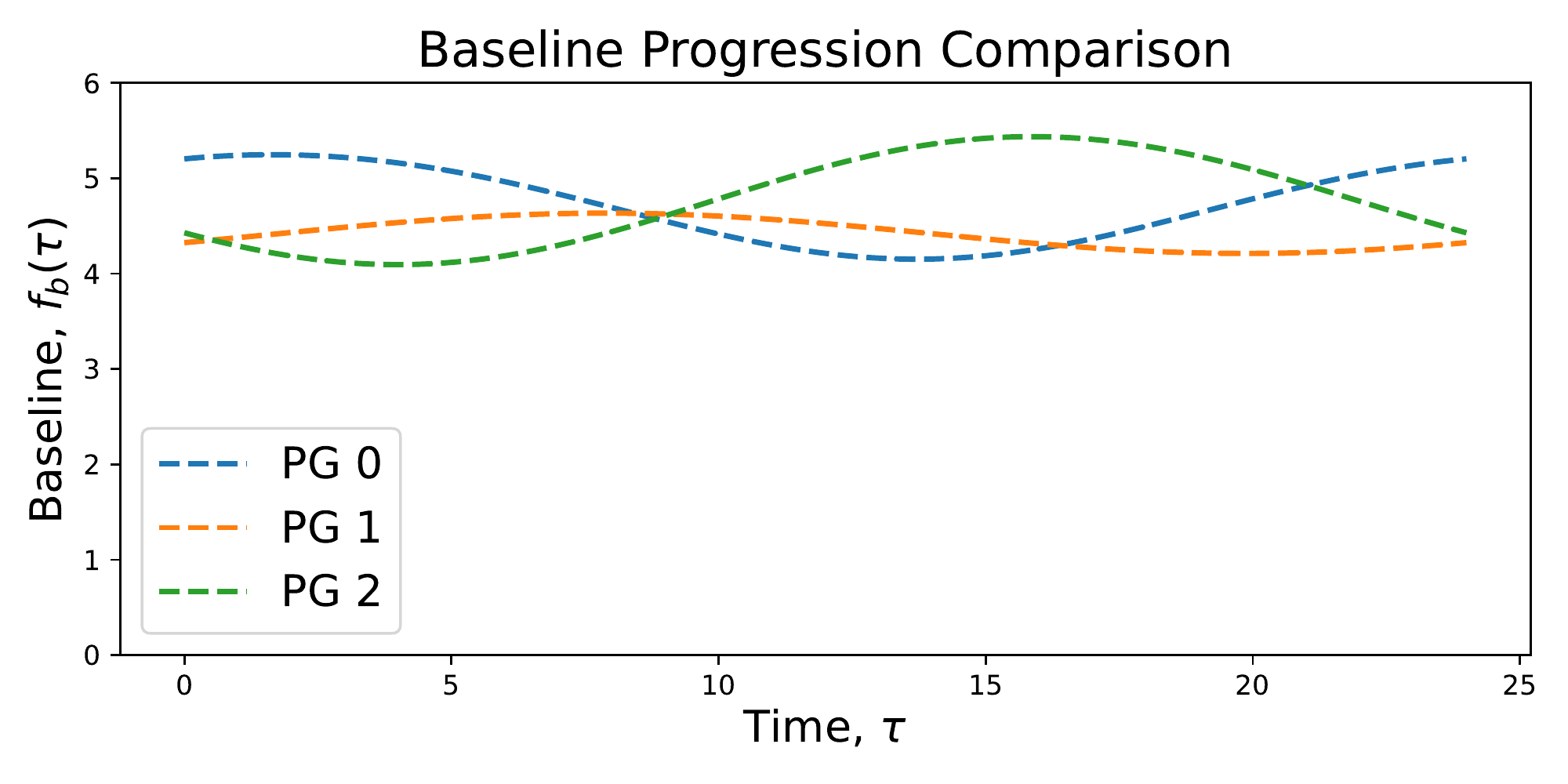}
        \caption{Baseline progressions $f_b$ in the interval $[0, 24]$ (hours).}
        \label{fig:simulator_base}
     \end{subfigure}
     \begin{subfigure}{.49\textwidth}
        \centering
        \includegraphics[width=\textwidth]{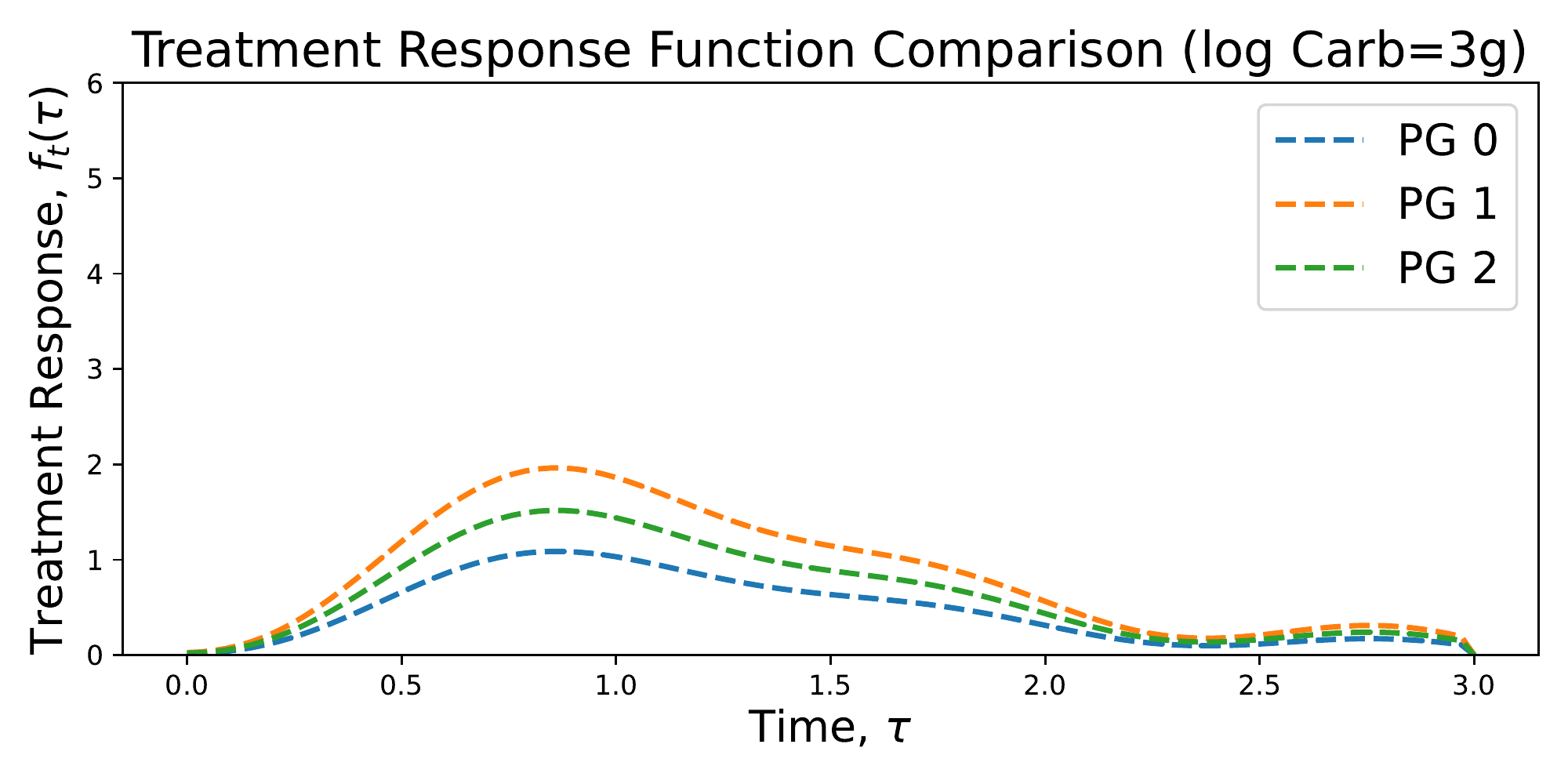}
        \caption{Treatment response curves $f_a$ in the interval $[0, 3]$ (hours).}
        \label{fig:simulator_trc}
     \end{subfigure}
    \caption{Baseline progression and treatment response functions of Patient 3 ({\color{blue}blue}), Patient 8 ({\color{orange}orange}) and Patient 12 ({\color{teal}green}), i.e., of the ground-truth outcome simulator.}
    \label{fig:gt-out}
\end{figure}

We choose three outcome simulator functions corresponding to the learned outcome models of three real-world patients (Patients \{3,8,12\}), by qualitatively assessing the learned functions with respect to the domain knowledge and findings about the effect of a meal on the blood glucose in \citet{wyatt2021postprandial}: (i) they have distinct daily baseline characteristics, fluctuating around a constant value (\cref{fig:simulator_base}) \citep{ashrafi2021computational} (ii) their treatment response curves are smooth functions with similar shapes to the bell-shaped response curves argued in \citet{wyatt2021postprandial} (\cref{fig:simulator_trc}) . The train and test fits of the outcome simulators for these patients are shown in Figure \ref{fig:gt-f_outcome}.

\subsubsection{Benchmarks}

For our estimation model, we use a joint model that has the same form with the simulator. In addition to ablations, we use two benchmark models used for comparison in Experiments:

\paragraph{\textsc{H21} \citep{hua2021personalized}.} They propose a joint treatment--outcome model, with a publicly available implementation. However, they assume treatments and outcomes occur jointly at the same continuous-time points, which is a strong assumption that does not hold for our problem definition. Therefore, we adapt their implementation by separating the joint likelihood into treatment and outcome parts, so that the model can work with treatments and outcomes that occur at different time points. Besides, we remove the irrelevant survival outcome part from the joint likelihood. The implementation uses a Markov chain Monte Carlo (\textsc{Mcmc}) algorithm. We run the algorithm for 20000 iterations, with a burn-in period of 5000 iterations. To deal with sample correlation, we use every $50^{\text{th}}$ sample after the burn-in period for inference. During inference, we use the mean values for the treatment and outcome model parameters.

\paragraph{\textsc{S17} \citep{schulam2017reliable}.} The outcome model proposed in \citet{schulam2017reliable} is a conditional \textsc{Gp}, with additive baseline and treatment response functions. We use their publicly available implementation that uses (i) the sum of a mixture of B-splines and a \textsc{Gp} prior with a Matern kernel for the baseline function and (ii) a constant treatment response. It uses marginal likelihood for learning and \textsc{Gp} posteriors for inference.

\subsubsection{Semi-synthetic Dataset}
\label{ssec:semi-synth}

\begin{figure}[h!]
    \begin{subfigure}[t]{.49\textwidth}
        \centering
        \includegraphics[width=\linewidth]{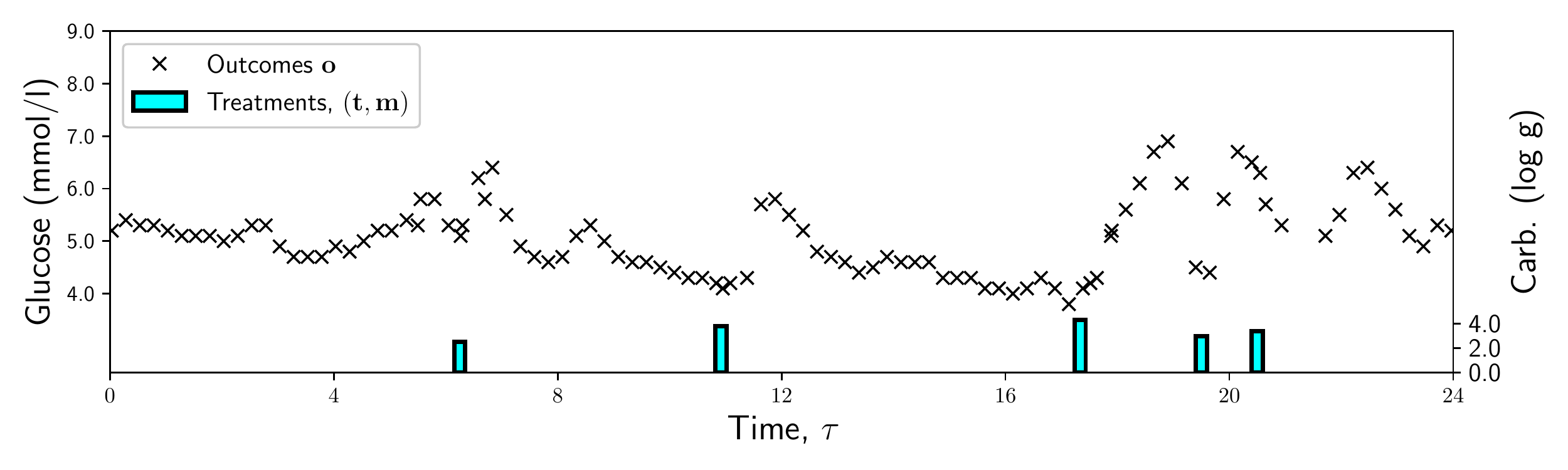}
        \caption{1-day treatment--outcome data of real-world Patient 3.}
     \end{subfigure}\hfill
     \begin{subfigure}[t]{.49\textwidth}
        \centering
        \includegraphics[width=\textwidth]{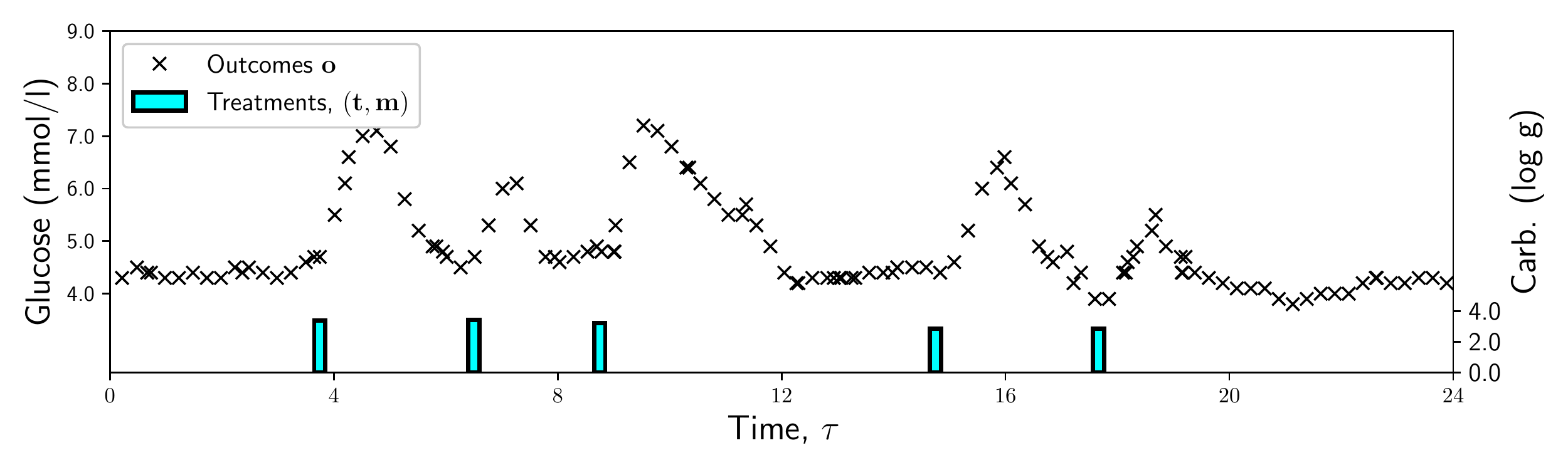}
        \caption{1-day treatment--outcome data of real-world Patient 8.}
     \end{subfigure}
     \begin{subfigure}[t]{.49\textwidth}
        \centering
        \includegraphics[width=\linewidth]{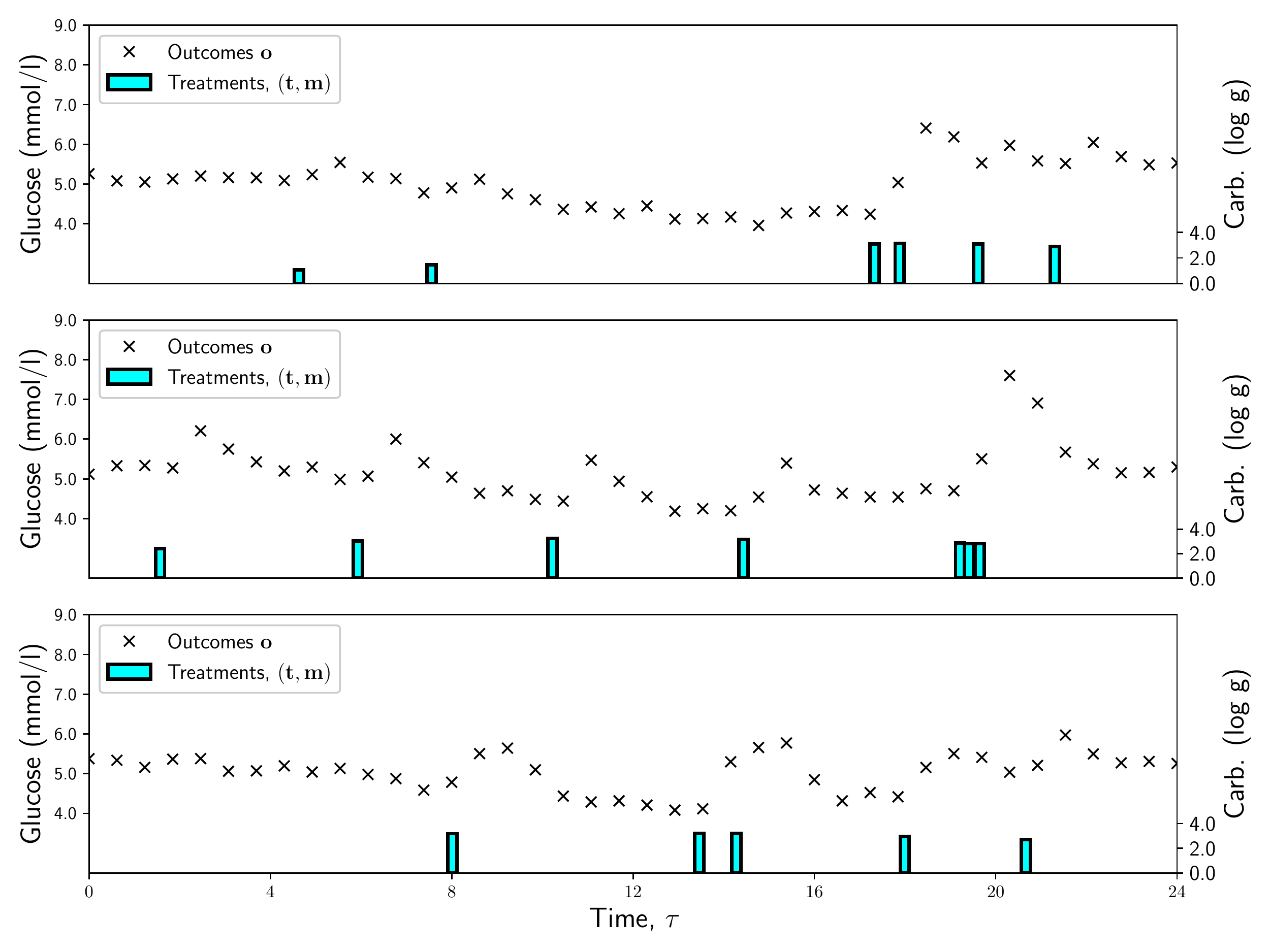}
        \caption{Outcome trajectory samples for 3 synthetic patients that use the baseline and response functions of Patient 3.}
     \end{subfigure}\hfill
     \begin{subfigure}[t]{.49\textwidth}
        \centering
        \includegraphics[width=\linewidth]{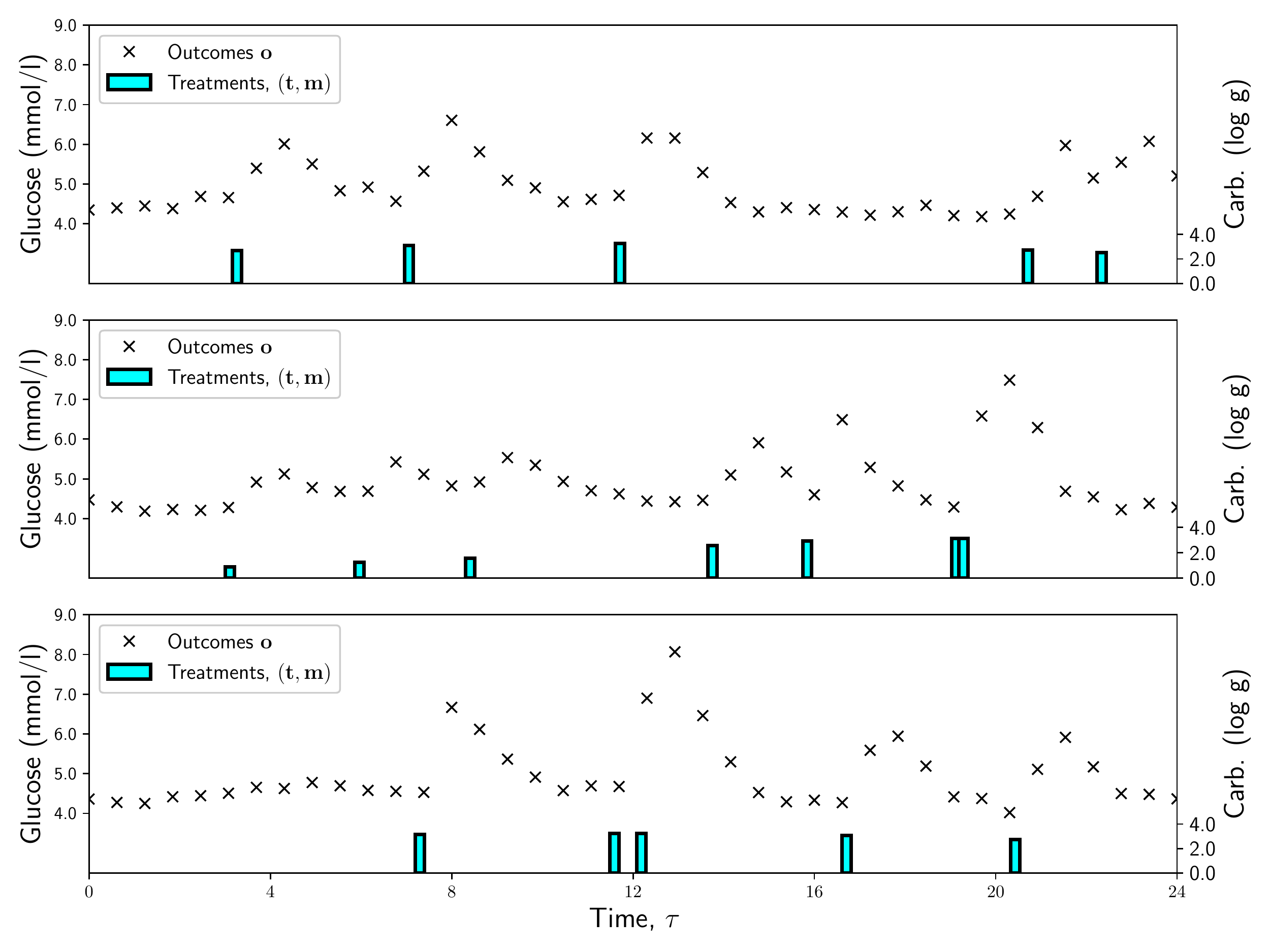}
        \caption{Outcome trajectory samples for 3 synthetic patients that use the baseline and response functions of Patient 8.}
     \end{subfigure}
    \caption{Observational outcome trajectory samples of patients belonging to two different patient groups: \textbf{(left)} patient group 1 and (\textbf{right}) patient group 2. \textbf{(a-b)} 1-day long examples of treatment--outcome sequences for Patient 3 and Patient 8, on which we learn the simulator functions for patient groups \{1,2\}. \textbf{(c-d)} Three joint trajectory samples from both patient groups.}
    \label{fig:synth_trajectory}
\end{figure}

We use the selected treatment and outcome simulators described in \cref{sssec:simulator} to simulate semi-synthetic samples from observational, interventional and counterfactual distributions. For the observational data set, we simulate 1-day long treatment--outcome sequences for 50 patients: $\mathcal{D} = \{\mathcal{D}^{(v)}\}_{v=1}^{50}$. We assume the outcome trajectory is measured in regular intervals, similar to a continuous glucose monitoring device. We divide the 1-day long observation period into 40 intervals and sample joint trajectories using the treatment and outcome simulators. We show some of the sampled outcome trajectories in \cref{fig:synth_trajectory}, together with 1-day long examples of treatment--outcome sequences of two real-world patients \{3,8\}. We train all benchmark models on the observational data set, while using interventional and counterfactual data sets for test.

\begin{figure}[h]
  \centering
  \begin{minipage}{1.0\linewidth}
    \begin{algorithm}[H]
        \caption{Ogata's Thinning algorithm for $P$ processes with Fixed Noise}
        \label{alg:ogata_fix_noise}
        \textbf{Input:} Start $T_1$, End $T_2$, Interval function $l(\cdot)$, $P$ conditional intensities $\{\lambda_{p}^*\}_{p=1}^P$. \\
        \textbf{Output:} Samples $\mathcal{T} = \{t_1, \ldots, t_n\}$ for $P$ point processes in the interval $[T_1, T_2]$.
        \begin{algorithmic}[1]
        \Function{Sample-fixed-noise}{$T_1, T_2, l(\cdot), \{\lambda_{p}^*\}_{p=1}^P$, P}
            \State $\mathcal{T}_p = \emptyset, \forall p \in \{1,\cdots,P\}$. \Comment{Initialize point sets.}
            \State $\tau = T_1$. \Comment{Initialize time.}
            \While{$\tau < T_2$}
                \State $\lambda_{ub} = \sup_{s \in [\tau, \tau+l(\tau)]} \{\lambda^*_p(s): p \in \{1, \cdots, P\}\}$ \Comment{Shared upper-bound.}
                \State $u_{ub}, u_a  \sim \mathcal{U}(0,1)$.  \Comment{Sample and fix noise var. for current interval.}
                \State $t_i = -1/\lambda_{ub} * \log(u_{ub})$  \Comment{Draw the inter-arrival time, $t_i \sim Exp(\lambda_{ub})$.}
                \If {$t_i \leq l(\tau)$} \Comment{Candidate in the interval.}
                    \For {$p \in \{1, \cdots, P\}$}
                        \If {$u_a \leq \lambda^*_p(\tau+t_i) / \lambda_{ub}$} \Comment{Keep with probability $\lambda^*_p/\lambda_{ub}$ for each process $p$.}
                            \State $\mathcal{T}_p = \mathcal{T}_p \cup \{\tau+t_i\}$. \Comment{Point accepted.}
                        \EndIf
                    \EndFor
                    \State $\tau = \tau + t_i$. \Comment{Continue from the candidate point.}
                \Else  \Comment{No candidate point in the interval.}
                    \State $\tau = \tau + l(\tau)$. \Comment{Continue from the end of the interval.}
                \EndIf
            \EndWhile
            \State \Return $\{\mathcal{T}_p\}_{p=1}^P$.
        \EndFunction
        \end{algorithmic}
    \end{algorithm}
  \end{minipage}
\end{figure}

We train our hierarchical outcome model on the observational data set to obtain estimated outcome model components for all patients: $\{\hat{f}^{(v)}_b(\tau), \hat{f}^{(v)}_a(\tau; \mathbf{a}), \hat{\epsilon}^{(v)}(\tau)\}_{v=1}^{50}$. For each policy $\pi \in \{\pi_A, \pi_B\}$, we train a treatment model on the population data which follows the target policy: $\{\hat{\lambda}^{*}_{A}(\tau), \hat{\lambda}^{*}_{B}(\tau)\}$. Overall, we have the following estimated model components: $\mathcal{M}_{est} = \{\hat{\lambda}^{*}_{A}(\tau), \hat{\lambda}^{*}_{B}(\tau)\} \cup \{\hat{f}^{(v)}_b(\tau), \hat{f}^{(v)}_a(\tau; \mathbf{a}), \hat{\epsilon}^{(v)}(\tau)\}_{v=1}^{50}$.

\paragraph{Test datasets and metric.} After learning model components on the observational data, we sample test data sets from the interventional and counterfactual distributions resulting from the policy interventions $[\Tilde{\pi}_{>1d}]$ and $[\Tilde{\pi}_{\leq 1d}]$ respectively. For the interventional data set, we sample the second day of each patient under the policy intervention $[\tilde{\pi}_{>1d}]$, where the treatment policy of the patient is switched. For the counterfactual data set, we condition on the observed data $\mathcal{D}^{(r)}$ for each patient and sample a hypothetical first day under a policy intervention $[\tilde{\pi}_{\leq 1d}]$, using the posterior of the noise variables. This is performed using the simulator functions and the counterfactual sampling algorithm detailed in \cref{sssec:cf_algo}.

More specifically, our policy intervention target has the following form:
\begin{align}
\label{eq:target-metric}
    P(\mathbf{Y_q}[\tilde{\pi}_{>T}] \mid \mathcal{H}_{\leq T})
    = \sum_{\tilde{\mathbf{a}}_{>T}} \prod_{k=0}^{m-1} & \underbrace{P(\tilde{\mathbf{a}}_{[q_k, q_{k+1})} \mid \tilde{\pi}_{>T}, \mathcal{H}_{\leq q_k})}_{\text{Treatment Model}} \underbrace{P(Y_{q_k} \mid \tilde{\mathbf{a}}_{[q_k, q_{k+1})}, \mathcal{H}_{\leq q_k})}_{\text{Outcome Model}} .
\end{align}
Notice that the marginal distribution of the outcome trajectory $\mathbf{Y_q}[\tilde{\pi}_{>T}]$ is not available in closed form, as we cannot integrate out treatments $\mathbf{a}$ in \cref{eq:target-metric}. Nevertheless, we have a generative treatment--outcome model, from which we can sample joint treatment--outcome trajectories to compare how close estimated outcome trajectories are to the ground-truth outcome trajectories in terms of the mean squared error (\textsc{Mse}). 

\begin{figure}[h]
    \centering
    \includegraphics[width=0.9\linewidth]{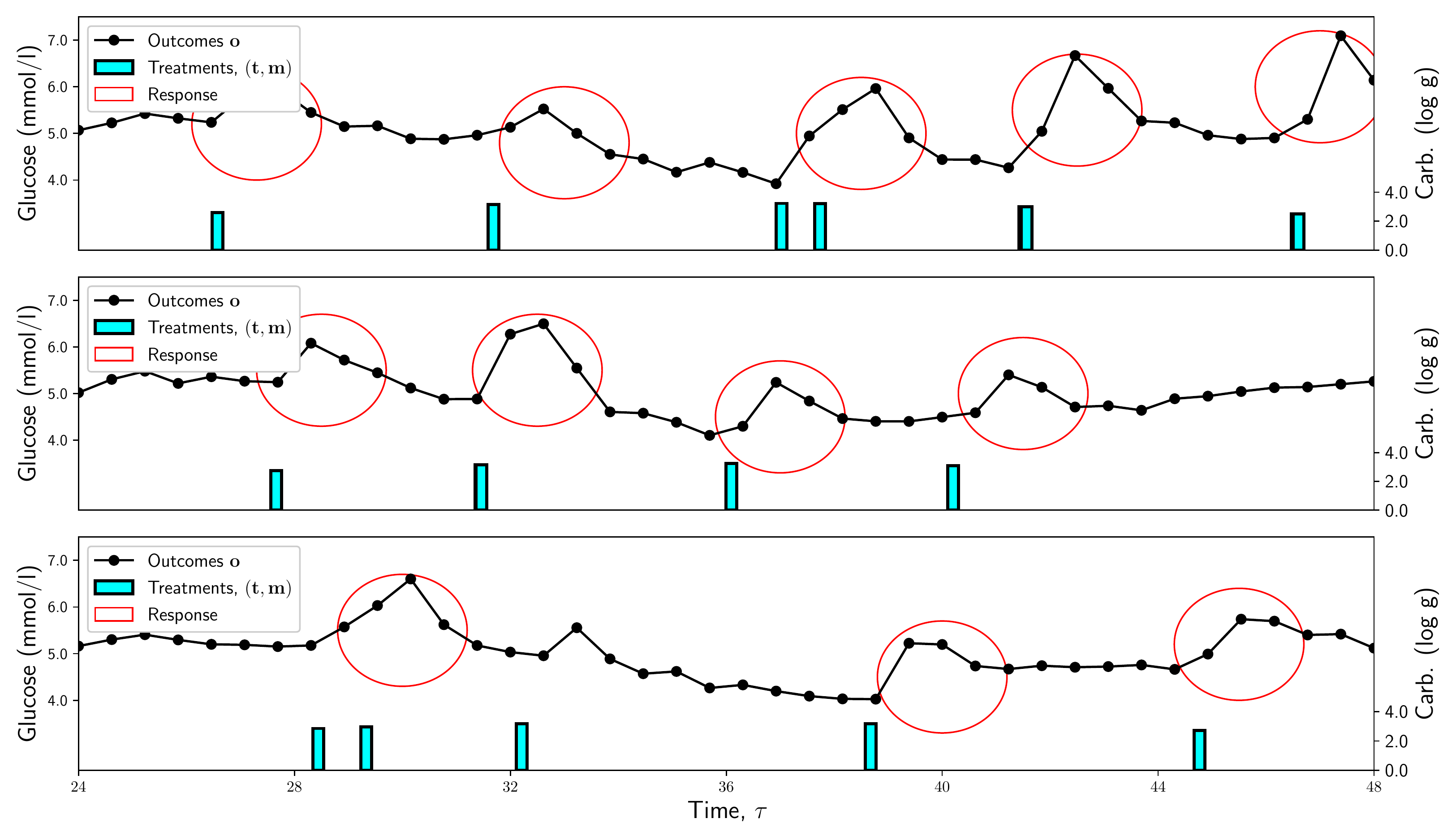}
    \caption{Three ground-truth i.i.d.~treatment samples from policy A and patient group 0. Notice that sampled treatments ({\color{cyan}cyan}) are not aligned, i.e., they do not occur at similar times each day. Hence, their treatment responses (annotated by {\color{red}red} circles) are also misaligned and estimating the \textsc{Mse} between them naively would lead to high values.}
    \label{fig:pp_stochasticity}
\end{figure}

Naively sampling joint trajectories leads to misaligned treatment times, and hence misaligned treatment response curves, no matter how close estimated intensities $\{\hat{\lambda}^{*}_{A}, \hat{\lambda}^{*}_{B}\}$ are to the ground-truth simulator intensities $\{\lambda^{*}_{A}, \lambda^{*}_{B}\}$. As an example, see three i.i.d.~treatment samples $\mathbf{a}^{(v)} \sim$ $p(\mathbf{a}^{(v)} \mid \lambda^{*}_{A}), v=1,...,V$ from the same simulator intensity  $\lambda^{*}_{A}$ in \cref{fig:pp_stochasticity}, where the misaligned treatment responses are annotated by red circles. This misalignment is due to the inherent stochasticity of the point process sampling.

The source of the stochasticity in the Ogata's sampling algorithm is two-folds: (i) the candidate point sampled from the upper-bound Poisson process and (ii) the uniform noise variable sampled for an accept/reject decision. To achieve aligned point process realisations and hence comparable treatment responses, we fix the noise variables responsible for both sources of randomness and sample a ground-truth and an estimated trajectory simultaneously. If an estimated intensity $\hat{\lambda}_{\text{est}}$ and a simulator (oracle) intensity $\lambda_{\text{oracle}}$ are close, the fixed-noise sampling algorithm will produce a similar set of treatments. In the case where two intensities $\hat{\lambda}_{\text{est}}$ and $\lambda_{\text{oracle}}$ are equal, we will obtain an equal set of treatments $\mathbf{a}^{(v)}_{\text{est}} \equiv \mathbf{a}^{(v)}_{\text{oracle}}, v=1,...,V$. The algorithm is shown in Algorithm \ref{alg:ogata_fix_noise}.

\begin{figure}[h]
    \centering
    \includegraphics[width=1.0\linewidth]{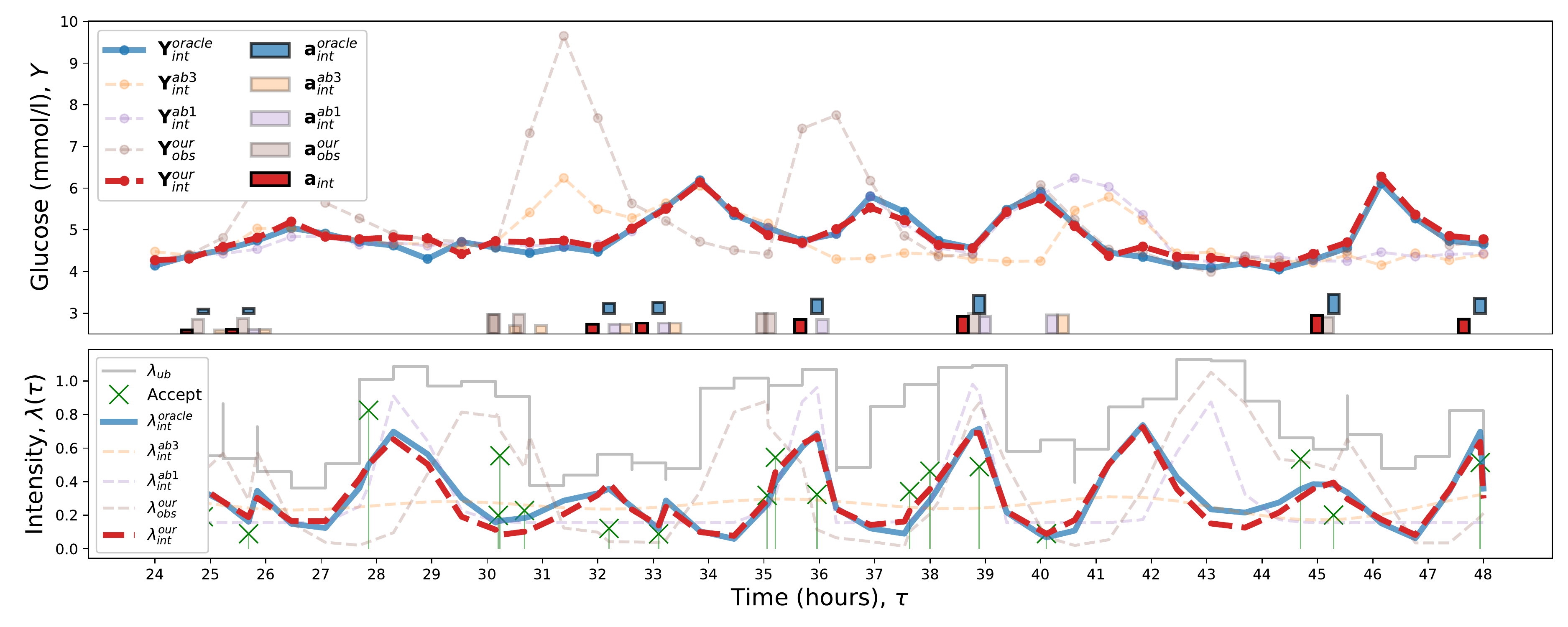}
    \caption{Interventional trajectory examples from the fixed-noise sampling algorithm (Algorithm \ref{alg:ogata_fix_noise}), where treatment noise variables $\mathcal{E}_{a}$ are fixed for the ground-truth and the estimated models. The ground-truth intensity $\lambda^{\text{oracle}}_{\text{int}}$, the ground-truth treatments $\mathbf{a}^{\text{oracle}}_{\text{int}}$ and the ground-truth trajectory $\mathbf{Y}^{\text{oracle}}_{\text{int}}$ are shown in {\color{blue}blue}, while the predictions of our model \textsc{Our}$_{\text{int}}$ are shown in {\color{red}red}. The estimated treatment intensity $\lambda^{\text{our}}_{\text{int}}$ and the estimated outcome trajectory $\mathbf{Y}^{\text{our}}_{\text{int}}$ follow the ground-truth intensity and trajectory well, while benchmark models fail to do so.}
    \label{fig:pair_sampling}
\end{figure}

For the policy intervention task, we illustrate the sampling process that compares the ground-truth (oracle) simulator with a set of estimated models (\textsc{Our}$_{\text{int}}$, \textsc{Our}$_{\text{obs}}$, \textsc{Ab1}$_{\text{int}}$, \textsc{Ab3}$_{\text{int}}$) in \cref{fig:pair_sampling}, where the subscripts `obs' (observational) and `int' (interventional) denote the target distribution. Similarly, the superscripts `oracle', `ab1', `ab3' and `our' denote the target model. The ground-truth intensity $\lambda^{\text{oracle}}_{\text{int}}$, the ground-truth treatments $\mathbf{a}^{\text{oracle}}_{\text{int}}$ and the ground-truth trajectory $\mathbf{Y}^{\text{oracle}}_{\text{int}}$ are shown in blue, while the predictions of our model \textsc{Our}$_{\text{int}}$ are shown in red. For the model \textsc{Our}$_{\text{int}}$, the estimated treatment intensity $\lambda^{\text{our}}_{\text{int}}$ and the outcome trajectory $\mathbf{Y}^{\text{our}}_{\text{int}}$ follow the ground-truth intensity and trajectory well. For the model \textsc{Our}$_{\text{obs}}$, the estimated treatment distribution is different than the ground-truth treatment distribution and the estimated trajectory $\mathbf{Y}^{\text{our}}_{\text{int}}$ is not close to the ground-truth trajectory. For the model \textsc{Ab3}$_{\text{int}}$, we see that the history-independent (\textsc{Nhpp}) intensity $\lambda^{ab3}_{int}$ does not take past treatments and outcome into account and only predicts a conservative approximation on the treatment distribution. The Gamma-based treatment intensity $\lambda^{ab1}_{\text{int}}$ of the model \textsc{Ab1}$_{\text{int}}$ is a better approximation than the history-independent intensity $\lambda^{ab3}_{\text{int}}$, however, it lacks the flexibility of our non-parametric intensity $\lambda^{\text{our}}_{\text{int}}$.

\end{document}